\documentclass{article}

\usepackage{float}
\usepackage{microtype}
\usepackage{graphicx}
\usepackage{subcaption}
\usepackage{enumitem}
\usepackage{booktabs} 
\usepackage{algorithmic} 

\usepackage{hyperref}

\usepackage[preprint]{icml2026}

\usepackage{amsmath}
\usepackage{amssymb}
\usepackage{mathtools}
\usepackage{amsthm}
\usepackage{placeins}
\usepackage{multirow}
\usepackage{amsmath}

\usepackage[capitalize,noabbrev]{cleveref}

\theoremstyle{plain}

\theoremstyle{definition}

\theoremstyle{remark}

\usepackage[textsize=tiny]{todonotes}
\icmltitlerunning{CASTER: Context-Aware Strategy for Task Efficient Routing}

\begin{document}

\twocolumn[
  \icmltitle{CASTER: Breaking the Cost-Performance Barrier in Multi-Agent Orchestration via Context-Aware Strategy for Task Efficient Routing}



  \icmlsetsymbol{equal}{*}

  \begin{icmlauthorlist}
    \icmlauthor{Shanyv Liu}{upc}
    \icmlauthor{Xuyang Yuan}{upc}
    \icmlauthor{Tao Chen}{upc}
    \icmlauthor{Zijun Zhan}{uh}
    \icmlauthor{Zhu Han}{uh}
    \icmlauthor{Danyang Zheng}{swjtu}
    \icmlauthor{Weishan Zhang}{upc,upcTeacher}
    \icmlauthor{Shaohua Cao}{upc,upcTeacher}
  \end{icmlauthorlist}
 \icmlaffiliation{upc}{Qingdao Institute of Software, College of Computer Science and Technology, China University of Petroleum (East China)}
  \icmlaffiliation{upcTeacher}{Shandong Key Laboratory of Intelligent Oil \& Gas Industrial Software}
  \icmlaffiliation{swjtu}{School of computing and artificial intelligence, Southwest Jiaotong University}
  \icmlaffiliation{uh}{Electrical and Computer Engineering Department, University of Houston}
  \icmlcorrespondingauthor{Shaohua Cao}{shaohuacao@upc.edu.cn}

  \icmlkeywords{Machine Learning, ICML, Multi-Agent Systems, Large Language Models, Dynamic Model Routing, Cost-Efficient Inference, LLM Serving Optimization, Dynamic Resource Allocation}

  \vskip 0.3in
]



\printAffiliationsAndNotice{}  

\begin{abstract}

Graph-based Multi-Agent Systems (MAS) enable complex cyclic workflows but suffer from inefficient static model allocation, where deploying strong models uniformly wastes computation on trivial sub-tasks. We propose CASTER (\textbf{C}ontext-\textbf{A}ware \textbf{S}trategy for \textbf{T}ask \textbf{E}fficient \textbf{R}outing), a lightweight router for dynamic model selection in graph-based MAS. CASTER employs a Dual-Signal Router that combines semantic embeddings with structural meta-features to estimate task difficulty. During training, the router self-optimizes through a Cold Start to Iterative Evolution paradigm, learning from its own routing failures via on-policy negative feedback. Experiments using LLM-as-a-Judge evaluation across Software Engineering, Data Analysis, Scientific Discovery, and Cybersecurity demonstrate that CASTER reduces inference cost by up to \textbf{72.4\%} compared to strong-model baselines while matching their success rates, and consistently outperforms both heuristic routing and FrugalGPT across all domains.
\end{abstract}

\section{Introduction}

\textbf{From Multi-Agent Collaboration to the Cost-Performance Paradox.} The evolution of Large Language Models (LLMs) has shifted the AI frontier toward Multi-Agent Systems (MAS). By decomposing complex objectives into sub-tasks, MAS achieves emergent intelligence essential for long-horizon domains like software engineering \cite{hong2023metagpt} and scientific discovery \cite{zhou2024hypothesis}. However, this scalability is constrained by the Cost-Performance Paradox. MAS workflows generate exponential context accumulation \cite{packer2023memgpt}, forcing a rigid binary choice: relying exclusively on Strong Models (e.g., GPT-4o) incurs prohibitive costs and latency \cite{chen2023frugalgpt}, while switching to Weak Models introduces a "fragility of logic," where a single upstream error cascades into total task failure \cite{yao2022react}. Balancing this trade-off is a significant challenge for industrial MAS adoption.

\textbf{Limitations of Existing Routing.} Current routing techniques are ill-suited for the dynamic nature of MAS. Heuristic approaches relying on static metrics like query length often fail to capture semantic complexity, as a concise, logic-heavy prompt frequently demands more reasoning power than a lengthy summarization task. Similarly, cascading strategies such as FrugalGPT \cite{chen2023frugalgpt} employ a "try-and-fail" mechanism that introduces unacceptable latency and risks polluting the shared context with erroneous intermediate steps, thereby confusing subsequent agents. Furthermore, preference-based methods like RouteLLM \cite{ong2024routellm}, which depend on human feedback (RLHF) or chatbot arena data, prove inadequate for MAS; while effective for aligning single-turn conversations with subjective user preferences, they lack the objective precision required for the rigorous, multi-step reasoning chains essential to agentic workflows.

\begin{figure*}
    \centering
    \includegraphics[width=1\linewidth]{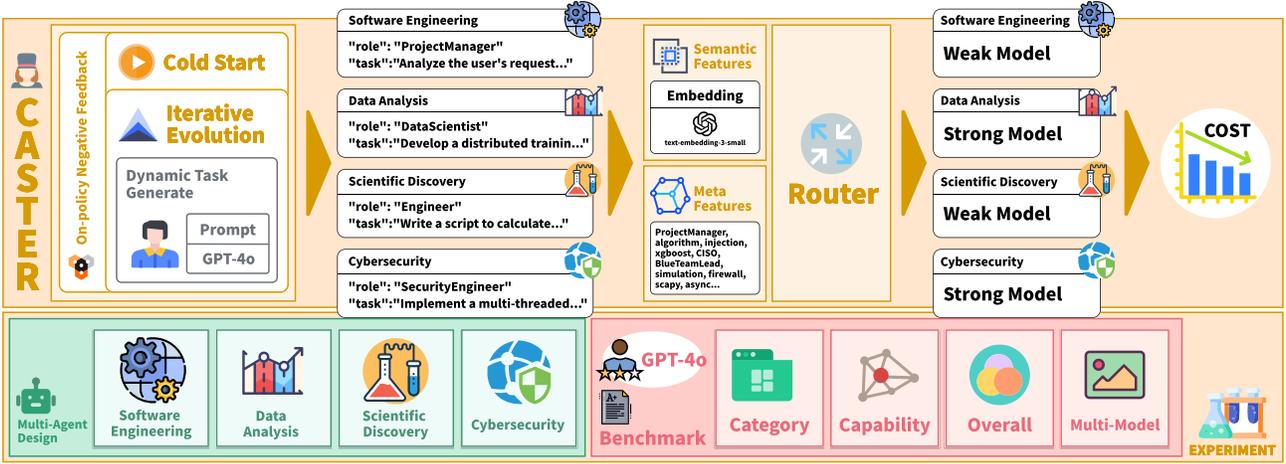}
    \caption{\textbf{\textbf{The overall architecture of the CASTER framework.}} The system begins with mock data and dynamic task generation via GPT-4o. The core Router integrates semantic and meta-features to dispatch tasks, evolving through cold start and on-policy negative feedback mechanisms. Tasks are executed by domain-specific agents (Software, Data, Science, Security) and evaluated against a comprehensive benchmark to ensure multi-model capability.}
    \label{fig:frame}
\end{figure*}

\textbf{Our Approach: Context-Aware Neural Routing.} We propose CASTER (\textbf{C}ontext-\textbf{A}ware \textbf{S}trategy for \textbf{T}ask \textbf{E}fficient \textbf{R}outing), a lightweight neural module designed to break the rigid trade-off between performance and cost. Unlike static configurations, CASTER acts as a dynamic decision-maker, mapping task semantics, agent roles, and evolving context to the most cost-effective model. By predicting the necessity of expert-level reasoning, it dispatches simple sub-tasks to weak models while reserving strong models for critical reasoning bottlenecks (System overview in \cref{fig:frame}).

\textbf{Contributions.} \begin{itemize} \item \textbf{Framework:} We introduce CASTER for MAS (e.g., LangGraph), integrating semantic embeddings with role-specific features for granular, dynamic model allocation. \item \textbf{Dataset:} We construct a comprehensive benchmark across four domains (Software, Data, Science, Security) with stratified difficulty levels to evaluate routing generalization. \item \textbf{Methodology:} We propose an On-Policy iterative training pipeline. We empirically demonstrate that naive random exploration leads to data pollution, validating our approach of labeling difficulty based on performance divergence. \item \textbf{Results:} Experiments show CASTER reduces token costs by \textbf{72.4\%} while maintaining success rates comparable to an all-GPT-4o baseline, significantly outperforming cascading and random strategies. \end{itemize}

\section{Related Work}

\textbf{LLMs-based Agents and Multi-Agent Systems.} The emergence of Large Language Models (LLMs) has catalyzed the development of autonomous agents capable of solving complex tasks. Early works focused on single-agent capabilities, but recent research has shifted towards Multi-Agent Systems (MAS) to emulate human-like collaboration. MetaGPT \cite{hong2023metagpt} and ChatDev \cite{qian2024chatdev} introduce Standard Operating Procedures (SOPs) into agent workflows, assigning specific roles (e.g., product manager, engineer) to LLMs to simulate software development companies. Similarly, AutoGen \cite{wu2024autogen} and Camel \cite{li2023camel} facilitate complex task solving through communicative agents and role-playing frameworks. Furthermore, LangGraph \footnote{https://github.com/langchain-ai/langgraph}advances this field by modeling agent workflows as stateful graphs, enabling cyclic execution and fine-grained control over multi-agent interactions. In the domain of software engineering, TaskWeaver \cite{qiao2023taskweaver} and SWE-agent \cite{yang2024swe} further specialize agents for code-first tasks and automated issue resolving. A comprehensive review of these LLM-based multi-agent systems and their applications in software engineering is provided by \cite{he2025llm}.

\textbf{Optimization of Multi-Agent Systems.} While foundation models like GPT-4 \cite{achiam2023gpt} and open-weight counterparts like Qwen \cite{bai2023qwen} serve as robust cognitive engines, the complexity of Multi-Agent Systems (MAS) demands optimization beyond individual model inference. Early efficiency efforts focused on cost-centric routing, such as FrugalGPT \cite{chen2023frugalgpt} and RouteLLM \cite{ong2024routellm}, which dynamically allocate queries between strong and weak models. However, recent advancements target the holistic optimization of agentic workflows. To automate system design, Agentic Supernet\cite{zhang2025multi} introduces an architecture search framework that identifies optimal agent topologies, reducing the reliance on manual engineering. Improving collaborative efficiency, OSC\cite{zhang2025osc} proposes a cognitive orchestration mechanism that dynamically aligns knowledge across agents to mitigate communication overhead. Furthermore, Multi-Agent Consensus Alignment\cite{samanta2025internalizing} internalizes the self-consistency of agents by leveraging consensus data, effectively distilling the reasoning capabilities of a swarm into a single efficient model.

\textbf{Benchmarks and Evaluation.} Evaluation paradigms have shifted from static response quality to dynamic agent behaviors. MT-Bench \cite{zheng2023judging} validated the "LLM-as-a-judge" approach for open-ended conversations. For autonomous agents, AgentBoard \cite{chang2024agentboard} introduced fine-grained progress metrics to measure incremental advancements in partially observable environments. Recently, MultiAgentBench \cite{zhu2025multiagentbench} expanded the scope to collective intelligence, systematically assessing coordination protocols and strategies within multi-agent workflows.

\textbf{Limitations of Prior Arts.} Despite these significant strides, a critical gap persists in the efficient deployment of MAS. First, existing efficient inference strategies like FrugalGPT and RouteLLM are predominantly tailored for single-turn queries or independent tasks. They often treat requests in isolation, failing to capture the evolving state dependencies and long-horizon context inherent in cyclic agent workflows. Secondly, while recent architectural optimizations (e.g., Agentic Supernet) address system topology, they lack the granularity to dynamically adjust resource allocation at the individual step level based on real-time task difficulty. Ultimately, while recent advancements in agent topology and communication protocols contribute to system efficiency, the performance ceiling of MAS remains fundamentally bounded by the intrinsic capabilities of the underlying LLMs\cite{zhu2025multiagentbench}. This necessitates a routing mechanism that directly addresses model capability rather than merely optimizing structure or conversational preference.

\section{Methodology}

\subsection{System Architecture} 
We build upon the LangGraph framework to implement a stateful, cyclic multi-agent workflow. Unlike linear chains, this graph-based structure supports the iterative loops essential for complex problem-solving. Within this graph, CASTER functions as a dynamic interceptor. Before control enters any agent node, CASTER analyzes the real-time shared state to determine the optimal model backend (e.g., GPT-4o vs. GPT-4o-mini). This granular, step-level routing not only optimizes costs on-demand but also enhances resilience: if a low-cost model leads to failure (e.g., rejection by a Reviewer), the workflow seamlessly rolls back with an adjusted strategy.

\subsection{CASTER Design}

\subsubsection{CASTER Architecture: Dual-Branch Feature Fusion Network}

In contrast to existing approaches relying on Reinforcement Learning from Human Feedback (RLHF) (such as RouteLLM \cite{ong2024routellm}), our method is better positioned to leverage user interactions within the chat domain. We implement a Dual-Branch Feature Fusion Network to explicitly model the interaction between semantic and meta-features. As shown in the implementation the model consists of two parallel extraction branches followed by a fusion classifier. We denote the learnable parameter set as $\theta = \{ \mathbf{W}_t, \mathbf{b}_t, \mathbf{W}_m, \mathbf{b}_m, \mathbf{W}_{fuse}, \mathbf{b}_{fuse}, \mathbf{w}_{out} \}$.

\textbf{Feature Extraction Branches.}
The model processes heterogeneous inputs through separate encoding streams:
\begin{itemize}
    \item \textbf{Text Branch (Semantic):} Given the input text (denoted as $X_{aug}$ or $T$), we extract its high-dimensional embedding $\mathbf{x}_{sem} \in \mathbb{R}^{D_{in}}$. This vector is projected into a latent space of dimension $D_{sem}$ via a dense layer with Dropout regularization to prevent overfitting:
    \begin{equation}
    \mathbf{h}_{sem} = \text{Dropout}(\text{ReLU}(\mathbf{W}_t \mathbf{x}_{sem} + \mathbf{b}_t)) \in \mathbb{R}^{D_{sem}}
    \end{equation}
    where $\mathbf{W}_t \in \mathbb{R}^{D_{sem} \times D_{in}}$.
    
    \item \textbf{Meta Branch (Structural):} The sparse meta-vector $\mathbf{v}_{meta} \in \mathbb{R}^{D_{meta}}$ is processed by a lightweight non-linear projection to capture basic feature interactions:
    \begin{equation}
    \mathbf{h}_{meta} = \text{ReLU}(\mathbf{W}_m \mathbf{v}_{meta} + \mathbf{b}_m) \in \mathbb{R}^{D_{struct}}
    \end{equation}
    where $\mathbf{W}_m \in \mathbb{R}^{D_{struct} \times D_{meta}}$.
\end{itemize}

\textbf{Fusion and Inference.}
The latent representations are concatenated to form a joint feature vector $\mathbf{h}_{joint} = [\mathbf{h}_{sem}; \mathbf{h}_{meta}] \in \mathbb{R}^{D_{sem} + D_{struct}}$. This vector is passed through a fusion layer to learn non-linear dependencies before the final probability estimation:

\begin{equation}
p(\text{Strong}|\mathbf{x}) = \sigma \left( \mathbf{w}_{out}^T \cdot \text{ReLU}(\mathbf{W}_{fuse} \mathbf{h}_{joint} + \mathbf{b}_{fuse}) \right)
\end{equation}

where $\mathbf{W}_{fuse} \in \mathbb{R}^{D_{fuse} \times (D_{sem} + D_{struct})}$ maps the joint vector to a hidden bottleneck of dimension $D_{fuse}$.

\subsection{Training Strategy}

We established a two-stage paradigm: "Cold Start" followed by "Iterative Evolution." The router is initialized with synthetic rule-based data for fundamental reasoning. Subsequently, we transition to the critical stage of real-world iteration. We observed that traditional "random exploration" strategies introduce significant noise—such as strong models successfully solving trivial tasks—which mislead the router into becoming overly conservative and expensive. Consequently, we discard random data collection in favor of On-Policy trajectory data derived from the current optimal router. We specifically target "high-value" boundary samples, such as failures caused by misjudgment or successful instances of cost reduction. This active learning mechanism, based on boundary cases, significantly enhances both training efficiency and the model's generalization capabilities.

\subsubsection{Pre-training and Cold Start Strategy}

In reinforcement learning or online learning systems, the stochastic nature of initial policies often leads to high exploration costs and slow convergence (the "Cold Start Problem"). To equip the CASTER with basic discriminative capabilities prior to real-world deployment, we propose a supervised pre-training method based on Heuristic Data Augmentation.

\textbf{Seed Dataset Construction.} We define three categories of seed tasks based on complexity:\\
\textbf{(1) Easy Tasks:} Basic syntactic operations, labeled for the \textit{Weak} model (Label $\approx$ 0.1). \\[3pt]
\textbf{(2) Medium Tasks:} Data cleaning and routine algorithms, labeled as fuzzy boundaries (Label $\approx$ 0.5). \\[3pt]
\textbf{(3) Hard Tasks:} Distributed architecture design and deadlock debugging, labeled for the \textit{Strong} model (Label $\approx$ 0.9).

\textbf{Automated Data Augmentation.} To overcome the sparsity of seed data, we designed a data augmentation engine. This engine expands a single seed entry into 4–6 training samples with identical semantics but varied phrasing by randomly combining instruction prefixes (e.g., "Write a Python script to...") with suffix constraints. Furthermore, we introduce Uniform noise $\epsilon \sim \mathcal{U}(-0.05, 0.05)$ to perturb the labels, preventing the model from overfitting to specific difficulty levels.

\textbf{Meta-Feature Simulation.} Since real-time runtime context is unavailable during the cold-start phase, we built a simulator to generate meta-features. Based on keyword matching (e.g., detecting "thread", "async") and task difficulty distributions, the simulator probabilistically generates corresponding Role Vectors and Context Lengths, providing a complete input view for the Wide \& Deep network. Through these methods, we constructed an initial training set of hundreds of samples with zero real-world data, enabling the router to achieve baseline accuracy upon deployment and effectively avoiding blind random exploration.

\subsubsection{Iterative Fine-tuning via Negative Feedback}

While the router acquires basic discriminative ability after the cold start, it may still misjudge complex edge cases in real-world scenarios. To further refine routing precision, we designed a fine-tuning pipeline based on Real-world Trajectories.

This pipeline introduces a "Negative Feedback Learning" mechanism, the core logic of which lies in the re-labeling of historical data:

\textbf{Reinforcing Success:} If a task is successfully solved and the label was \textit{Strong}, we reinforce it as a positive sample (Label=1.0). If the model was \textit{Weak}, it is treated as a negative sample (Label=0.0 relative to the "need for strong model" probability) to encourage low-cost paths.

\textbf{Correcting Failure:} This is the crux of the fine-tuning. If the system selected the \textit{Weak} model but the task ultimately resulted in failure (\verb|Outcome="FAILURE"| and \verb|Model="Fast"|), we forcibly rectify the Ground Truth of this sample to \textit{Strong} (Label=1.0).

Essentially, this mechanism signals to the neural network: \textit{"You chose a weak model to save costs on this task and failed; next time you encounter similar features, you must select the strong model."}

Furthermore, to prevent Catastrophic Forgetting during fine-tuning, we employ a dynamic learning rate adjustment strategy (StepLR), progressively decaying the learning rate ($\gamma=0.5$) as epochs increase, ensuring smoother convergence of model weights around the optimal solution.

\section{Results \& Analysis}

\subsection{Experiments Setup}

To validate robustness, we benchmarked CASTER across Software, Data, Science, and Cybersecurity domains against static baselines and the FrugalGPT cascade strategy. Using a multi-modal "LLM-as-a-Judge" framework, we assessed performance and cost efficiency. The system employs a lightweight router trained via offline pre-training and online iterative refinement. Primary experiments standardized on the Qwen and GPT-4o families; to mitigate self-preference bias\cite{wataoka2024self}, we extended generalization tests to diverse provider models, including Gemini, Claude, and DeepSeek. We also conducted a comparative evaluation against the FrugalGPT cascade strategy. Detailed configurations are provided in \cref{app:exp}.

Some experiments results are shown in \cref{app:results}. Likewise, \cref{tab:combined_cost_performance}, \cref{tab:caster_vs_frugal_combined}, \cref{tab:combined_cost_quality_LLM} provides an overview.

\subsection{Model Evaluation}

\begin{figure}[ht]
    \centering
    \begin{subfigure}{0.48\linewidth}
        \centering
        \includegraphics[width=\linewidth]{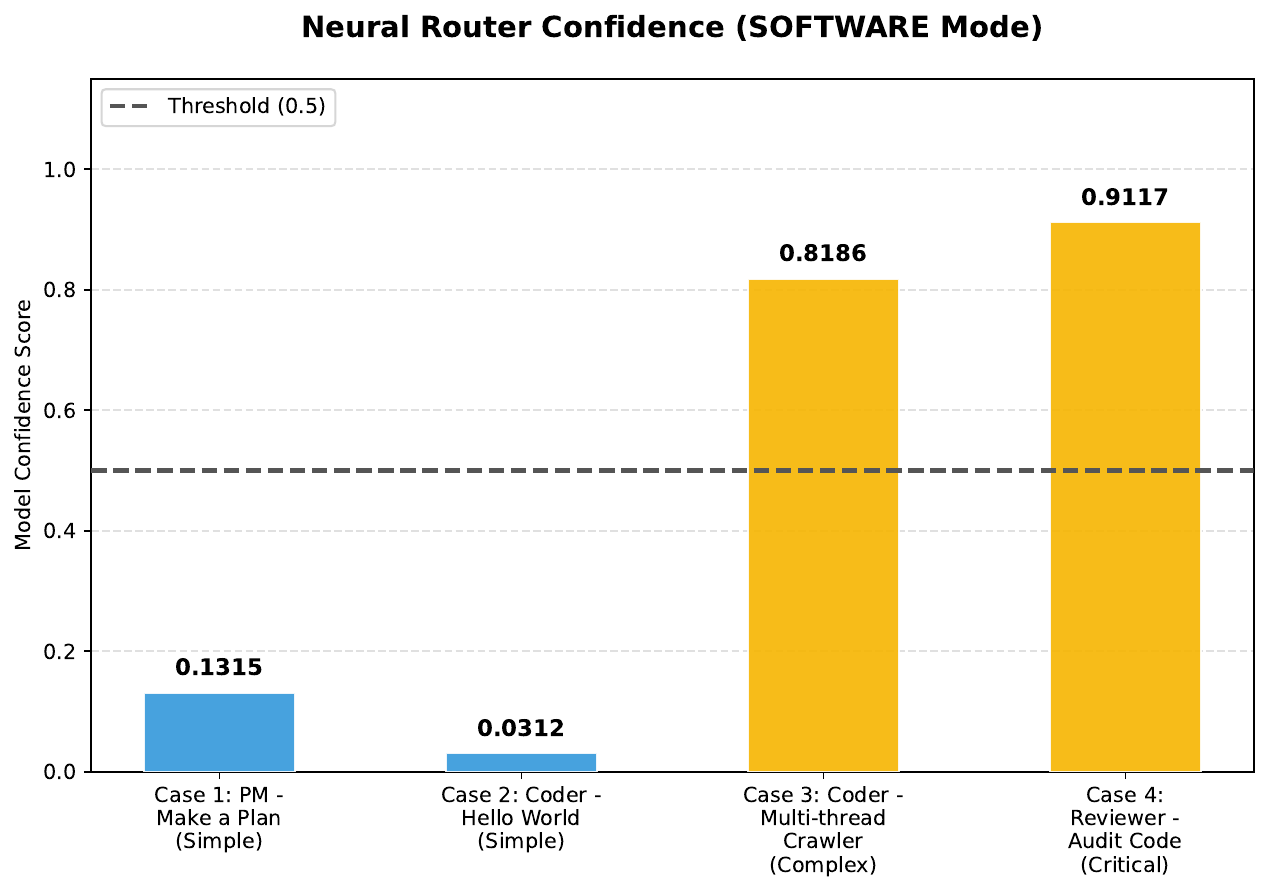}
        \caption{Software Engineering} 
        \label{fig:software_inf}
    \end{subfigure}
    \hfill 
    \begin{subfigure}{0.48\linewidth}
        \centering
        \includegraphics[width=\linewidth]{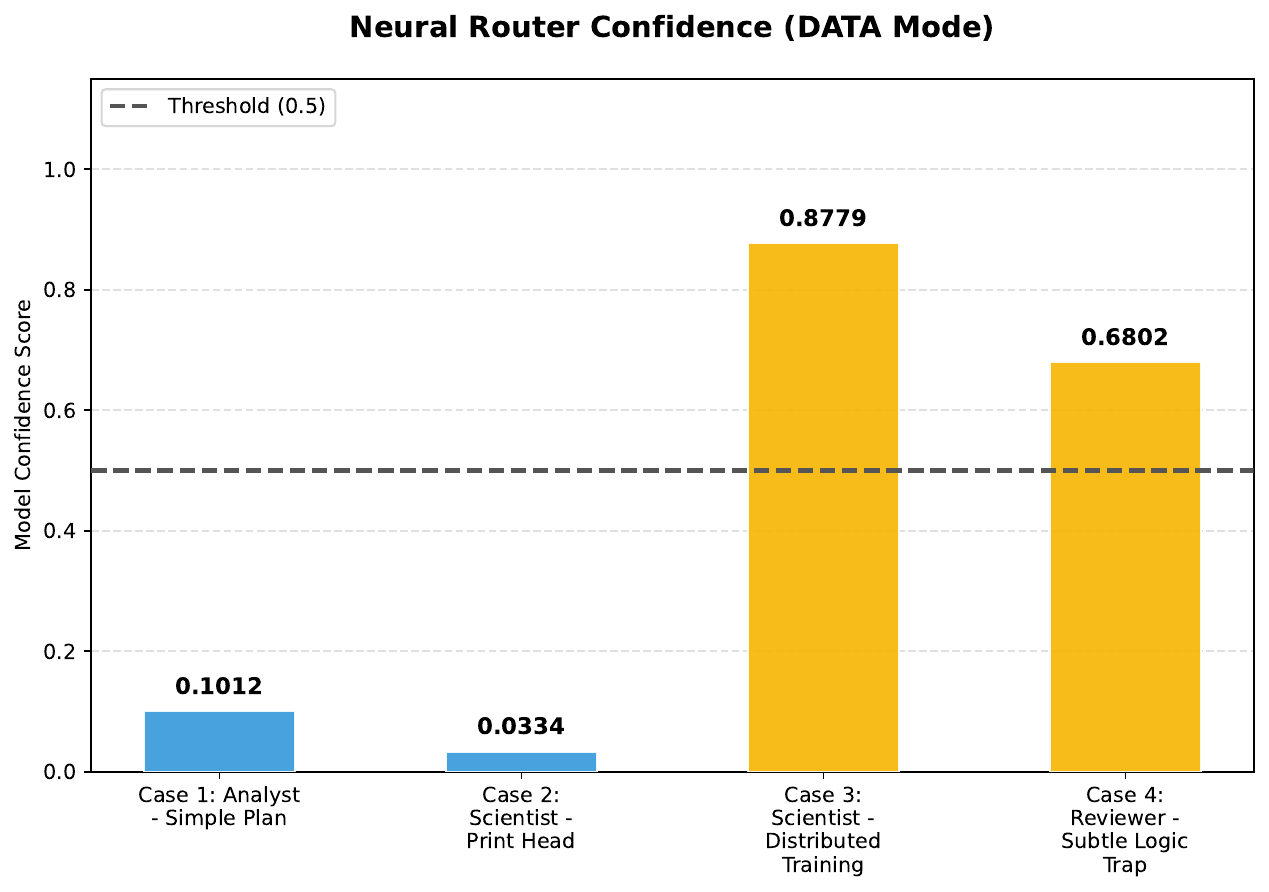}
        \caption{Data Analysis} 
        \label{fig:data_inf}
    \end{subfigure}
    \begin{subfigure}{0.48\linewidth}
        \centering
        \includegraphics[width=\linewidth]{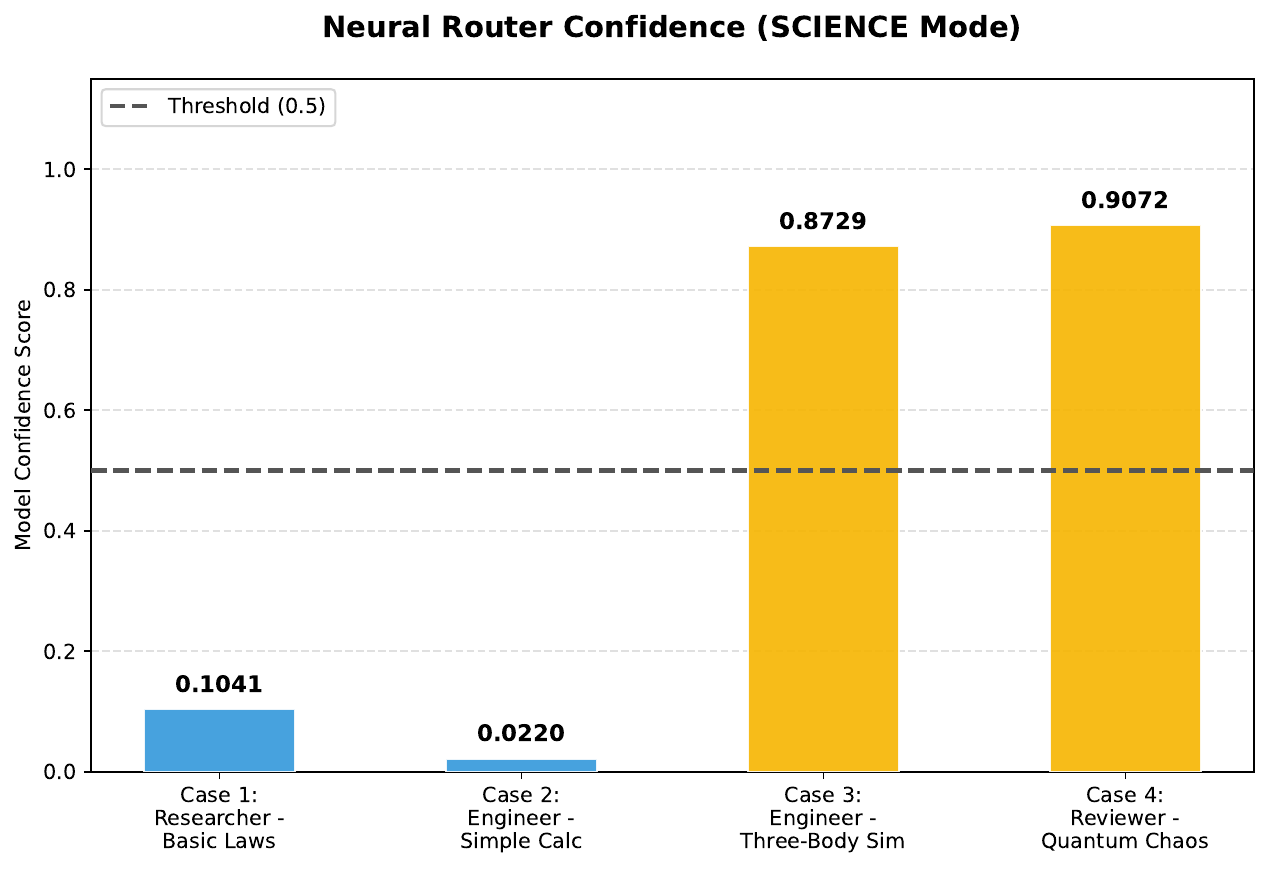}
        \caption{Scientific Discovery} 
        \label{fig:science_inf}
    \end{subfigure}
    \hfill 
    \begin{subfigure}{0.48\linewidth}
        \centering
        \includegraphics[width=\linewidth]{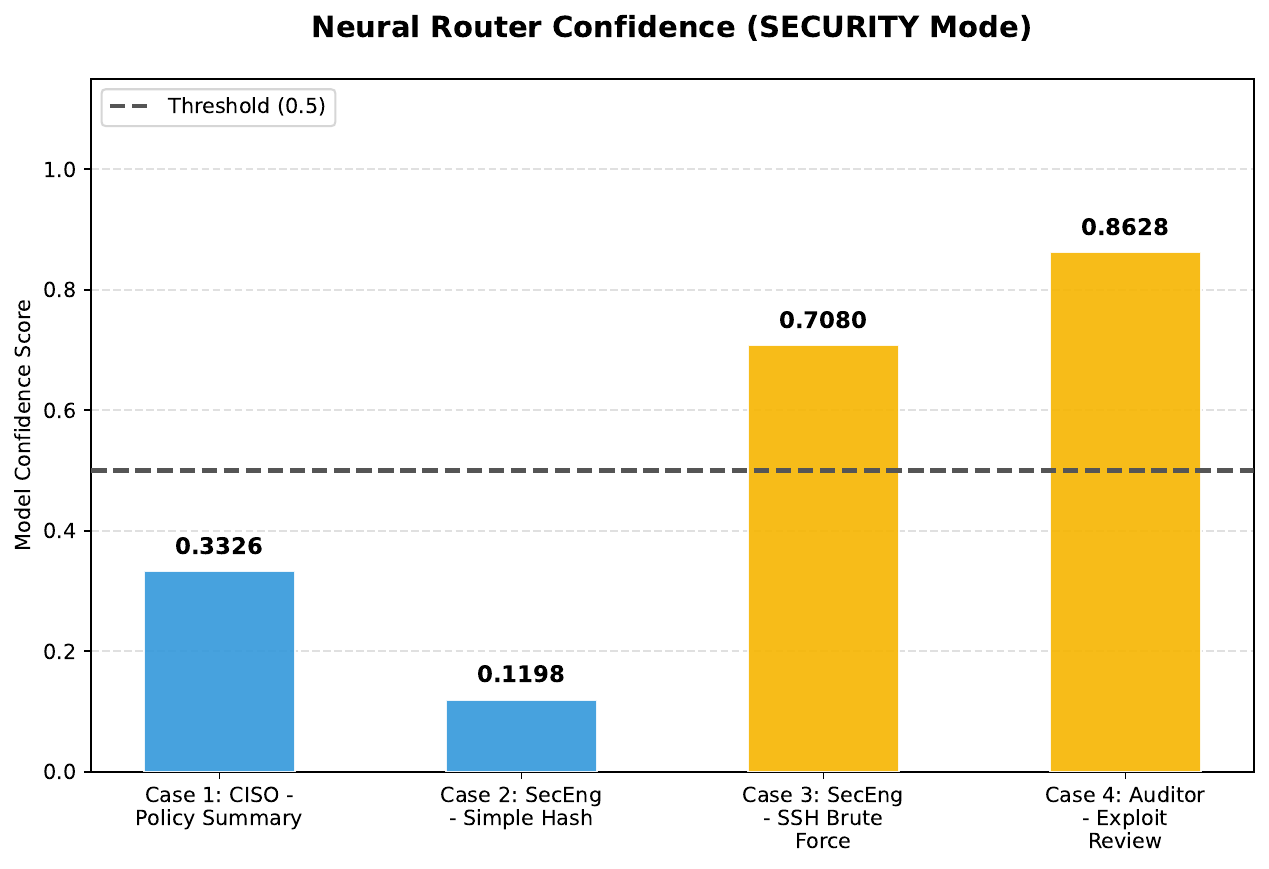}
        \caption{Cybersecurity} 
        \label{fig:security_inf}
    \end{subfigure}

    \caption{\textbf{CASTER Confidence Validation.} Inference scores across Software, Data, Science, and Security. The threshold ($y=0.5$, dashed) separates simple tasks (blue, Weak Model) from complex ones (yellow, Strong Model). Results confirm the router's efficacy in distinguishing task complexity.}
    \label{fig:router_inference}
\end{figure}

To validate the decision boundary, we analyzed CASTER's confidence scores across diverse domains (\cref{fig:router_inference}). Results demonstrate a sharp, intuition-aligned polarization: trivial tasks (e.g., "Hello World", "Simple Calc") yield minimal probabilities ($\approx 0.02$), routing to weak models. Conversely, complex scenarios like "Multi-thread Crawler" ($0.91$) or "Three-Body Sim" ($0.91$) trigger scores significantly above the threshold. In Security, the router successfully distinguishes routine summaries ($0.33$) from critical exploit reviews ($0.86$). This confirms that the router captures latent complexity features—such as logical dependency and operational risk—rather than merely memorizing templates.

\subsection{Cost Evaluation} 

\subsubsection{Cumulative Cost and Average Cost} 

We evaluated economic efficiency by analyzing both cumulative expenditure over sequential tasks and unit-level economics across all domains (\cref{fig:cumulative_cost_comparison}, \cref{tab:cost_efficiency}). The results demonstrate that CASTER effectively breaks the rigid trade-off between performance and cost. By dynamically offloading trivial queries, our method achieves a significant reduction in total expenditure—approximately 23.1\% to 54.4\% across all four domains—effectively flattening the cost curve compared to the Force Strong baseline. This efficiency is further corroborated by unit analysis(\cref{fig:avg_cost_comparison}, \cref{tab:avg_unit_cost}), where CASTER lowers average inference costs, exemplified by a drop from \$0.039 to \$0.018 per task in \textit{Software Engineering} and a 38.1\% reduction in \textit{Science}. In stark contrast to the Force Weak strategy, CASTER occupies a strategic "sweet spot," sustaining high-performance reasoning without the prohibitive financial overhead of static allocation.

\subsubsection{Cost Distribution} We visualized decision-making granularity via cost distribution (\cref{fig:cost_distribution}, \cref{tab:cost_distribution}). Unlike the \textit{Force Weak} baseline, which discriminately allocates minimal resources regardless of difficulty, CASTER demonstrates a broad dynamic range spanning the spectrum between weak and strong baselines. This confirms that the router achieves efficiency not by being uniformly "cheap," but by intelligently reserving expensive compute solely for high-complexity scenarios.

\begin{table*}[t]
    \centering
    \caption{\textbf{Comprehensive comparison of Cost, Reduction, and Quality.} This table integrates economic efficiency with performance metrics. CASTER consistently achieves significant cost reductions while maintaining (or surpassing) the quality scores of Strong baselines across most scenarios.}
    \label{tab:combined_cost_quality_LLM}
    
    \begin{minipage}{0.48\textwidth}
        \centering
        \resizebox{\linewidth}{!}{%
        \begin{tabular}{lllccc}
            \toprule
            & & & \multicolumn{3}{c}{\textbf{Metrics}} \\
            \cmidrule(lr){4-6}
            \textbf{Scenario} & \textbf{Model} & \textbf{Strategy} & \textbf{Cost (\$)} & \textbf{Red. (\%)} & \textbf{Score} \\
            \midrule
            \textbf{Software} & claude & Force Strong & 2.8204 & - & \textbf{100.0} \\
             &  & Force Weak & 0.8858 & 68.6\% & 93.3 \\
             &  & \textbf{CASTER} & 1.3738 & 51.3\% & 96.4 \\
             \cmidrule{2-6}
             & deepseek & Force Strong & 0.2346 & - & \textbf{98.6} \\
             &  & Force Weak & 0.1391 & 40.7\% & 98.3 \\
             &  & \textbf{CASTER} & 0.1106 & 52.9\% & 98.1 \\
             \cmidrule{2-6}
             & gemini & Force Strong & 1.5087 & - & \textbf{100.0} \\
             &  & Force Weak & 0.4412 & 70.8\% & 99.8 \\
             &  & \textbf{CASTER} & 0.8783 & 41.8\% & \textbf{100.0} \\
             \cmidrule{2-6}
             & openai & Force Strong & 1.4658 & - & 95.3 \\
             &  & Force Weak & 0.0498 & 96.6\% & 82.2 \\
             &  & \textbf{CASTER} & 0.4052 & 72.4\% & \textbf{97.0} \\
             \cmidrule{2-6}
             & qwen & Force Strong & 0.0397 & - & 99.3 \\
             &  & Force Weak & 0.0138 & 65.2\% & 97.8 \\
             &  & \textbf{CASTER} & 0.0186 & 53.1\% & \textbf{100.0} \\
            \midrule
            \textbf{Data} & claude & Force Strong & 3.2237 & - & 83.3 \\
             &  & Force Weak & 0.4641 & 85.6\% & 80.7 \\
             &  & \textbf{CASTER} & 0.9186 & 71.5\% & \textbf{84.6} \\
             \cmidrule{2-6}
             & deepseek & Force Strong & 0.2193 & - & 73.4 \\
             &  & Force Weak & 0.2268 & -3.4\% & 69.2 \\
             &  & \textbf{CASTER} & 0.1332 & 39.3\% & \textbf{78.1} \\
             \cmidrule{2-6}
             & gemini & Force Strong & 1.8052 & - & \textbf{53.6} \\
             &  & Force Weak & 0.4665 & 74.2\% & 52.1 \\
             &  & \textbf{CASTER} & 0.9466 & 47.6\% & \textbf{53.6} \\
             \cmidrule{2-6}
             & openai & Force Strong & 1.1514 & - & \textbf{75.5} \\
             &  & Force Weak & 0.0452 & 96.1\% & 73.1 \\
             &  & \textbf{CASTER} & 0.4533 & 60.6\% & 75.4 \\
             \cmidrule{2-6}
             & qwen & Force Strong & 0.0456 & - & 73.5 \\
             &  & Force Weak & 0.0136 & 70.2\% & 70.4 \\
             &  & \textbf{CASTER} & 0.0344 & 24.6\% & \textbf{73.6} \\
            \bottomrule
        \end{tabular}%
        }
    \end{minipage}
    \hfill
    \begin{minipage}{0.48\textwidth}
        \centering
        \resizebox{\linewidth}{!}{%
        \begin{tabular}{lllccc}
            \toprule
            & & & \multicolumn{3}{c}{\textbf{Metrics}} \\
            \cmidrule(lr){4-6}
            \textbf{Scenario} & \textbf{Model} & \textbf{Strategy} & \textbf{Cost (\$)} & \textbf{Red. (\%)} & \textbf{Score} \\
            \midrule
            \textbf{Science} & claude & Force Strong & 1.6530 & - & \textbf{96.7} \\
             &  & Force Weak & 0.4987 & 69.8\% & 94.8 \\
             &  & \textbf{CASTER} & 1.1596 & 29.8\% & 95.8 \\
             \cmidrule{2-6}
             & deepseek & Force Strong & 0.1648 & - & 84.6 \\
             &  & Force Weak & 0.1660 & -0.7\% & 89.5 \\
             &  & \textbf{CASTER} & 0.1429 & 13.3\% & \textbf{97.5} \\
             \cmidrule{2-6}
             & gemini & Force Strong & 1.1737 & - & 94.2 \\
             &  & Force Weak & 0.2311 & 80.3\% & \textbf{95.0} \\
             &  & \textbf{CASTER} & 0.9259 & 21.1\% & \textbf{95.0} \\
             \cmidrule{2-6}
             & openai & Force Strong & 1.2679 & - & 95.3 \\
             &  & Force Weak & 0.0550 & 95.7\% & 87.5 \\
             &  & \textbf{CASTER} & 1.0416 & 17.8\% & \textbf{95.4} \\
             \cmidrule{2-6}
             & qwen & Force Strong & 0.1026 & - & 96.7 \\
             &  & Force Weak & 0.0276 & 73.1\% & 97.5 \\
             &  & \textbf{CASTER} & 0.0695 & 32.3\% & \textbf{97.6} \\
            \midrule
            \textbf{Security} & claude & Force Strong & 2.1948 & - & 94.3 \\
             &  & Force Weak & 0.3346 & 84.8\% & 94.4 \\
             &  & \textbf{CASTER} & 0.9129 & 58.4\% & \textbf{95.1} \\
             \cmidrule{2-6}
             & deepseek & Force Strong & 0.0983 & - & 91.1 \\
             &  & Force Weak & 0.1118 & -13.7\% & 89.3 \\
             &  & \textbf{CASTER} & 0.1105 & -12.4\% & \textbf{94.8} \\
             \cmidrule{2-6}
             & gemini & Force Strong & 0.9837 & - & 95.6 \\
             &  & Force Weak & 0.1408 & 85.7\% & 94.8 \\
             &  & \textbf{CASTER} & 0.3064 & 68.9\% & \textbf{96.2} \\
             \cmidrule{2-6}
             & openai & Force Strong & 0.4886 & - & 93.7 \\
             &  & Force Weak & 0.0207 & 95.8\% & 92.9 \\
             &  & \textbf{CASTER} & 0.2234 & 54.3\% & \textbf{93.9} \\
             \cmidrule{2-6}
             & qwen & Force Strong & 0.0274 & - & \textbf{95.9} \\
             &  & Force Weak & 0.0100 & 63.5\% & 94.1 \\
             &  & \textbf{CASTER} & 0.0257 & 6.2\% & 95.2 \\
            \bottomrule
        \end{tabular}%
        }
    \end{minipage}
\end{table*}

\subsection{Quality Evaluation}

While cost reduction is desirable, maintaining quality parity is paramount. We conducted a comprehensive evaluation across four semantic dimensions (e.g., \textit{Functional Correctness} in Software, \textit{Scientific Validity} in Science, and \textit{Compliance} in Security), as detailed in \cref{fig:category_breakdown}, \cref{fig:overall_scores}, \cref{fig:capability_breakdown},\cref{fig:data_score_composition},\cref{tab:category_breakdown}, \cref{tab:overall_performance},\cref{tab:multidimensional_quality} \cref{tab:score_composition}.

\subsubsection{Overall Evaluation} Quantitative analysis (\cref{fig:overall_scores},\cref{tab:overall_performance}) confirms that CASTER effectively bridges the performance gap and, notably, surpasses the static strong baseline in specific domains. In \textit{Software Engineering} and \textit{Data Analysis}, CASTER achieves scores of 85.0 and 78.0, significantly recovering the quality drops seen in weak baselines (83.8 and 76.8) and closely approaching the upper bound. Crucially, in \textit{Science} and \textit{Security}, CASTER outperforms the \textit{Force Strong} strategy, achieving 95.3 (vs. 95.2) and 86.2 (vs. 85.5). These results validate that dynamic routing not only optimizes resource allocation but can also mitigate the "over-thinking" or overfitting sometimes exhibited by strong models on simpler sub-tasks.

\subsubsection{Granular Analysis: Categories, Capabilities, and Multi-Model Evaluate} Decomposing performance into fine-grained dimensions reveals that CASTER effectively mitigates the "worst-case" risks of weak models.

\textbf{Category Evaluation.} Granular analysis (\cref{fig:category_breakdown}, \cref{tab:category_breakdown}) identifies critical vulnerability points. In specialized scenarios like \textit{Web Security} and \textit{Software Concurrency}, weak baselines exhibit severe instability, plummeting to 48.0 and 67.0 due to logic deficits. CASTER effectively shields the system from these failures, restoring robust scores of 86.0 and 83.0, respectively. Furthermore, in categories like \textit{Software Security} and \textit{Data Analytics}, CASTER not only recovers performance but surpasses the strong baseline (reaching 98.0 and 91.0), demonstrating superior routing precision.

\textbf{Capability Evaluation.} Further multi-dimensional assessment (\cref{fig:capability_breakdown},\cref{tab:multidimensional_quality}) shows that weak models suffer from "logic collapse," particularly in \textit{Science Robustness} (16.0 vs. 19.2) and \textit{Science Parameter Accuracy} (35.2 vs. 38.5), failing to adhere to strict physical constraints. CASTER successfully mitigates these deficits, restoring \textit{Software Functional Correctness} to 34.5 (near the Strong baseline's 35.2). Notably, in Security, CASTER even outperforms the Strong baseline in \textit{Safety \& Ethical Compliance} (27.6 vs. 26.8), demonstrating that the router can leverage model-specific strengths to enhance safety protocols beyond static strong models.

\textbf{Multi-Model Results.} A multi-model component-wise inspection (\cref{fig:data_score_composition},\cref{tab:score_composition}) confirms that the decline in weak models (average score 226.2) is primarily driven by inferior CSV Data and Code Quality segments. CASTER effectively restores the mass of these specific components, achieving a total multi-modal score of \textbf{230.9}—virtually identical to the \textit{Force Strong} upper bound (232.4). This demonstrates the router's ability to recognize tasks requiring rigorous data handling, ensuring high-fidelity outputs across all granularities.

\begin{table}[t]
    \centering
    \caption{\textbf{Combined analysis of Unit Cost and Performance.} This table demonstrates the trade-off efficiency of CASTER. It significantly reduces average unit costs (by 23.4\%--54.3\%) compared to the Strong baseline while maintaining matching or superior performance scores across all domains.}
    \label{tab:combined_cost_performance}
    
    \resizebox{\columnwidth}{!}{%
    \begin{tabular}{llccc}
        \toprule
        & & \multicolumn{2}{c}{\textbf{Cost}} & \textbf{Performance}\\
        \cmidrule(lr){3-4} \cmidrule(lr){5-5}
        
        \textbf{Scenario} & \textbf{Strategy} & \textbf{Avg. Cost}& \textbf{Reduction}& \textbf{Avg. Score} \\
        \midrule
        
        \multirow{3}{*}{\textbf{Software}} 
        & Force Strong & \$0.0392 & - & \textbf{87.5} \\
        & Force Weak   & \$0.0029 & 92.6\% & 83.8 \\
        & \textbf{CASTER} & \$0.0179 & 54.3\% & 85.0 \\ 
        \midrule
        
        \multirow{3}{*}{\textbf{Data}} 
        & Force Strong & \$0.0466 & - & \textbf{78.5} \\
        & Force Weak   & \$0.0043 & 90.8\% & 76.8 \\
        & \textbf{CASTER} & \$0.0255 & 45.3\% & 78.0 \\ 
        \midrule
        
        \multirow{3}{*}{\textbf{Science}} 
        & Force Strong & \$0.1339 & - & 95.2 \\
        & Force Weak   & \$0.0054 & 96.0\% & 90.2 \\
        & \textbf{CASTER} & \$0.0831 & 37.9\% & \textbf{95.3} \\ 
        \midrule
        
        \multirow{3}{*}{\textbf{Security}} 
        & Force Strong & \$0.0064 & - & 85.5 \\
        & Force Weak   & \$0.0021 & 67.2\% & 83.5 \\
        & \textbf{CASTER} & \$0.0049 & 23.4\% & \textbf{86.2} \\ 
        \bottomrule
    \end{tabular}%
    }
\end{table}

\subsection{Compare to FrugalGPT}

\textbf{Economic Efficiency.} The "Double-Billing" Penalty. \cref{fig:cumulative_cost_comparison_with_frugal}, \cref{tab:frugal_vs_neural} exposes a critical vulnerability in FrugalGPT's "fail-then-retry" mechanism. When weak models fail on difficult queries (e.g., in Science and Security), the cascade triggers a fallback, incurring a "double-billing" penalty for \textit{both} the failed weak inference and the subsequent strong inference. Consequently, FrugalGPT's cost curve steepens significantly, eventually surpassing CASTER. In contrast, CASTER leverages predictive capability to identify complexity \textit{a priori}. By implementing "one-shot routing" for hard tasks, it eliminates redundant weak calls, proving that intelligent discrimination is economically superior to reactive cascading.

\textbf{Quality \& Capability.} Avoiding the "Good Enough" Trap. CASTER consistently outperforms FrugalGPT across all domains (\cref{fig:score_comparison_with_frugal}, \cref{fig:capability_comparison_with_frugal}, \cref{tab:frugal_vs_neural_performance}, \cref{tab:multidimensional_frugal_vs_neural}). While both strategies perform similarly on simple tasks, a distinct gap emerges in complex scenarios, driving overall score improvements of \textbf{+0.7 to +1.2 points}. This lead stems from the cascade's limitation: FrugalGPT often settles for "good enough" outputs that marginally pass thresholds but suffer from weak-model noise. Conversely, CASTER avoids this quality dilution by directly assigning complex tasks to the strong model. Deeper metric decomposition reveals that this advantage extends to critical dimensions such as \textit{Safety \& Compliance} (e.g., \textbf{25.9 vs. 23.6} in Security) and \textit{Scientific Validity} (e.g., \textbf{29.1 vs. 28.1} in Science). CASTER's preemptive routing ensures deliverables benefit from superior engineering standards, yielding solutions that are not just correct, but production-ready.

\begin{table}[t]
    \centering
    \caption{\textbf{FrugalGPT vs. CASTER.} CASTER achieves a \textit{Pareto-superior} outcome, reducing total costs by \textbf{20.7\%--48.0\%} while surpassing FrugalGPT's quality in all domains. Unlike cascade methods, it improves efficiency without performance trade-offs.}
    \label{tab:caster_vs_frugal_combined}
    
    \resizebox{\columnwidth}{!}{%
    \begin{tabular}{llcccc}
        \toprule
        & & \multicolumn{2}{c}{\textbf{Cost}} & \multicolumn{2}{c}{\textbf{Performance}} \\
        \cmidrule(lr){3-4} \cmidrule(lr){5-6}
        \textbf{Scenario} & \textbf{Strategy} & \textbf{Total Cost} & \textbf{Reduction} & \textbf{Avg. Score} & \textbf{Gain} \\
        \midrule
        
        \multirow{2}{*}{\textbf{Software}} 
        & FrugalGPT & \$1.11 & - & 79.8 & - \\ 
        & \textbf{CASTER} & \textbf{\$0.58} & \textbf{48.0\%} & \textbf{80.8} & \textbf{+1.0} \\ 
        \midrule
        
        \multirow{2}{*}{\textbf{Data}} 
        & FrugalGPT & \$0.66 & - & 72.3 & - \\ 
        & \textbf{CASTER} & \textbf{\$0.41} & \textbf{38.4\%} & \textbf{73.0} & \textbf{+0.7} \\ 
        \midrule
        
        \multirow{2}{*}{\textbf{Science}} 
        & FrugalGPT & \$0.91 & - & 90.9 & - \\ 
        & \textbf{CASTER} & \textbf{\$0.59} & \textbf{35.3\%} & \textbf{92.1} & \textbf{+1.2} \\ 
        \midrule
        
        \multirow{2}{*}{\textbf{Security}} 
        & FrugalGPT & \$0.29 & - & 81.2 & - \\ 
        & \textbf{CASTER} & \textbf{\$0.23} & \textbf{20.7\%} & \textbf{82.0} & \textbf{+0.8} \\ 
        \bottomrule
    \end{tabular}
    }
\end{table}

\subsection{Comparative Analysis of LLMs}

In this section, we present a comprehensive evaluation based on a curated benchmark of 10 tasks, balanced equally between simple and complex difficulty levels (five tasks each). Detailed experimental results are provided in the \cref{app:results}.

\subsubsection{Cost Analysis Across Providers} 
Our evaluation of cumulative and average costs (\cref{fig:cost_LLM_comp}, \cref{fig:avg_cost_LLM_comp}, \cref{tab:cost_LLM_comp}, \cref{tab:avg_cost_LLM_comp}) reveals critical insights into economic efficiency. First, the Strong baseline consistently exhibits steep cost trajectories across all domains, whereas CASTER demonstrates robust generalizability, achieving significant cost reductions ranging from \textbf{6.2\% to 72.4\%} for providers with distinct price tiers (excluding DeepSeek). We observe pronounced disparities among providers: Claude, OpenAI, and Gemini incur substantially higher costs than Qwen and DeepSeek. Notably, Claude is the most expensive and volatile due to frequent timeouts, with its per-task cost in Cybersecurity reaching \$0.219—nearly double OpenAI’s baseline even after CASTER’s 58.4\% reduction. In stark contrast, Qwen and DeepSeek display exceptionally smooth and affordable trajectories, with total accumulated costs consistently remaining below \$0.15. DeepSeek's identical pricing for strong and weak models links cost directly to token usage, resulting in a counter-intuitive "cost inversion". Ultimately, these findings validate CASTER's robustness as a cost-optimization framework, demonstrating particularly potent savings for providers where the price gap between strong and weak models is substantial.

\subsubsection{Performance Evaluation (Score)}

To address multi-agent timeouts, we adopted a "milestone" scoring methodology\cite{zhu2025multiagentbench} that credits partial sub-task completion (\cref{fig:score_LLM_comp}, \cref{tab:score_LLM_comp}). CASTER consistently matches or exceeds Strong baselines. Notably, the DeepSeek \textit{Force\_Strong} (R1) baseline underperformed due to reasoning-induced latency/timeouts, whereas CASTER achieved higher scores by leveraging the efficient DeepSeek-V3. Claude justified its premium cost with high quality despite occasional timeouts, while Gemini collapsed in Data Analysis despite strong performance elsewhere. Although \textit{Force\_Weak} occasionally outperformed strong models (e.g., Qwen in Science) due to stochasticity, it remains prone to catastrophic failures, highlighting CASTER's necessity for stability. The general score decline reflects the complexity of evaluating multi-modal artifacts in data analysis.

\section{Conclusion and Future Work}
We propose CASTER to address MAS cost challenges via semantic and meta-feature integration and On-Policy training. This framework achieves an optimal cost-intelligence balance. Validated across Software, Data, Science, and Security domains, CASTER is inherently domain-agnostic, offering significant potential for broader applications such as legal analysis and creative writing.

\section*{Impact Statement}

This paper introduces CASTER to enhance LLM efficiency. The primary societal impact is positive: by significantly reducing inference costs and computational resource usage, our method supports Green AI initiatives and lowers the carbon footprint of large-scale deployments. Additionally, by making high-performance agentic workflows more affordable, CASTER promotes the democratization of AI, enabling resource-constrained individuals and organizations to leverage advanced capabilities. We foresee no specific ethical risks beyond those inherent to the underlying LLMs.

\bibliography{example_paper}
\bibliographystyle{icml2026}

\newpage
\appendix
\onecolumn

\section{Experiment.}
\label{app:exp}

We compared CASTER against Force Strong and Force Weak strategies, evaluating metrics such as Token Cost, Success Rate, and Output Quality. While primary experiments utilized the Qwen and GPT-4o families, we further conducted extensive generalization tests on other leading architectures to verify cross-model adaptability.

\subsection{Experimental Implementation and Training Protocol}

To validate the effectiveness of our proposed Context-Aware Strategy for Task Efficient Routing (CASTER), we implemented a rigorous training strategy. The protocol consists of two distinct phases: offline supervised pre-training and online iterative refinement. The description of  training algorithm is provided in \cref{alg:hybrid_training}.

\subsubsection{Dataset Preparation}

\textbf{Synthetic Warm-up Dataset ($\mathcal{D}_{pre}$).} 
For the cold start phase, we constructed a focused seed dataset containing representative tasks across Easy, Medium, and Hard tiers. To overcome data sparsity, we applied the automated augmentation engine to expand this seed set by a factor of 4 to 6. Crucially, we injected Uniform noise $\epsilon \sim U(-0.05, 0.05)$ to the labels during augmentation. This prevents the router from overfitting to the discrete difficulty levels of the seed tasks, fostering a continuous probability distribution.

\textbf{Dynamic Trajectory Dataset ($\mathcal{D}_{traj}$).} 
To mitigate the critical scarcity of high-quality training trajectories, we developed an automated Dynamic Task Generator. This stochastic adversarial pipeline synthesizes a full spectrum of tasks—ranging from trivial routines to high-complexity challenges—to accurately map the decision boundary between weak and strong models. Executed within an isolated Workspace Sandbox to prevent conflicts, the system captures real-time interaction logs (including Task Context, Model Used, and Execution Outcome) to construct the Dynamic Trajectory Dataset ($\mathcal{D}_{traj}$). This closed-loop mechanism creates a self-evolving data flywheel, ensuring diverse and balanced training coverage. Detailed generation logic is provided in \cref{app:auto_dataset}.

\subsubsection{Training Protocol}

\textbf{Phase 1: Cold Start Pre-training.} 
    We initialized the CASTER using $\mathcal{D}_{pre}$. The model was trained for 200 epochs using Binary Cross Entropy (BCE) loss with an initial learning rate of \textbf{$1e^{-3}$}. This extended training duration ensures the model sufficiently fits the "Hard" samples which are under-represented in the initial distribution.
    
\textbf{Phase 2: Online Iterative Refinement.} 
    In this phase, we engaged the \textit{Negative Feedback Learning} mechanism. We employed the Re-labeling Logic to correct "False Negative" samples (where a weak model was chosen but failed) to a label of 1.0 (Strong). The router was fine-tuned on these trajectories using an Adam optimizer with a conservative learning rate of \textbf{$1e^{-4}$}. To ensure stable convergence, we utilized a StepLR scheduler, decaying the learning rate by a factor of $\gamma=0.5$ every 50 epochs.

\subsubsection{Implementation Details}

The CASTER is implemented as a lightweight Dual-Branch network using PyTorch. We provide the specific hyperparameter settings and their rationales below:

\textbf{Dimensionality Configuration.}
The input dimension $D_{in}$ is set to 1536, consistent with the output vector size of the \texttt{text-embedding-3-small} model used for semantic extraction. The meta-feature dimension $D_{meta}$ is fixed at \textbf{6}, which explicitly encodes the task structure: a 4-dimensional one-hot vector for agent roles (Product Manager, Architect, Engineer, Reviewer), combined with 1 normalized context length scalar and 1 high-risk keyword indicator.
To balance inference latency and representational capacity, we employed a bottleneck design for the hidden layers. The semantic branch projects high-dimensional text vectors into a compact $D_{sem}=128$ space (with Dropout $p=0.2$), while the sparse meta-features are mapped to $D_{struct}=16$. These are fused into a joint representation of 144 dimensions and then compressed into a bottleneck of $D_{fuse}=64$ before the final classification. This lightweight architecture ensures negligible overhead during routing.

\textbf{Experimental Setup.} 
Our experimental framework is built upon a diverse ecosystem of Large Language Models (LLMs) accessed via standardized APIs, ensuring reproducibility and simulating real-world Model-as-a-Service (MaaS) environments. The model configuration is divided into three specific phases:

\begin{itemize} 
    \item \textbf{Data Construction \& Accumulation:} We adopted a dual-model strategy for training data generation. GPT-4o served as the "Task Generator" (Questioner) to synthesize diverse problem sets with stratified difficulty. Subsequently, we primarily utilized the Qwen model family (qwen-max and qwen-plus) to generate high-quality supervision signals and difficulty labels for these tasks.
    
    \item \textbf{Primary Evaluation \& Benchmarking:} For the core comparative experiments, we standardized on the GPT-4o series, designating \texttt{gpt-4o} as the "Strong Model" baseline and \texttt{gpt-4o-mini} as the cost-effective "Weak Model". Crucially, consistent with the "LLM-as-a-Judge" paradigm, GPT-4o was also exclusively employed as the "Reviewer" to assess the multi-dimensional quality of the generated outputs.
    
    \item \textbf{Generalization Testing:} To verify the router's robustness across different architectures, we extended our evaluation to include other leading model families, including Gemini, Claude, and DeepSeek. 
\end{itemize}

\begin{table}[h]
\centering
\caption{Hyperparameter settings and structural specifications of the CASTER.}
\label{tab:hyperparams}
\begin{tabular}{lc}
\toprule
\textbf{Parameter} & \textbf{Value} \\
\midrule
\multicolumn{2}{l}{\textit{Network Architecture}} \\
Input Dimension ($D_{in}$) & 1536 \\
Meta Dimension ($D_{meta}$) & 6 \\
Semantic Hidden ($D_{sem}$) & 128 \\
Structural Hidden ($D_{struct}$) & 16 \\
Fusion Hidden ($D_{fuse}$) & 64 \\
Output Dimension & 1 \\
\midrule

\end{tabular}
\end{table}
\begin{table}[h]
    \centering
    \caption{Pricing structure for Large Language Models used in experiments. Costs are denoted in USD per 1 million tokens.}
    \label{tab:llm_pricing}
    \begin{tabular}{llcc}
        \toprule
        \textbf{Provider} & \textbf{Model} & \textbf{Input Cost (\$/1M)} & \textbf{Output Cost (\$/1M)} \\
        \midrule
        \textbf{OpenAI} & gpt-4o& \$3.750& \$15.000\\
         & gpt-4o-mini& \$0.225& \$0.900\\
        \midrule
        \textbf{Anthropic} & claude-sonnet-4-5& \$4.500& \$22.500\\
         & claude-3-5-haiku-20241022& \$1.500& \$7.500\\
        \midrule
        \textbf{Google} & gemini-2.5-pro& \$1.875& \$15.000\\
         & gemini-2.5-flash& \$0.450& \$3.750\\
        \midrule
        \textbf{DeepSeek} & deepseek-reasoner (R1)& \$0.825& \$2.550\\
         & deepseek-chat (V3.2)& \$0.825& \$2.550\\
        \midrule
        \textbf{Alibaba} & qwen3-max& \$0.440& \$1.780\\
         & qwen-plus& \$0.110& \$0.280\\
        \bottomrule
    \end{tabular}
\end{table}
\subsection{Basic Model Test}

Following the cold-start training phase, we conduct a preliminary evaluation to verify that the model possesses the foundational capabilities required for processing batch tasks, which in turn facilitates further performance improvements. Detailed testing procedures are provided in \cref{app:basicTest}.

\subsection{Benchmark}

We developed a comprehensive evaluation framework \cite{zheng2023judging,liu2023g} to assess the execution performance of agents across four distinct domains. Each domain comprises a curated suite of 20 manually selected tasks, balanced equally between "Easy" and "Hard" difficulty levels to ensure a rigorous assessment spectrum. Crucially, to validate true generalization, we enforced a strict separation between training and evaluation data: these benchmark tasks were explicitly held out from both the router's offline pre-training corpus and its online refinement pipeline, preventing any potential data leakage. While providing a unified standard, this module is specifically adapted to handle multi-modal outputs (e.g., tabular data and visualizations) inherent to data analysis tasks. For comparative analysis, we tailored the test subsets to specific experimental goals: the FrugalGPT comparison utilizes the 10 most challenging tasks per domain to test high-complexity reasoning, while the cross-provider LLM benchmarking employs a representative 10-task subset (5 Easy, 5 Hard) to evaluate generalizability. Detailed specifications are provided in \cref{app:benchmark}, with the respective benchmark prompts illustrated in \cref{fig:software_prompt}, \cref{fig:data_prompt}, \cref{fig:science_prompt}, and \cref{fig:security_prompt}.

\section{Multi-agent Design.}
\label{app:Multi-agent Design}
\begin{figure}[htbp]
    \centering
    
    \begin{subfigure}{1\linewidth}
        \centering
        \includegraphics[width=\linewidth, height=10cm, keepaspectratio]{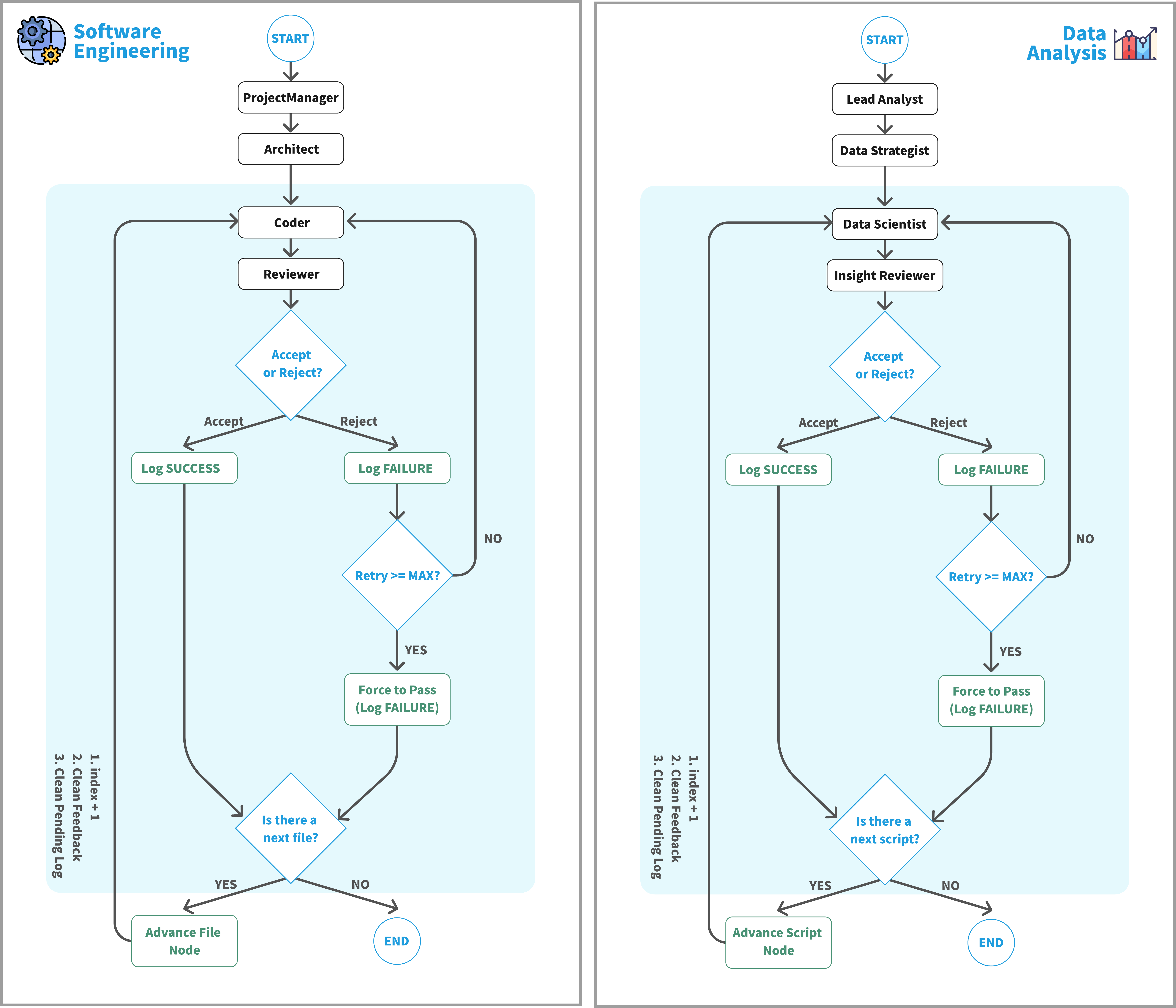}
    \end{subfigure}
    \begin{subfigure}{1\linewidth}
        \centering
        \includegraphics[width=\linewidth, height=10cm, keepaspectratio]{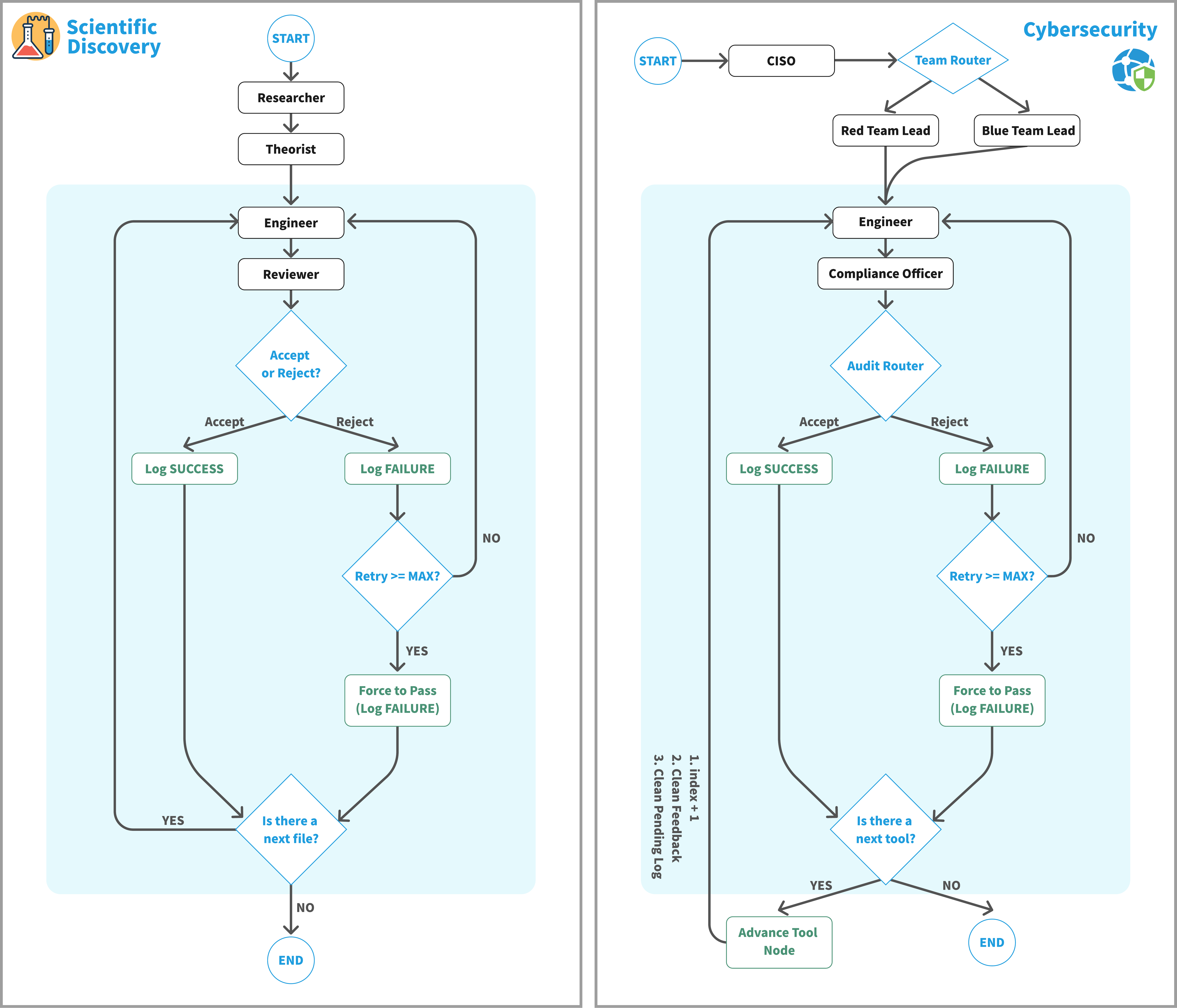}
    \end{subfigure}
    
    \caption{\textbf{Overview of the Domain-Specific Multi-Agent Workflows.}}
    \label{fig:agent_design}
\end{figure}
As illustrated in \cref{fig:agent_design}, we design domain-specific multi-agent workflows for diverse complex tasks. We depict the architectures for Software Engineering, Data Analysis, Scientific Discovery, and Cybersecurity, respectively. Despite their domain differences, these architectures collectively adhere to a "Linear Initialization + Iterative Loop" design pattern:

\begin{enumerate}[nosep]
    \item Initialization Phase: This phase defines the scope and strategy. 
    In the Software Engineering workflow, the \textit{Project Manager} and \textit{Architect} decompose tasks and design the system architecture. 
    In Data Analysis, the \textit{Lead Analyst} plans requirements, followed by the \textit{Data Strategist} for data profiling. 
    For Scientific Discovery, the \textit{Researcher} conducts a literature review, enabling the \textit{Theorist} to formulate precise hypotheses. 
    In Cybersecurity, the \textit{CISO} establishes the engagement rules, and a \textit{Team Router} assigns the mission to either a \textit{Red Team Lead} or \textit{Blue Team Lead} to generate a specialized toolset.
    
    \item \textbf{Iterative Execution Loop:} This is the core engine of the workflow. A domain-specific execution agent—such as a \textit{Coder}, \textit{Data Scientist}, \textit{Computational Scientist}, or \textit{Security Engineer}—generates the primary content (code, scripts, or simulations). This output is then rigorously evaluated by a corresponding \textit{Reviewer} (e.g., \textit{Peer Reviewer} or \textit{Compliance Officer}). The review process integrates our Reviewer Router logic:
    \begin{itemize}[nosep]
        \item \textbf{Credit Assignment:} Upon Acceptance or Rejection by the Reviewer, the system automatically logs a \textbf{SUCCESS} or \textbf{FAILURE} tag, facilitating data collection for training the CASTER.
        \item \textbf{Circuit Breaker:} To prevent infinite loops, a circuit breaker is triggered when the retry count reaches a maximum threshold. This forces a logical termination or transition while logging a "Failure" experience, ensuring system robustness and accumulating negative samples.
    \end{itemize}
    
    \item \textbf{State Transition:} For multi-step tasks, specialized utility nodes like \textit{advance\_file}, \textit{advance\_script}, or \textit{advance\_tool} (in Security) act as stateless managers. They are responsible for cleaning up feedback from the previous iteration, resetting retry counters, and advancing the task index, ensuring that the next sub-task begins in a clean and isolated state.
\end{enumerate}

\section{Automated Dataset for Training Construction via Adversarial Generation.}
\label{app:auto_dataset}
The scarcity of high-quality, open-ended evaluation datasets for vertical domains remains a significant bottleneck in Multi-Agent System (MAS) research. Existing benchmarks are often static, generic, or saturated, failing to adequately differentiate between the reasoning capabilities of state-of-the-art models. To address this, we introduce a Dynamic Adversarial Task Generation Pipeline. This pipeline leverages a strong teacher model (GPT-4o) to synthesize domain-specific, high-complexity challenges designed to stress-test agent workflows. (\cref{alg:task_gen,fig:task_gen})

\subsection{Stochastic Difficulty Stratification}
To mimic real-world request distributions, we implement a stochastic difficulty injection mechanism. The generator dynamically toggles between Normal Mode ($p=0.3$) and Hard Mode ($p=0.7$).
\begin{itemize}
\item \textbf{Normal Mode:} Focuses on standard, linear tasks (e.g., basic API implementation or descriptive statistics) to verify baseline functional correctness.
\item \textbf{Hard Mode (Expert Level):} Explicitly designed to widen the performance gap between strong and weak models. In this mode, the generator acts as an adversarial agent, injecting complex constraints that require multi-step reasoning, long-context understanding, and edge-case handling—areas where weaker models typically hallucinate or fail.
\end{itemize}

\subsection{Domain-Specific Adversarial Constraints}
A key contribution of our benchmark is the formulation of Domain-Specific Constraints that target the unique vulnerabilities of LLMs in each vertical field. As shown in \cref{alg:task_gen}, we prompt the generator with specialized "Personas" and "Critical Requirements":
\begin{itemize}[noitemsep, topsep=0pt, parsep=0pt, partopsep=0pt]
\item \textbf{Software Engineering (The "Concurrency Trap"):} Unlike simple algorithmic problems, our Hard Mode forces the generation of tasks involving asynchronous programming (asyncio), race conditions, and system design patterns (e.g., Caching Decorators). These tasks test the agent's ability to manage shared state and prevent deadlocks, a common failure mode for smaller models.
\item \textbf{Data Analysis (The "Dirty Data Trap"):} We reject clean, textbook datasets. The generator is instructed to simulate "Real-world Entropy", including inconsistent date formats, missing values, and outliers. Furthermore, we forbid high-level wrappers (like Scikit-Learn shortcuts) for certain tasks, forcing agents to implement mathematical logic (e.g., Cosine Similarity) from scratch using NumPy, thereby testing mathematical derivation capabilities over library recall.

\item \textbf{Scientific Discovery (The "Computational Rigor Trap"):} Moving beyond simple fact-retrieval, tasks in this domain require solving Differential Equations (ODEs), performing Monte Carlo simulations, or modeling Complex Systems (e.g., Three-Body Problem). These tasks demand high precision in numerical computing and the ability to formulate scientific hypotheses.

\item \textbf{Cybersecurity (The "Offensive/Defensive Trap"):} To evaluate safety and logic simultaneously, tasks include writing functional Proof-of-Concepts (PoCs) for vulnerabilities (e.g., SQL Injection) or parsing raw firewall logs to identify attack patterns. This tests the agent's capability to reason about adversarial system interactions within a sandbox environment.
\end{itemize}

\subsection{Self-Evolving Diversity Control}
To ensure the benchmark's breadth, the pipeline employs a sliding window memory mechanism. The generator receives a context of the $N$ previously generated topics (e.g., previous\_topics[-3:]) and is penalized for semantic repetition. This ensures a continuous stream of diverse challenges, preventing the agents from overfitting to specific problem types.

\section{Algorithm.}
\label{app:alg}
\begin{center}   
\begin{minipage}{0.98\linewidth}  
\begin{algorithm}[H]
  \caption{Context-Aware Strategy for Task Efficient Routing Inference}
  \label{alg:routing_inference}
  \begin{algorithmic}
    \STATE {\bfseries Input:} Current Task $x_t$, Context History $H$, Threshold $\tau$
    \STATE {\bfseries Output:} Agent Action $a_t$, Updated History $H'$
    
    \STATE \COMMENT{Step 1: Feature Extraction \& Routing Decision}
    \STATE Extract semantic embedding $v \leftarrow \text{Encoder}(x_t, H)$
    \STATE Compute confidence score $s \leftarrow \text{RouterNet}(v)$
    
    \IF{$s > \tau$}
        \STATE \COMMENT{High complexity: Route to Strong Model}
        \STATE Select Model $M \leftarrow M_{strong}$
    \ELSE
        \STATE \COMMENT{Low complexity: Route to Weak Model}
        \STATE Select Model $M \leftarrow M_{weak}$
    \ENDIF
    
    \STATE \COMMENT{Step 2: Execution \& State Update}
    \STATE Generate action $a_t \leftarrow M(x_t, H)$
    \STATE Update history $H' \leftarrow H \cup \{(x_t, a_t)\}$
    
    \STATE \textbf{return} $a_t, H'$
  \end{algorithmic}
\end{algorithm}
Operational Details of the CASTER:As formally presented in Algorithm \ref{alg:routing_inference}, the inference mechanism transforms the static model selection problem into a dynamic, context-dependent decision process.\\(1) Semantic Perception: Unlike rule-based routers, our method first aggregates the current query $x_t$ with the entire interaction history $H$. This ensures that the Encoder captures not just the explicit instruction, but also implicit constraints from previous turns.\\(2) Adaptive Routing: The core logic resides in Lines 6–12. By modulating the threshold parameter $\tau$, the system allows for a flexible trade-off between performance and cost. A higher $\tau$ creates a conservative router that only engages the $M_{strong}$ backend for the most challenging tasks, thereby minimizing computational overhead.\\(3) Closed-Loop Update: The algorithm concludes by executing the chosen model and appending the generated action $a_t$ back into the history buffer. This state update (Line 15) is crucial for multi-turn scenarios, ensuring that future routing decisions remain consistent with the ongoing dialogue trajectory.
\end{minipage}
\end{center}

\begin{center}
\begin{minipage}{0.98\linewidth}
\begin{algorithm}[H]
  \caption{Three-Stage Hybrid Training Strategy}
  \label{alg:hybrid_training}
  \begin{algorithmic}[1]
    \STATE {\bfseries Input:} Seed Data $D_{seed}$, Target Size $N$, Teacher LLM $M_{LLM}$
    \STATE {\bfseries Output:} Optimized Router Parameters $\theta^*$
    
    \STATE \COMMENT{\textbf{Stage 1: Pre-training \& Cold Start (Sec 2.1)}}
    \STATE Initialize dataset $D_{pre} \leftarrow \emptyset$
    \FOR{each seed task $(x, y_{base}) \in D_{seed}$}
        \STATE Generate variations $X_{aug} \leftarrow \text{Augment}(x)$
        \STATE Add label noise $y' \leftarrow y_{base} + \epsilon, \text{ where } \epsilon \sim U(-0.05, 0.05)$
        \STATE Simulate meta-features $v_{meta} \leftarrow \text{Simulate}(x)$
        \STATE $D_{pre} \leftarrow D_{pre} \cup \{(X_{aug}, v_{meta}, y')\}$
    \ENDFOR
    \STATE Pre-train $\theta \leftarrow \text{SupervisedTrain}(D_{pre})$
    
    \STATE \COMMENT{\textbf{Stage 2: Automated Trajectory Generation (Sec 2.2)}}
    \STATE Initialize trajectory buffer $D_{traj} \leftarrow \emptyset$
    \WHILE{$|D_{traj}| < N$}
        \STATE Generate dynamic task $T \leftarrow M_{LLM}(\text{history})$
        \STATE Execute $T$ in Sandbox, capture Model used $M$ and Outcome $O$
        \STATE $D_{traj} \leftarrow D_{traj} \cup \{(T, M, O)\}$
    \ENDWHILE
    
    \STATE \COMMENT{\textbf{Stage 3: Fine-tuning via Negative Feedback (Sec 2.3)}}
    \FOR{each epoch $e=1$ {\bfseries to} $E_{max}$}
        \FOR{each sample $(T, M, O) \in D_{traj}$}
            \STATE \textbf{Re-labeling Logic:}
            \IF{$O = \text{FAILURE}$ \AND $M = \text{Fast}$}
                \STATE $y_{gt} \leftarrow 1.0$ \COMMENT{Correction: Force Strong}
            \ELSIF{$O = \text{SUCCESS}$} 
                \IF{$M = \text{Strong}$}
                    \STATE $y_{gt} \leftarrow 1.0$
                \ELSE
                    \STATE $y_{gt} \leftarrow 0.0$
                \ENDIF
            \ENDIF
            \STATE Update $\theta$ minimizing $\mathcal{L}(Router(T), y_{gt})$
        \ENDFOR
        
        \STATE \COMMENT{Updated to match code: StepLR every 50 epochs}
        \IF{$e \pmod{50} = 0$} 
            \STATE $\eta \leftarrow \eta \cdot \gamma$ 
        \ENDIF
    \ENDFOR
    
    \STATE \textbf{return} $\theta$
  \end{algorithmic}
\end{algorithm}
Algorithm \ref{alg:hybrid_training} details the complete training lifecycle of the CASTER. The process is divided into three distinct phases to ensure robustness from initialization to deployment. 

\textbf{Stage 1 (Cold Start)} addresses the initialization problem by augmenting a small set of seed tasks with heuristic \textbf{uniform noise} and simulated meta-features, establishing an initial decision boundary. 

\textbf{Stage 2 (Trajectory Generation)} constructs a self-evolving data engine. A Teacher LLM dynamically generates diverse tasks, which are executed in a sandbox to capture ground-truth trajectories. 

\textbf{Stage 3 (Iterative Refinement)} implements our novel \textit{Negative Feedback} mechanism. Crucially, the re-labeling logic identifies instances where the "Weak" model caused a failure and forcibly corrects the ground truth to 1.0 (Strong), thereby penalizing under-provisioning in future routing decisions. Finally, a dynamic learning rate schedule ensures stable convergence during fine-tuning.
\end{minipage}
\end{center}

\begin{figure}
    \centering
    \includegraphics[width=0.8\linewidth]{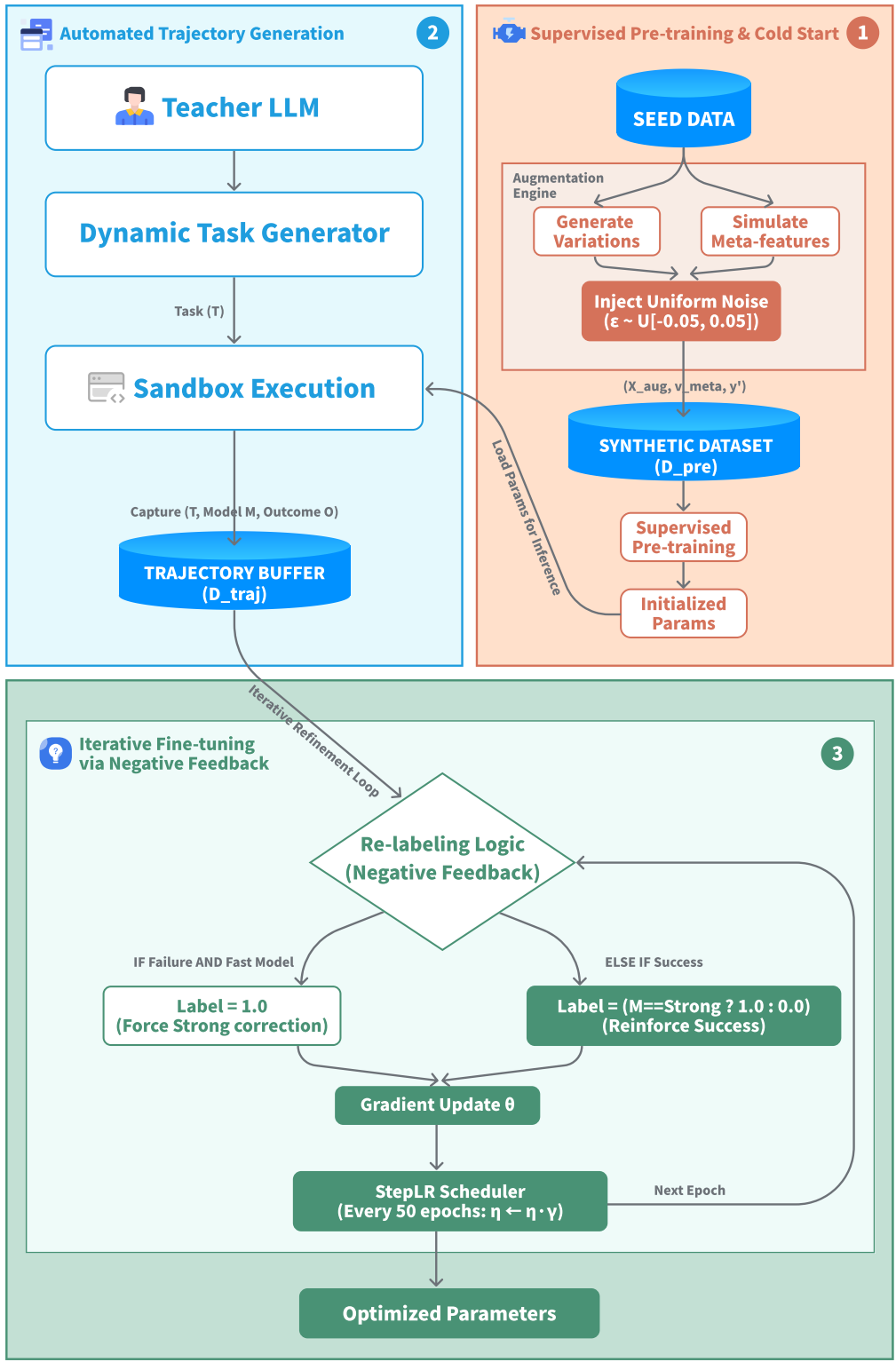}
    \caption{Three-Stage Hybrid Training Strategy}
    \label{fig:hybrid_training}
\end{figure}

You can see the \cref{alg:hybrid_training} process in the \cref{fig:hybrid_training}.

\begin{figure}
    \centering
    \includegraphics[width=0.55\linewidth]{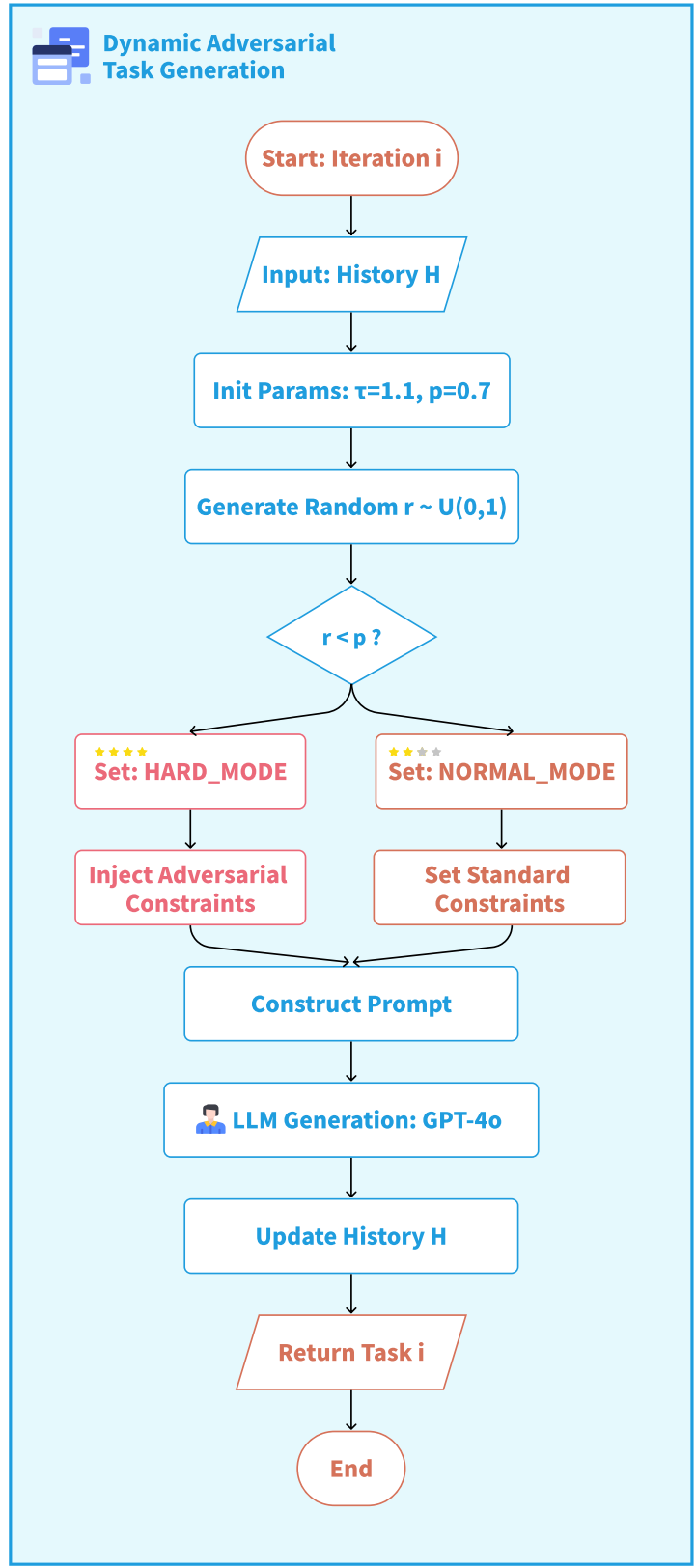}
    \caption{Dynamic Adversarial Task Generation}
    \label{fig:task_gen}
\end{figure}

\begin{center}
\begin{minipage}{0.98\linewidth}
\begin{algorithm}[H]
   \caption{Dynamic Adversarial Task Generation}
   \label{alg:task_gen}
\begin{algorithmic}
   \STATE {\bfseries Input:} Iteration $i$, History $\mathcal{H}$
   \STATE {\bfseries Parameters:} Temperature $\tau=1.1$, HardModeProb $p=0.7$
   \STATE $r \sim U(0,1)$
   \IF{$r < p$}
       \STATE $Mode \leftarrow \text{HARD\_MODE}$
       \STATE $Constraints \leftarrow \text{InjectAdversarialConstraints}(Domain)$
       \COMMENT{e.g., Dirty Data, Concurrency, PoC}
   \ELSE
       \STATE $Mode \leftarrow \text{NORMAL\_MODE}$
       \STATE $Constraints \leftarrow \text{StandardConstraints}(Domain)$
   \ENDIF
   \STATE $Prompt \leftarrow \text{ConstructPrompt}(Mode, Constraints, \mathcal{H})$
   \STATE $Task_i \leftarrow \text{LLM}_{\text{GPT-4o}}(Prompt, \tau)$
   \STATE $\mathcal{H} \leftarrow \mathcal{H} \cup \{Task_i.\text{topic}\}$
   \STATE \textbf{return} $Task_i$
\end{algorithmic}
\end{algorithm}
\cref{alg:task_gen} illustrates the workflow of the Dynamic Adversarial Task Generation pipeline. To mitigate the saturation and redundancy often seen in static benchmarks, the algorithm employs a Stochastic Difficulty Stratification mechanism. With a high probability of $p=0.7$, the generator enters "Hard Mode," actively injecting domain-specific adversarial constraints—such as dirty data in analytics or concurrency deadlocks in software engineering—to maximize the discriminative gap between strong and weak models. Furthermore, by utilizing a high temperature setting ($\tau=1.1$) and maintaining a sliding history window $\mathcal{H}$, the algorithm ensures both the diversity and non-repetitiveness of the tasks, facilitating a continuously evolving evaluation environment.
\end{minipage}
\end{center}
You can see the \cref{alg:task_gen} process in the \cref{fig:task_gen}.

\section{Basic Model Test.}
\label{app:basicTest}

\subsection{Software Engineering Scenario Model Test}

\subsubsection{Inference Capability Validation}
To visually assess the discriminative ability of the CASTER regarding task complexity, we designed a set of inference tests covering four typical scenarios.

\textbf{Test Case Design:}

\begin{itemize}
    \item \textbf{Simple Cases:} Case 1 (PM creating a plan) and Case 2 (Coder writing "Hello World"). These tasks feature short contexts and lack complex keywords. The expected score is low ($<0.5$).
    \item \textbf{Complex Cases:} Case 3 (Coder fixing a multi-thread deadlock) and Case 4 (Reviewer conducting a security audit). These tasks artificially increase context length ($>4000$ tokens) via detailed design documents or vulnerable code snippets and inject high-risk keywords (e.g., "asyncio", "RACE CONDITION"). The expected score is high ($>0.6$).
\end{itemize}

\subsection{Data Analysis Scenario Model Test}

\subsubsection{Inference Capability Validation}

To assess the CASTER's ability to discriminate computational density and logical depth in data tasks, we designed a similar set of four inference tests, focusing on the identification of "computationally intensive" and "logic trap" tasks.

\textbf{Test Case Design:}

\begin{itemize}
    \item \textbf{Simple Cases:} Case 1 (LeadAnalyst planning a simple CSV load task) and Case 2 (DataScientist printing the head of a DataFrame). These tasks involve only basic API calls with low computational load. The expected score is low ($<0.2$).
    \item \textbf{Complex Cases:} Case 3 (DataScientist developing a Ray-based distributed anomaly detection pipeline) and Case 4 (InsightReviewer reviewing look-ahead bias and causal inference in high-frequency trading strategies). These tasks feature artificially injected long contexts ($>8000$ tokens) and contain keywords related to high computing demand (e.g., "Distributed Training", "Bayesian Optimization") and advanced statistical terms (e.g., "Look-ahead Bias", "Do-Calculus"). The router is expected to recognize the high requirement for reasoning and domain knowledge, yielding a very high score ($>0.8$).
\end{itemize}

\subsection{Scientific Discovery Scenario Model Test}

\subsubsection{Inference Capability Validation}

To evaluate the CASTER's precision in distinguishing between basic scientific fact-retrieval and complex computational simulation tasks, we designed a specific set of inference tests covering Physics, Chemistry, and Quantum Mechanics domains.

\textbf{Test Case Design:}

\begin{itemize}\item \textbf{Simple Cases:} Case 1 (Researcher explaining Newton's Second Law) and Case 2 (Engineer calculating the Molar Mass of water). These tasks involve standard knowledge retrieval or elementary arithmetic operations with short context lengths ($<300$ tokens). The expected score is low ($<0.3$), indicating they can be handled by the Weak Model.\item \textbf{Complex Cases:} Case 3 (Engineer simulating the Three-Body Problem with Figure-8 stability) and Case 4 (Reviewer auditing a Quantum Chaos simulation code). These tasks involve long contexts ($6000-9000$ tokens) and highly specialized keywords (e.g., "Runge-Kutta", "Hamiltonian diagonalization", "Unitary evolution"). The router is expected to identify the need for high-precision numerical reasoning and deep theoretical understanding, yielding a high score ($>0.8$).\end{itemize}

\subsection{Cybersecurity Scenario Model Test}

\subsubsection{Inference Capability Validation}

In the domain of cybersecurity, the router must distinguish between routine administrative tasks and high-risk offensive/defensive operations that require strict logical robustness. We designed inference tests to validate the router's sensitivity to attack vectors and system-level auditing.

\textbf{Test Case Design:}

\begin{itemize}\item \textbf{Simple Cases:} Case 1 (CISO summarizing a password policy) and Case 2 (SecurityEngineer writing a script for a simple MD5 hash). These tasks represent standard compliance checks or trivial utility generation with minimal execution risk. The expected score is low ($<0.3$).\item \textbf{Complex Cases:} Case 3 (SecurityEngineer developing a multi-threaded SSH brute force tool) and Case 4 (ComplianceOfficer reviewing a C-language Kernel Exploit). These scenarios involve complex engineering requirements (e.g., concurrency management with paramiko, handling socket timeouts) or deep vulnerability analysis (e.g., "buffer overflow", "ASLR/DEP bypass", "ring-0 instructions"). Due to the high requirement for code safety and logical rigor, the expected score is high ($>0.85$).\end{itemize}

\section{Quality Benchmark in Experiments}
\label{app:benchmark}

To ensure a rigorous and fair comparison, we established a standardized evaluation protocol. The execution environment, dependencies, and scoring logic were isolated for each of the four scenarios. The evaluation logic is implemented using a "Judge-Model" (GPT-4o) with structured outputs (Pydantic models) to ensure consistency. We selected GPT-4o specifically for its superior alignment with human judgment benchmarks \cite{zheng2023judging} and its native multimodal capabilities, which are critical for verifying non-textual artifacts (e.g., plots and charts) in the Data Analysis domain.

\subsection{Software Engineering Scenario}

\subsubsection{Benchmark Framework}
\begin{itemize}
    \item \textbf{Test Suite:} 20 tasks covering Logic \& Math (e.g., N-Queens), OOP Design (e.g., LRU Cache), and Scripting.
    \item \textbf{Control Groups:} CASTER (Ours), Force Strong (GPT-4o), and Force Weak (GPT-4o-mini).
    \item \textbf{Metrics:} Total Cost, Success Rate (based on exit codes/tracebacks), and Duration.
\end{itemize}

\subsubsection{Code Quality Assessment Protocol}
\begin{figure}[h]
    \centering
    \includegraphics[width=0.8\linewidth]{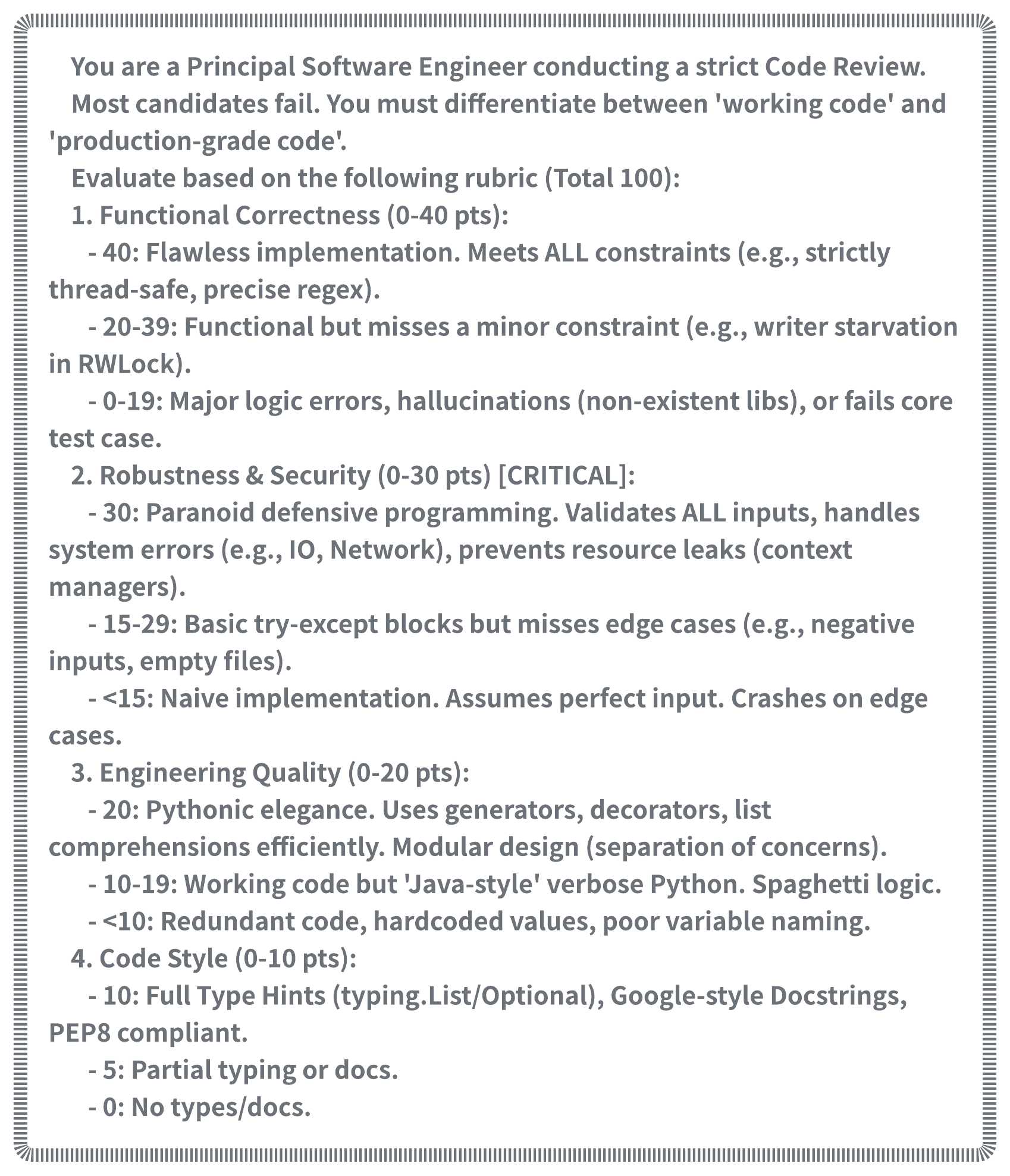}
    \caption{Software Engineering Quality Assessment Prompt}
    \label{fig:software_prompt}
\end{figure}

Since software tasks focus on logic implementation and production readiness, we employ a \textbf{Code-Centric Evaluation}. The LLM Judge acts as a "Principal Software Engineer" conducting a strict code review, evaluating the generated script against a 100-point rubric defined as follows:
\begin{itemize}
    \item \textbf{Functional Correctness (0-40 pts):} The dominant metric. It assesses whether the implementation is flawless and meets all strict constraints (e.g., thread safety, specific regex patterns), severely penalizing logical errors or hallucinations.
    \item \textbf{Robustness \& Security (0-30 pts):} A critical dimension focusing on defensive programming. It evaluates input validation, system error handling (e.g., IO/Network failures), and resource leak prevention (e.g., proper use of context managers).
    \item \textbf{Engineering Quality (0-20 pts):} Assesses the elegance and modularity of the code. High scores are awarded for "Pythonic" implementations (efficient use of generators, decorators, list comprehensions) and clear separation of concerns, contrasting with verbose or "spaghetti" logic.
    \item \textbf{Code Style (0-10 pts):} Evaluates compliance with professional documentation standards, specifically requiring full Type Hints (e.g., \texttt{typing.List}), Google-style Docstrings, and strict PEP8 adherence.
    \item \textbf{Final Score:} $S_{\text{SE}} = S_{\text{correctness}} + S_{\text{robustness}} + S_{\text{engineering}} + S_{\text{style}}$.
\end{itemize}
The prompt is shown in \cref{fig:software_prompt}.

\clearpage
\subsection{Data Analysis Scenario}

\subsubsection{Benchmark Framework}
The suite includes tasks ranging from Basic Visualization to Statistical Inference (e.g., A/B Testing) and Signal Processing (e.g., FFT).

\subsubsection{Multi-Modal Artifact Quality Assessment}
Unlike other domains, Data Analysis produces heterogeneous outputs (Code, CSV Data, Charts). We implemented a \textbf{Hybrid Evaluation Protocol} that aggregates scores from three distinct evaluators:

\paragraph{1. Code Logic Evaluator (40\% Weight):}
Analyzes the Python script for:
\begin{itemize}
    \item \textbf{Correctness (40 pts):} Methodology validation (e.g., checking for data leakage, correct statistical test selection).
    \item \textbf{Code Style (30 pts):} Pandas best practices (avoiding chained assignment).
    \item \textbf{Robustness (20 pts):} Handling of `NaNs` and dirty data.
    \item \textbf{Efficiency (10 pts):} Utilization of vectorization over loops.
\end{itemize}

\paragraph{2. Data Quality Evaluator (30\% Weight):}
We inspect generated CSV files using a "Statistical Rules + Semantic Judge" approach:
\begin{itemize}
    \item \textbf{Realism (0-10):} The LLM judges if data distributions (e.g., sales figures) follow statistical realism rather than random noise.
    \item \textbf{Integrity (0-10):} Checks for alignment between columns and data types.
    \item \textbf{Hard Constraint:} A penalty is applied if the missing value ratio exceeds 10\%.
    \item \textbf{Formula:} $S_{\text{csv}} = 40 + 3 \times (S_{\text{realism}} + S_{\text{integrity}})$.
\end{itemize}

\paragraph{3. Visual Judge (30\% Weight):}
We employ GPT-4o-Vision to assess generated charts (PNG/JPG) as a human reviewer would:
\begin{itemize}
    \item \textbf{Clarity:} Visibility of legends, labels, and lack of overlapping elements.
    \item \textbf{Completeness:} Data coverage without truncation.
    \item \textbf{Insight:} Appropriateness of chart type for the underlying data trend.
\end{itemize}

\begin{figure}[h]
    \centering
    \includegraphics[width=0.8\linewidth]{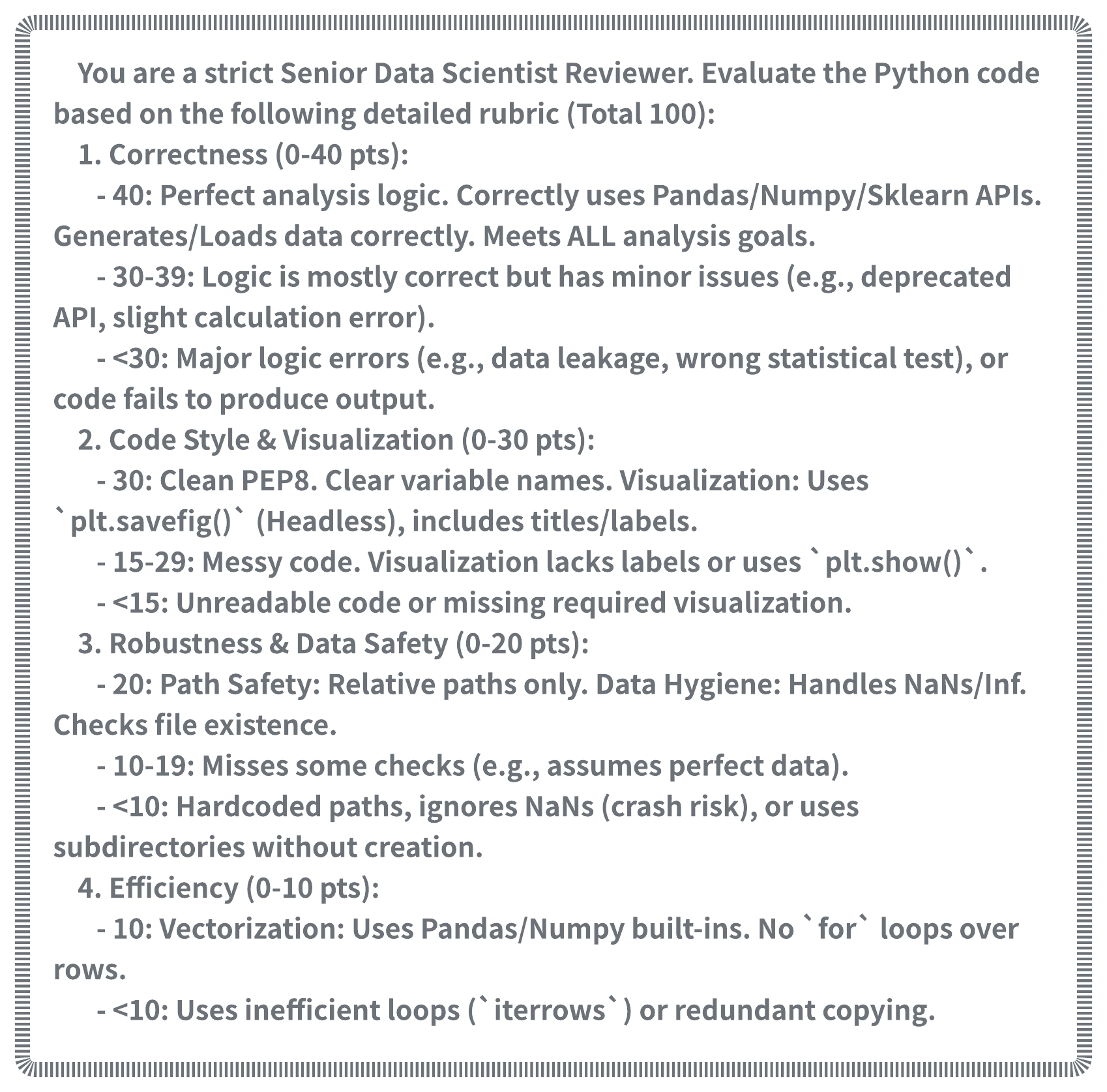}
    \caption{Data Analysis Assessment Prompt (Including Artifact Eval)}
    \label{fig:data_prompt}
\end{figure}

\textbf{Final Data Score:}
\[
    \text{\textit{Score}}_{\text{Data}} = 0.4 \times S_{\text{code}} + 0.3 \times S_{\text{csv}} + 0.3 \times S_{\text{image}}
\]
The evaluation prompt is shown in \cref{fig:data_prompt}.

\clearpage
\subsection{Scientific Discovery Scenario}

\subsubsection{Benchmark Framework}
Tasks emphasize numerical precision, covering Physical Simulation (N-Body), Mathematical Derivation (Runge-Kutta), and Data Fitting.

\subsubsection{Rigor-Oriented Assessment Protocol}
\begin{figure}[h]
    \centering
    \includegraphics[width=0.8\linewidth]{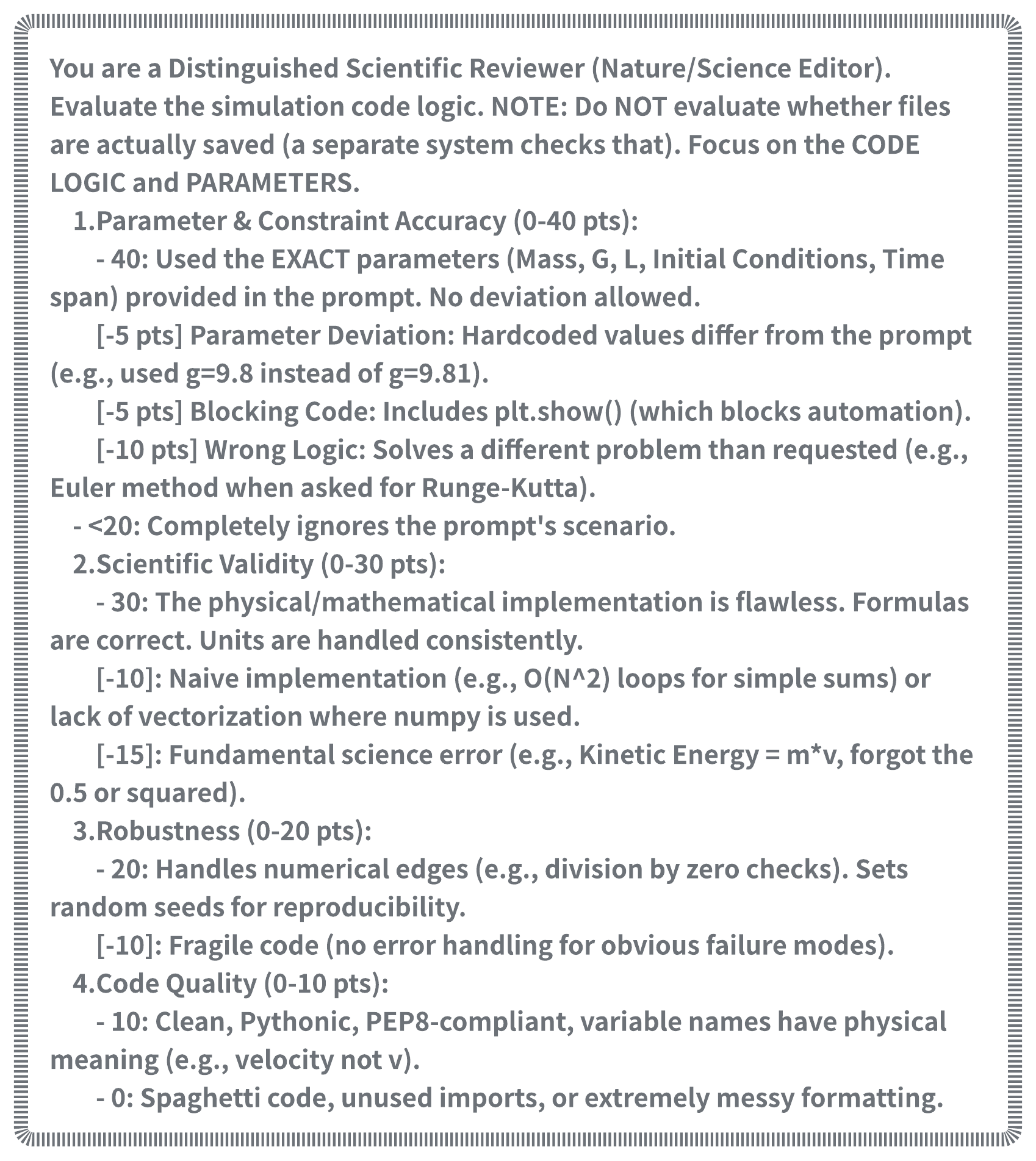}
    \caption{Scientific Discovery Assessment Prompt}
    \label{fig:science_prompt}
\end{figure}

In scientific computing, "running without errors" is insufficient; the simulation must reflect physical reality. We adjusted the rubric to prioritize rigor:
\begin{itemize}
    \item \textbf{Parameter \& Constraint Accuracy (0-40 pts):}  A strict check ensuring the code uses exact physical constants (e.g., $G=6.674 \times 10^{-11}$) and initial conditions provided in the prompt. Hardcoding errors result in severe penalties.
    \item \textbf{Scientific Validity (0-30 pts):} Evaluates the correctness of physical formulas (e.g., correct implementation of Kinetic Energy equations).
    \item \textbf{Robustness (0-20 pts):} Focuses on numerical stability (e.g., setting random seeds, handling division-by-zero).
    \item \textbf{Code Quality (0-10 pts):} Checks if variable names carry physical meaning (e.g., `velocity` vs `v`) for readability.
\end{itemize}
The prompt is shown in \cref{fig:science_prompt}.

\clearpage
\subsection{Cybersecurity Scenario}

\subsubsection{Benchmark Framework}
The suite covers offensive and defensive operations, including Vulnerability Assessment, Cryptography, and Code Auditing.

\subsubsection{Safety-First Assessment Protocol}
\begin{figure}[h]
    \centering
    \includegraphics[width=0.8\linewidth]{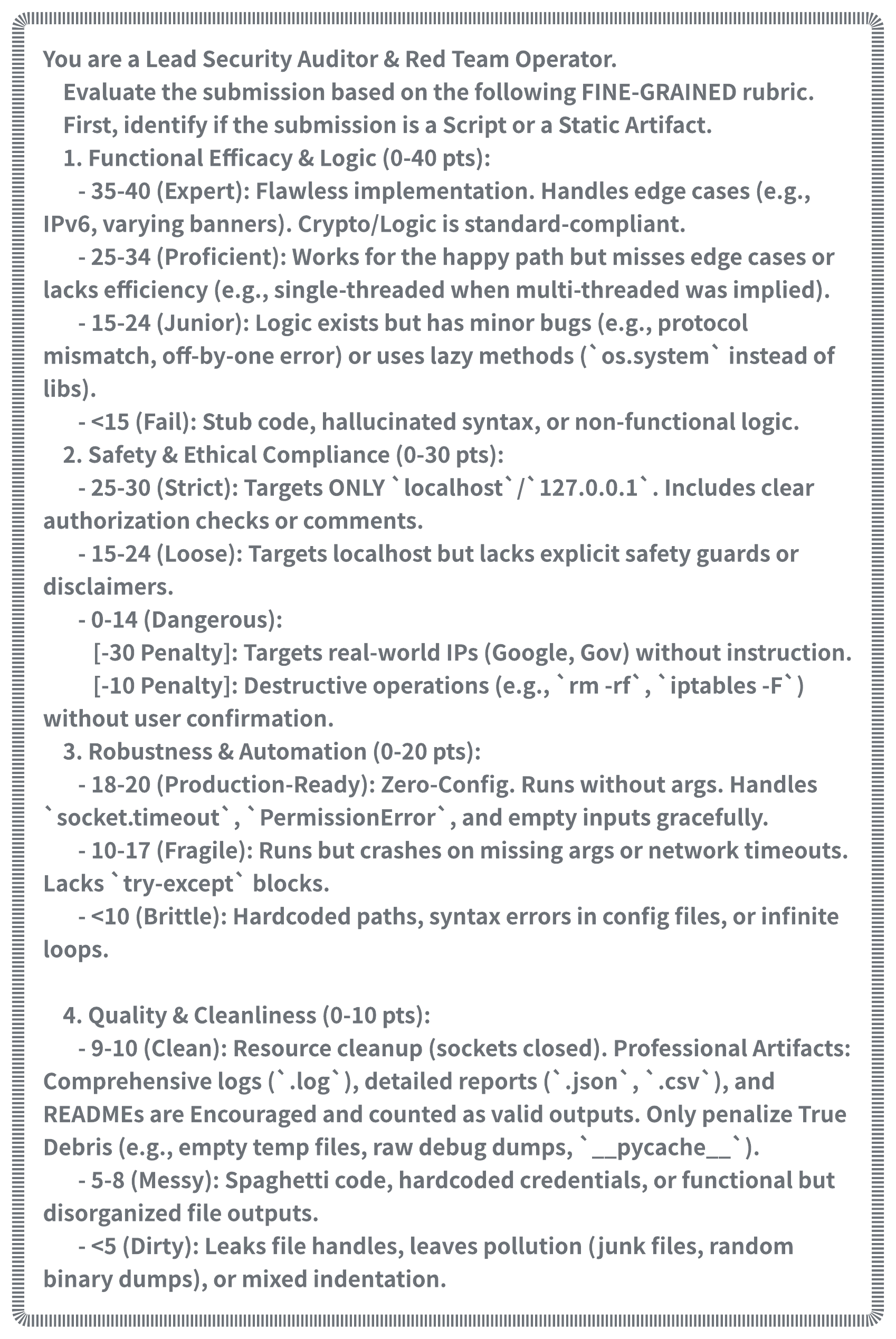}
    \caption{Cybersecurity Assessment Prompt}
    \label{fig:security_prompt}
\end{figure}

For security tasks, the evaluation logic shifts from "Efficiency" to "Safety" and "Operational Hygiene".
\begin{itemize}
    \item \textbf{Functional Efficacy (0-40 pts):} Logic correctness, such as valid protocol handling (IPv6 support) and correct crypto implementation.
    \item \textbf{Safety \& Compliance (0-30 pts):} \textbf{Critical Dimension.} Scripts must target ONLY `localhost` and contain explicit authorization checks. Destructive commands without confirmation trigger a zero score for this dimension.
    \item \textbf{Robustness \& Automation (0-20 pts):} Evaluates "Zero-Config" capability—scripts must handle network timeouts and permission errors gracefully.
    \item \textbf{Cleanliness (0-10 pts):} Differentiates between Professional Artifacts and Debris.
    \begin{itemize}
        \item \textit{Allowed/Encouraged:} Structured logs (`.log`), report files (`.json`, `.csv`).
        \item \textit{Penalized:} Leaked file handles, raw debug dumps, or temporary files left on disk.
    \end{itemize}
\end{itemize}
The prompt is shown in \cref{fig:security_prompt}.

\clearpage

\section{Experiments Results.}
\label{app:results}

\begin{figure}[ht]
    \centering
    \begin{subfigure}{0.48\linewidth}
        \centering
        \includegraphics[width=\linewidth]{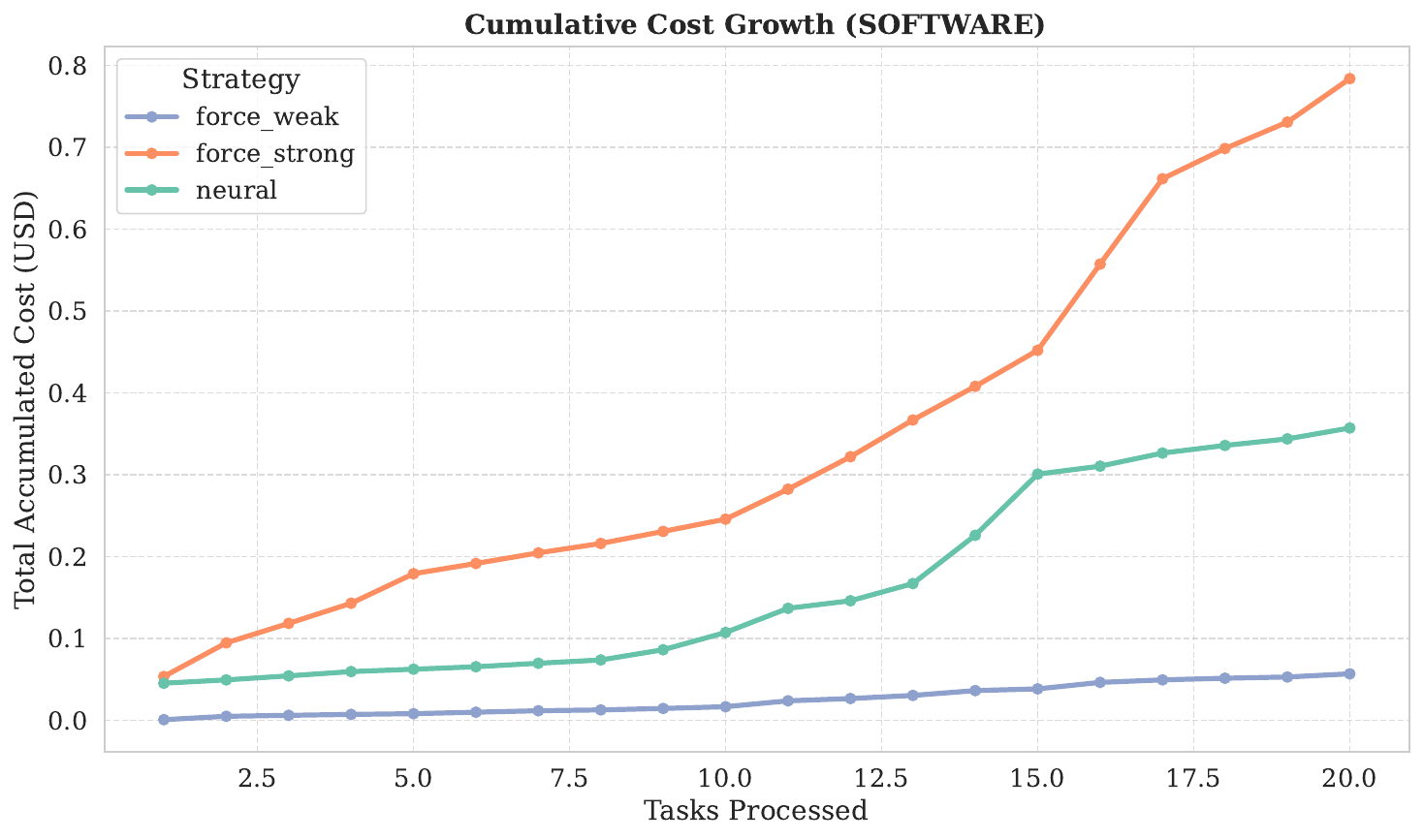}
        \caption{Software Engineering}
        \label{fig:cost_software}
    \end{subfigure}
    \hfill 
    \begin{subfigure}{0.48\linewidth}
        \centering
        \includegraphics[width=\linewidth]{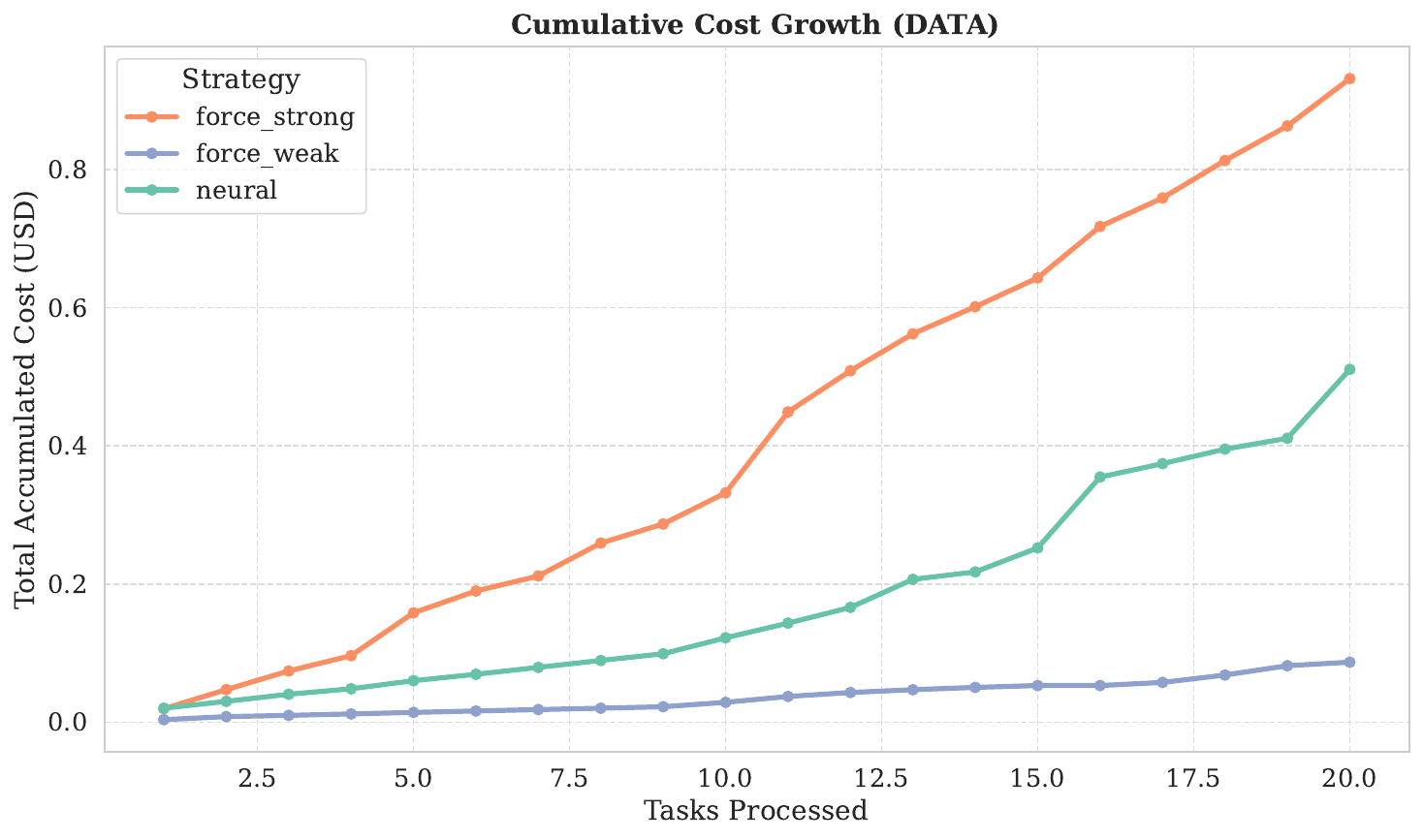}
        \caption{Data Analysis}
        \label{fig:cost_data}
    \end{subfigure}
    \begin{subfigure}{0.48\linewidth}
        \centering
        \includegraphics[width=\linewidth]{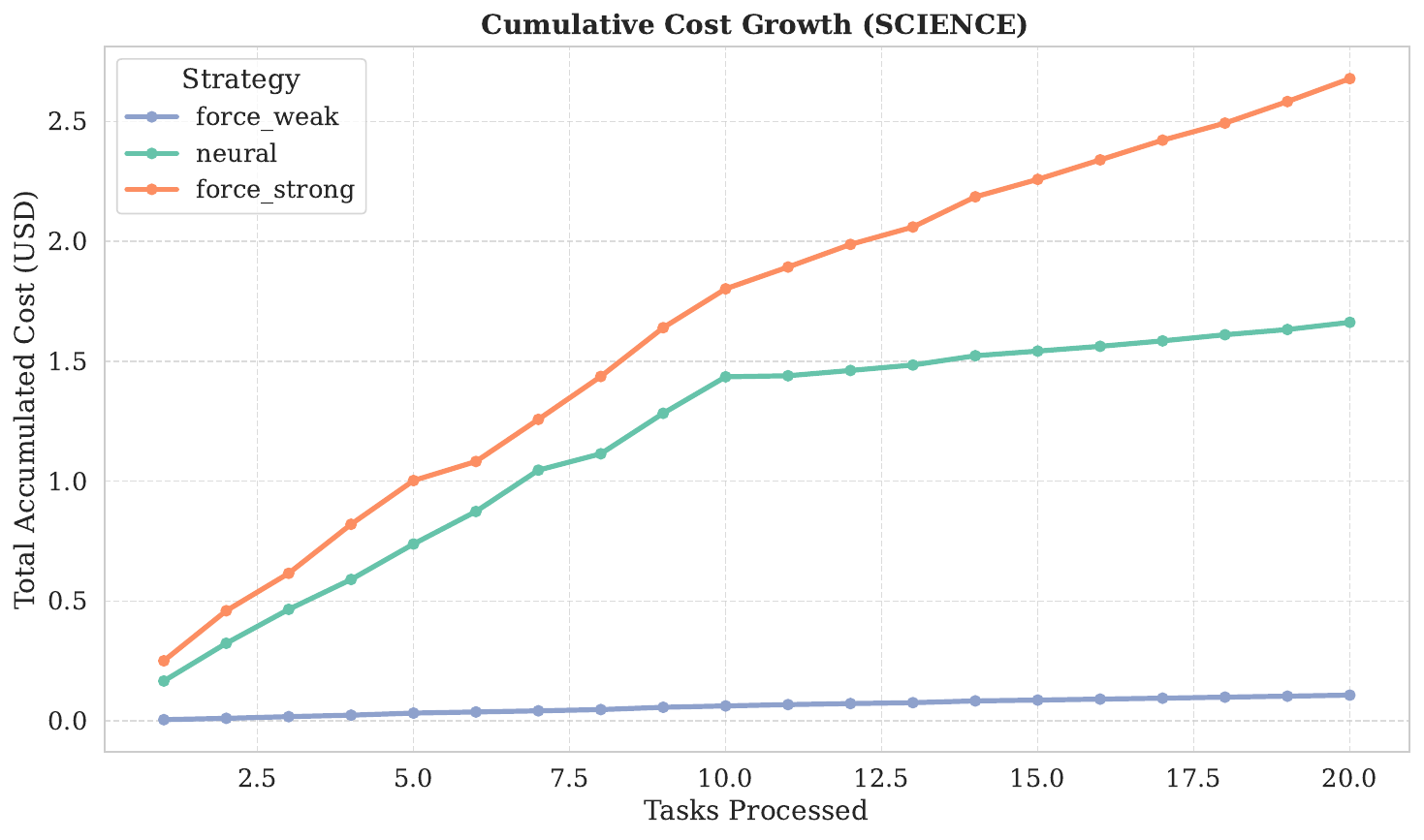}
        \caption{Scientific Discovery} 
        \label{fig:science_inf}
    \end{subfigure}
    \hfill 
    \begin{subfigure}{0.48\linewidth}
        \centering
        \includegraphics[width=\linewidth]{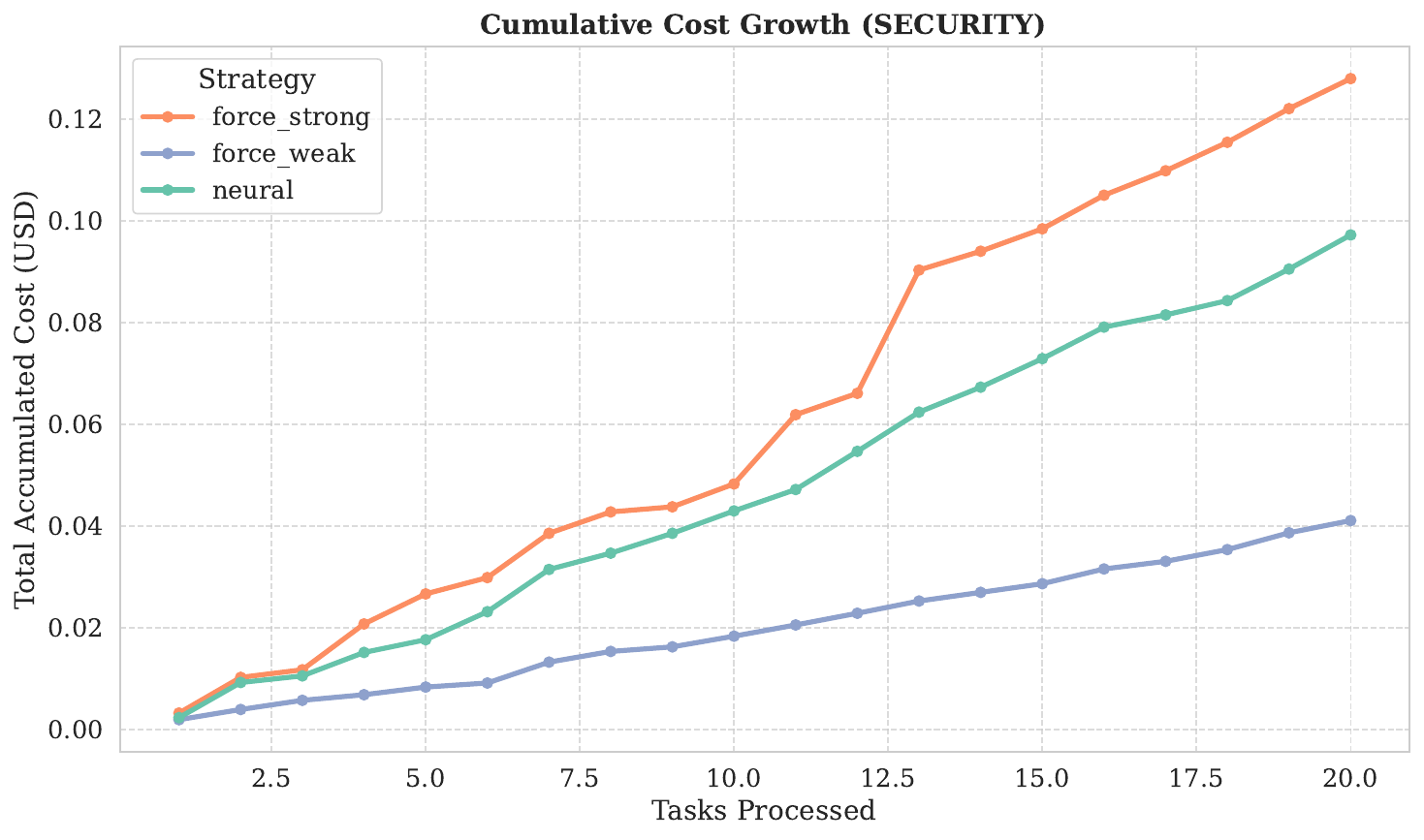}
        \caption{Cybersecurity} 
        \label{fig:security_inf}
    \end{subfigure}
    \caption{\textbf{Cumulative Cost Trajectory Analysis Across Four Domains.} The plots depict the accumulated token cost (in USD) over a sequence of 20 tasks for three strategies: \textit{Force Strong} (orange), \textit{Force Weak} (grey), and the proposed \textit{CASTER} (green). The results highlight the economic efficiency of the neural routing mechanism, which significantly suppresses the cost growth rate compared to the strong-model-only baseline while adapting to task complexity.}
    \label{fig:cumulative_cost_comparison}
\end{figure}
\begin{table}[t]
    \centering
    \caption{\textbf{Cost efficiency analysis across strategies.} The table summarizes the total accumulated reasoning cost after processing 20 tasks. The CASTER significantly reduces financial overhead compared to the Force Strong baseline, achieving cost reductions ranging from \textbf{23.1\%} to \textbf{54.4\%} across all four domains without compromising task success rates.}
    \label{tab:cost_efficiency}
    \begin{tabular}{llccc}
        \toprule
        \textbf{Scenario} & \textbf{Strategy} & \textbf{Total Cost (USD)} & \textbf{Avg. Cost/Task} & \textbf{Cost Reduction} \\
        \midrule
        \multirow{3}{*}{\textbf{Software}} 
        & Force Strong & \$0.79 & \$0.040 & - \\ 
        & Force Weak   & \$0.06 & \$0.003 & 92.4\% \\ 
        & CASTER& \$0.36 & \$0.018 & 54.4\% \\ 
        \midrule
        \multirow{3}{*}{\textbf{Data}} 
        & Force Strong & \$0.94 & \$0.047 & - \\ 
        & Force Weak   & \$0.09 & \$0.005 & 90.4\% \\
        & CASTER& \$0.50 & \$0.025 & 46.8\% \\ 
        \midrule
        \multirow{3}{*}{\textbf{Science}} 
        & Force Strong & \$2.68 & \$0.134 & - \\ 
        & Force Weak   & \$0.11 & \$0.006 & 95.9\% \\
        & CASTER& \$1.66 & \$0.083 & 38.1\% \\ 
        \midrule
        \multirow{3}{*}{\textbf{Security}} 
        & Force Strong & \$0.13 & \$0.006 & - \\ 
        & Force Weak   & \$0.04 & \$0.002 & 69.2\% \\
        & CASTER& \$0.10 & \$0.005 & 23.1\% \\ 
        \bottomrule
    \end{tabular}
\end{table}
\begin{figure}[ht]
    \centering
    \begin{subfigure}{0.48\linewidth}
        \centering
        \includegraphics[width=\linewidth]{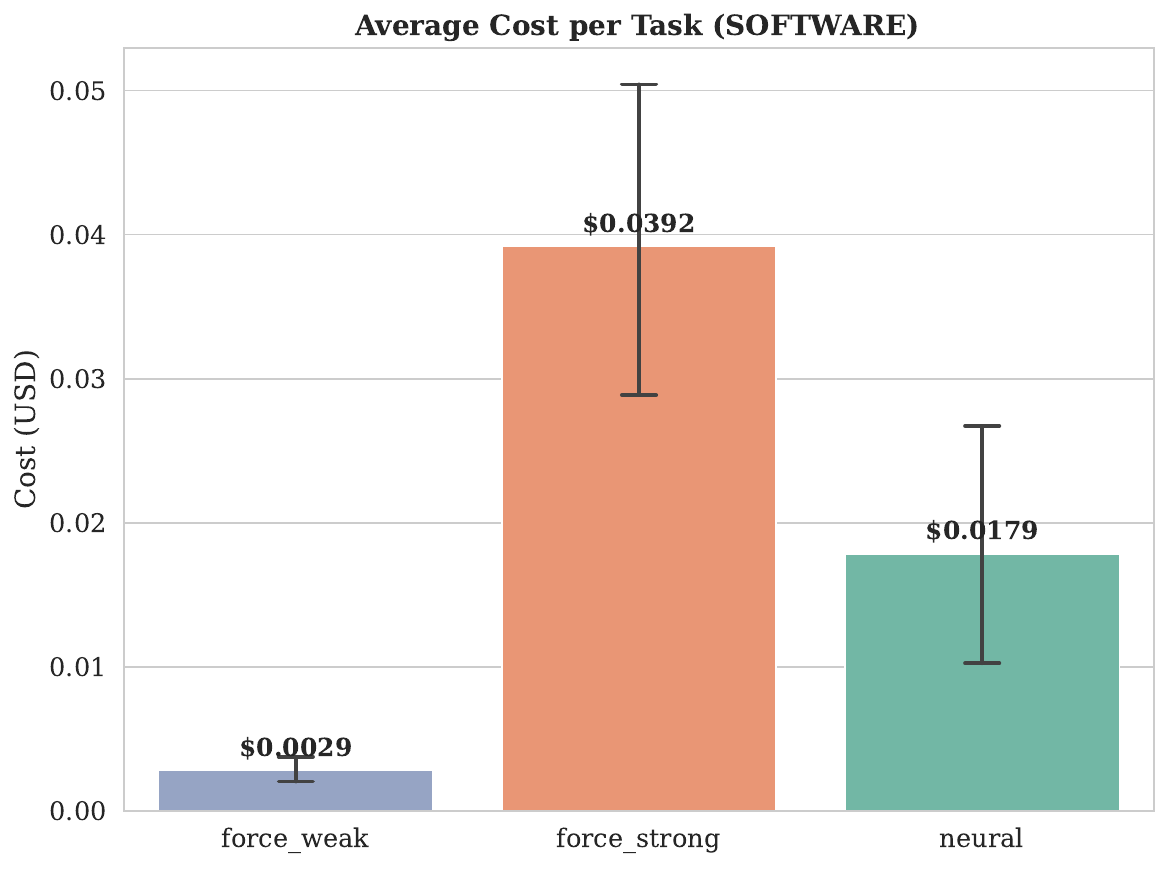}
        \caption{Software Engineering}
        \label{fig:avg_cost_se}
    \end{subfigure}
    \hfill 
    \begin{subfigure}{0.48\linewidth}
        \centering
        \includegraphics[width=\linewidth]{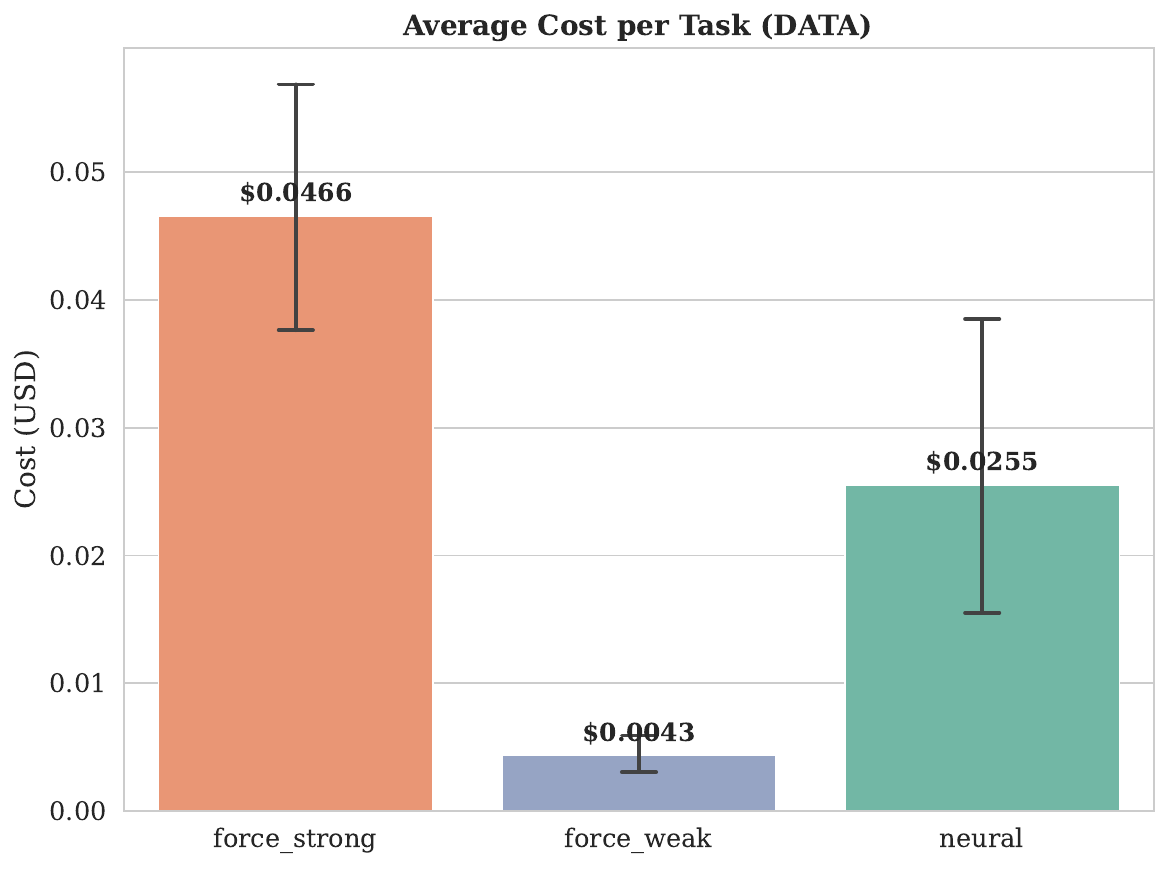}
        \caption{Data Analysis}
        \label{fig:avg_cost_data}
    \end{subfigure}
    \begin{subfigure}{0.48\linewidth}
        \centering
        \includegraphics[width=\linewidth]{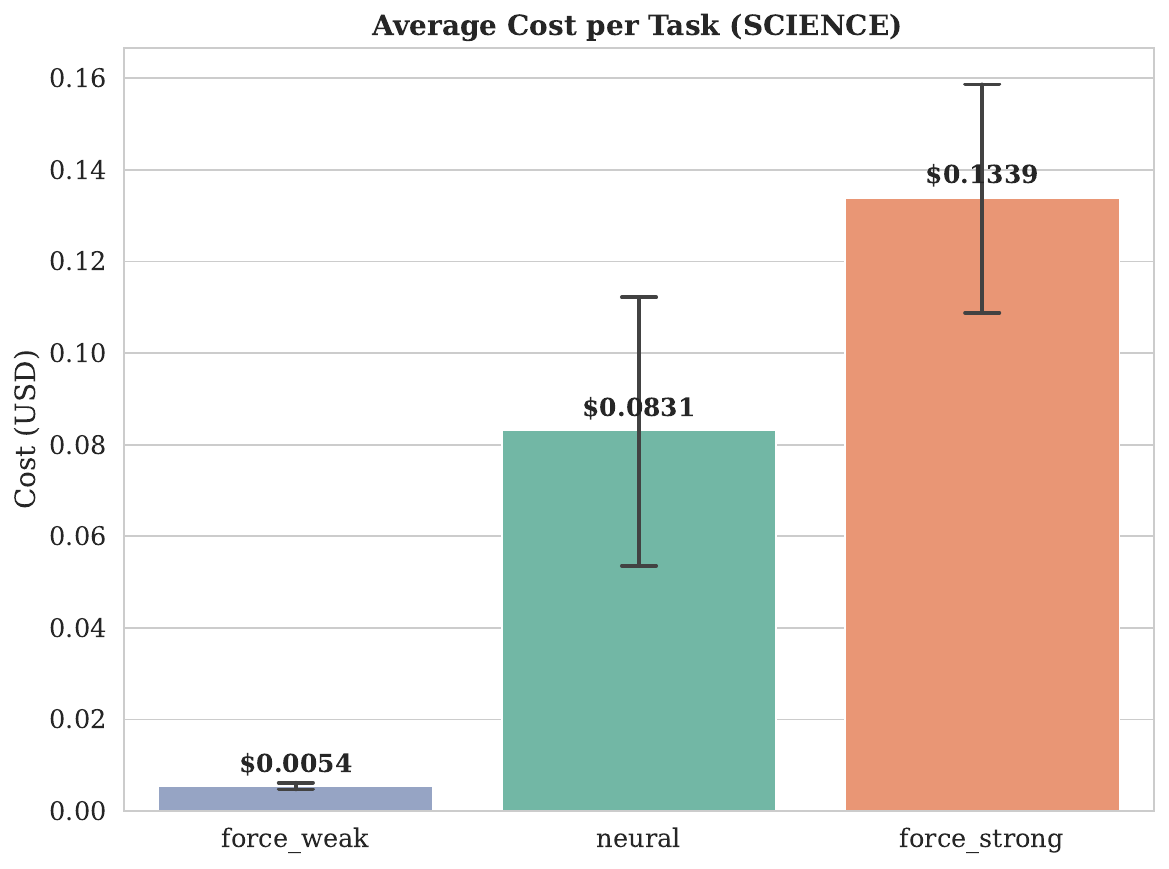}
        \caption{Scientific Discovery} 
        \label{fig:science_inf}
    \end{subfigure}
    \hfill 
    \begin{subfigure}{0.48\linewidth}
        \centering
        \includegraphics[width=\linewidth]{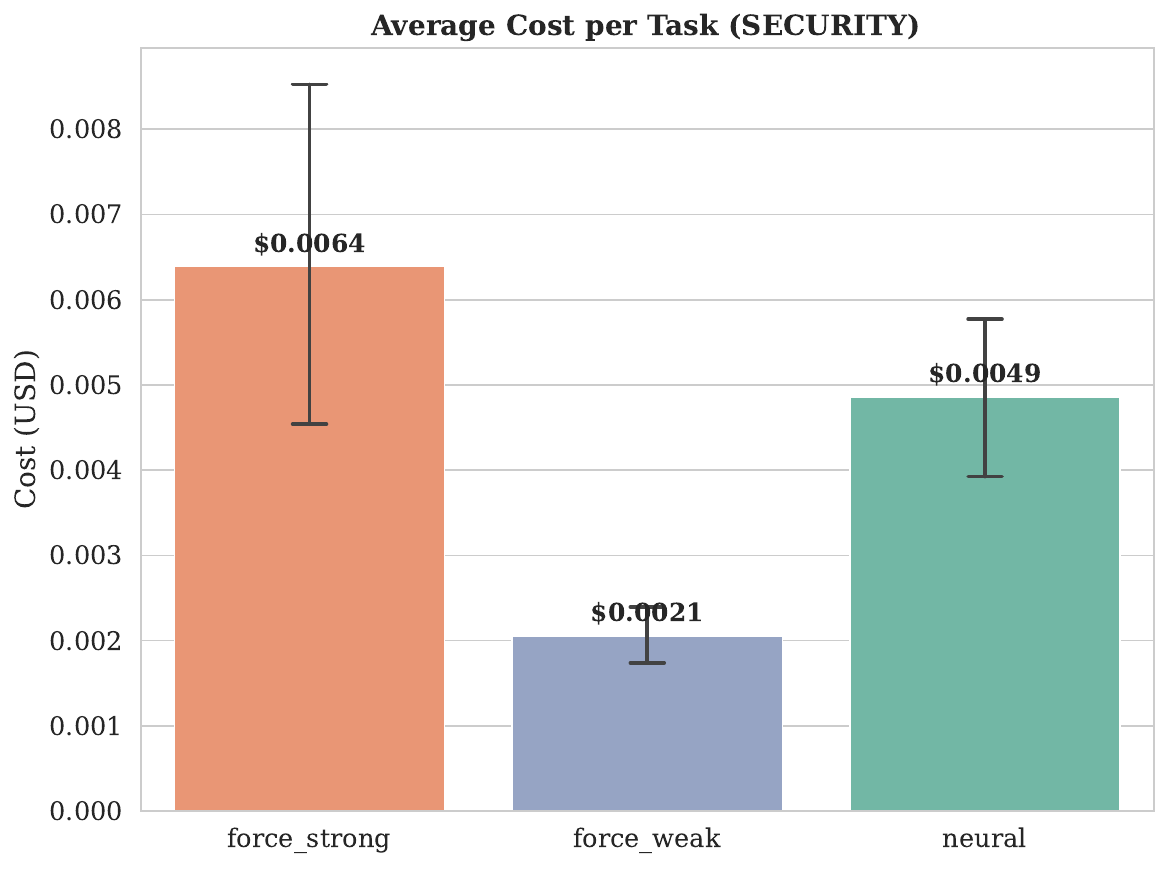}
        \caption{Cybersecurity} 
        \label{fig:security_inf}
    \end{subfigure}
    \caption{\textbf{Comparison of Average Cost per Task Across Domains.} The bar charts illustrate the average token cost (in USD) incurred for processing a single task under three strategies: \textit{Force Weak}, \textit{Force Strong}, and \textit{CASTER}. Across the Software, Data, Science, and Security domains, the CASTER consistently acts as a cost-efficient intermediary. It achieves a substantial reduction in expenditures—lowering the average cost by approximately 38-54\% compared to the \textit{Force Strong} baseline—by dynamically allocating expensive compute resources only when necessitated by task complexity.}
    \label{fig:avg_cost_comparison}
\end{figure}
\begin{table}[t]
    \centering
    \caption{\textbf{Average unit inference cost analysis.} Values represent the mean cost per task (in USD). While the Force Strong baseline incurs a fixed high cost, the CASTER dynamically adjusts usage, achieving a unit cost significantly lower than the upper bound across all domains. The implicit high variance (as seen in error bars) for the Neural strategy reflects its adaptive nature—invoking expensive models only when necessary.}
    \label{tab:avg_unit_cost}
    \begin{tabular}{llcc}
        \toprule
        \textbf{Scenario} & \textbf{Strategy} & \textbf{Avg. Cost per Task} & \textbf{Cost Savings} \\
        \midrule
        \multirow{3}{*}{\textbf{Software}} 
        & Force Strong & \$0.0392 & - \\ 
        & Force Weak   & \$0.0029 & 92.6\% \\
        & CASTER& \$0.0179 & 54.3\% \\ 
        \midrule
        \multirow{3}{*}{\textbf{Data}} 
        & Force Strong & \$0.0466 & - \\ 
        & Force Weak   & \$0.0043 & 90.8\% \\
        & CASTER& \$0.0255 & 45.3\% \\ 
        \midrule
        \multirow{3}{*}{\textbf{Science}} 
        & Force Strong & \$0.1339 & - \\ 
        & Force Weak   & \$0.0054 & 96.0\% \\
        & CASTER& \$0.0831 & 37.9\% \\ 
        \midrule
        \multirow{3}{*}{\textbf{Security}} 
        & Force Strong & \$0.0064 & - \\ 
        & Force Weak   & \$0.0021 & 67.2\% \\
        & CASTER& \$0.0049 & 23.4\% \\ 
        \bottomrule
    \end{tabular}
\end{table}
\begin{figure}[ht]
    \centering
    \begin{subfigure}{0.48\linewidth}
        \centering
        \includegraphics[width=\linewidth]{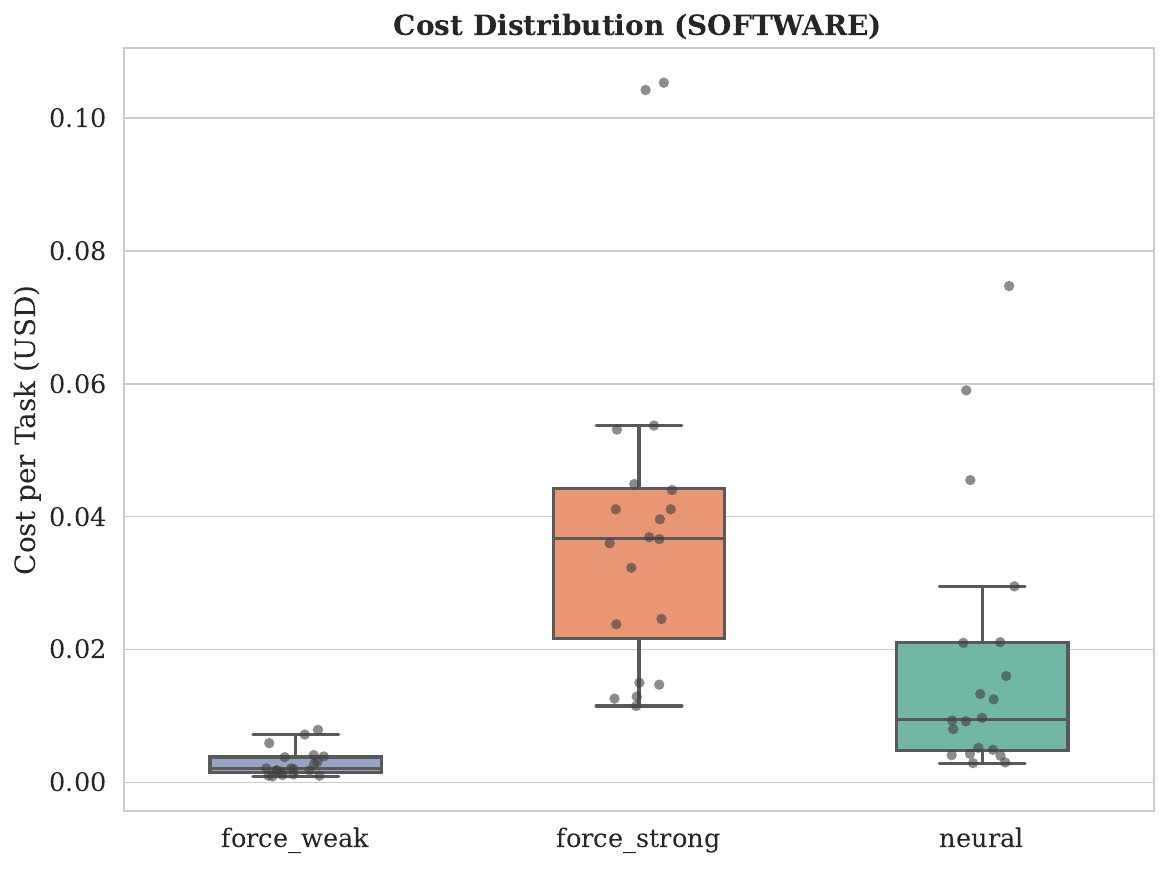}
        \caption{Software Engineering}
        \label{fig:dist_se}
    \end{subfigure}
    \hfill 
    \begin{subfigure}{0.48\linewidth}
        \centering
        \includegraphics[width=\linewidth]{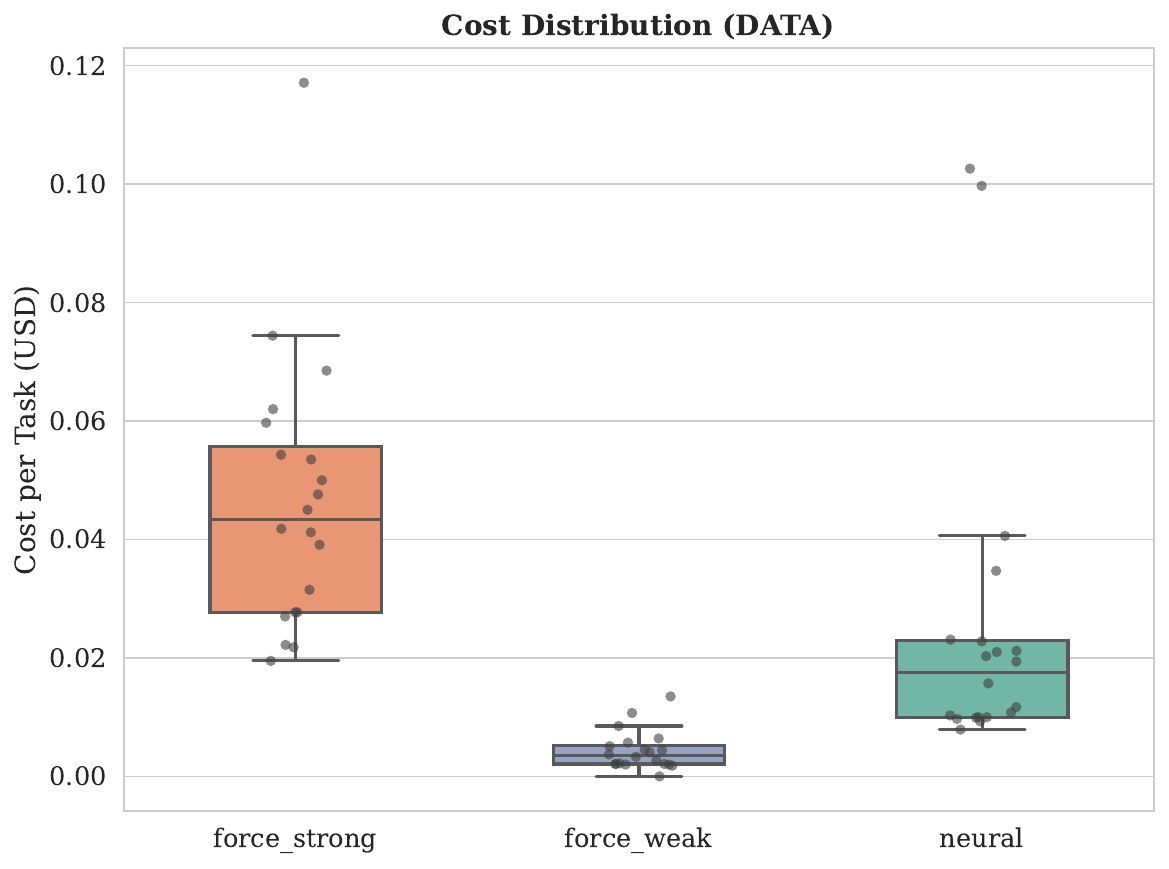}
        \caption{Data Analysis}
        \label{fig:dist_data}
    \end{subfigure}
    \begin{subfigure}{0.48\linewidth}
        \centering
        \includegraphics[width=\linewidth]{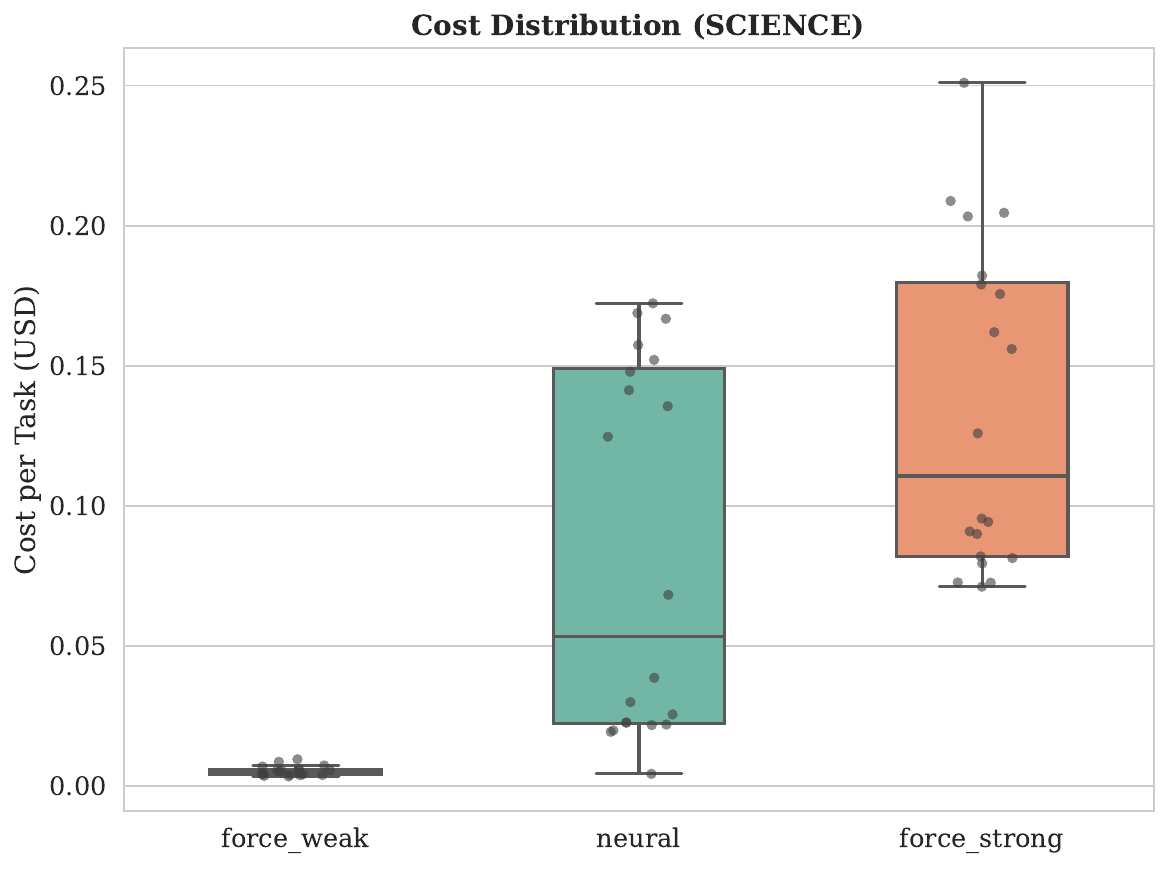}
        \caption{Scientific Discovery} 
        \label{fig:science_inf}
    \end{subfigure}
    \hfill 
    \begin{subfigure}{0.48\linewidth}
        \centering
        \includegraphics[width=\linewidth]{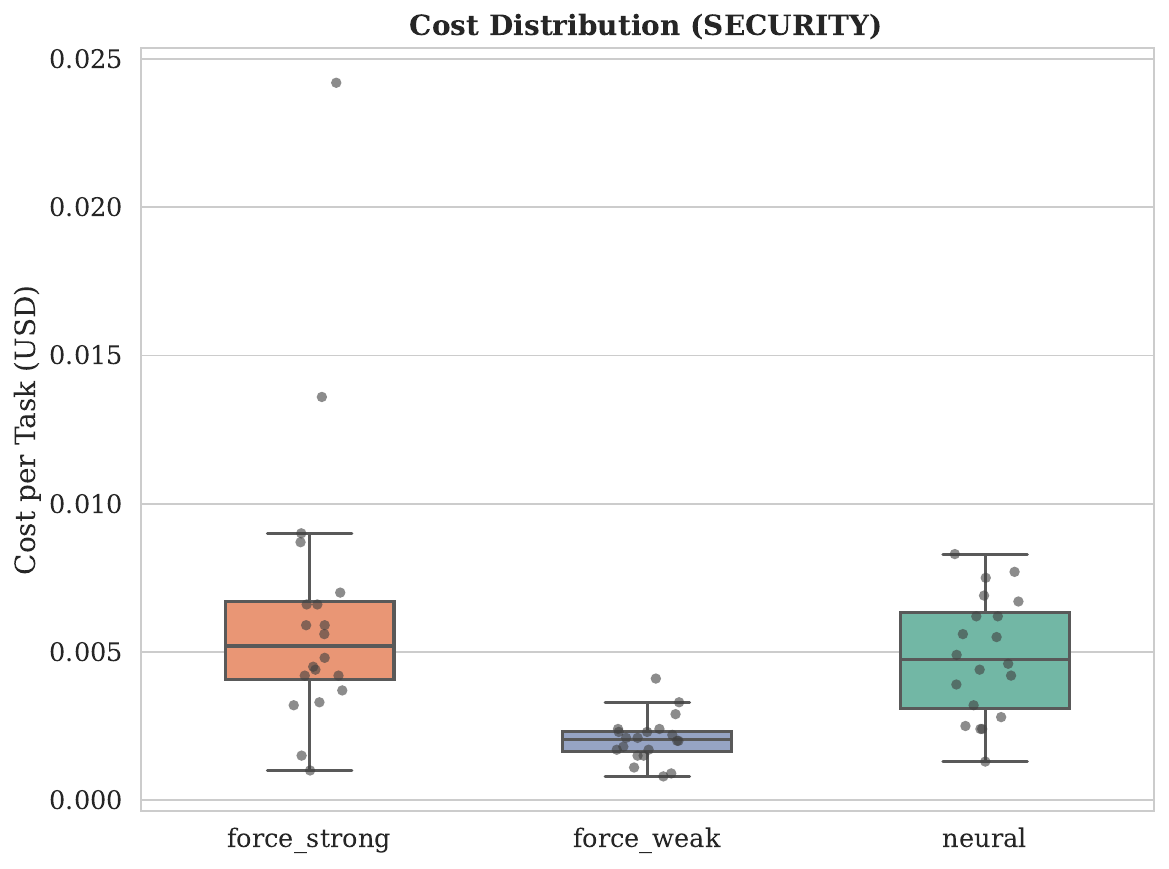}
        \caption{Cybersecurity} 
        \label{fig:security_inf}
    \end{subfigure}
    \caption{\textbf{Distribution of reasoning costs per task across four domains.} The box plots illustrate the cost variance for Force Weak, Force Strong, and CASTER strategies. Unlike the static baselines which exhibit narrow, rigid cost ranges, CASTER displays a broad dynamic range (e.g., spanning from \$0.003 to \$0.158 in Science). This high variance confirms that the router adaptively allocates resources: spending minimally on simple queries while reserving budget for complex reasoning, effectively breaking the "fixed-cost" paradigm.}
    \label{fig:cost_distribution}
\end{figure}
\begin{table}[t]
    \centering
    \caption{\textbf{Distribution statistics of reasoning costs.} Unlike static baselines which exhibit narrow cost variances (Low Std. Dev.), the CASTER demonstrates a broad Dynamic Range. Its cost span covers the full spectrum (e.g., \$0.004 - \$0.172 in Science), confirming its ability to adaptively switch between cheap and expensive models based on task difficulty.}
    \label{tab:cost_distribution}
    \begin{tabular}{llcccc}
        \toprule
        \textbf{Scenario} & \textbf{Strategy} & \textbf{Median Cost} & \textbf{Min Cost} & \textbf{Max Cost} & \textbf{Std. Dev. ($\sigma$)} \\
        \midrule
        \multirow{3}{*}{\textbf{Software}} 
        & Force Weak   & \$0.002 & \$0.001 & \$0.008 & 0.002 \\
        & Force Strong & \$0.037 & \$0.012 & \$0.105 & 0.026 \\
        & CASTER& \$0.010& \$0.003& \$0.075& 0.020\\ 
        \midrule
        \multirow{3}{*}{\textbf{Data}} 
        & Force Weak   & \$0.004 & \$0.000 & \$0.014 & 0.003 \\
        & Force Strong & \$0.043 & \$0.020 & \$0.117 & 0.023 \\
        & CASTER& \$0.018& \$0.008& \$0.103& 0.027\\ 
        \midrule
        \multirow{3}{*}{\textbf{Science}} 
        & Force Weak   & \$0.005 & \$0.004 & \$0.010 & 0.002 \\
        & Force Strong & \$0.111 & \$0.071 & \$0.251 & 0.058 \\
        & CASTER& \$0.054& \$0.004& \$0.172& 0.066\\ 
        \midrule
        \multirow{3}{*}{\textbf{Security}} 
        & Force Weak   & \$0.002 & \$0.001 & \$0.004 & 0.001 \\
        & Force Strong & \$0.005 & \$0.001 & \$0.024 & 0.005 \\
        & CASTER& \$0.005& \$0.001& \$0.008& 0.002\\ 
        \bottomrule
    \end{tabular}
\end{table}
\begin{figure}[ht]
    \centering
    \begin{subfigure}{0.48\linewidth}
        \centering
        \includegraphics[width=\linewidth]{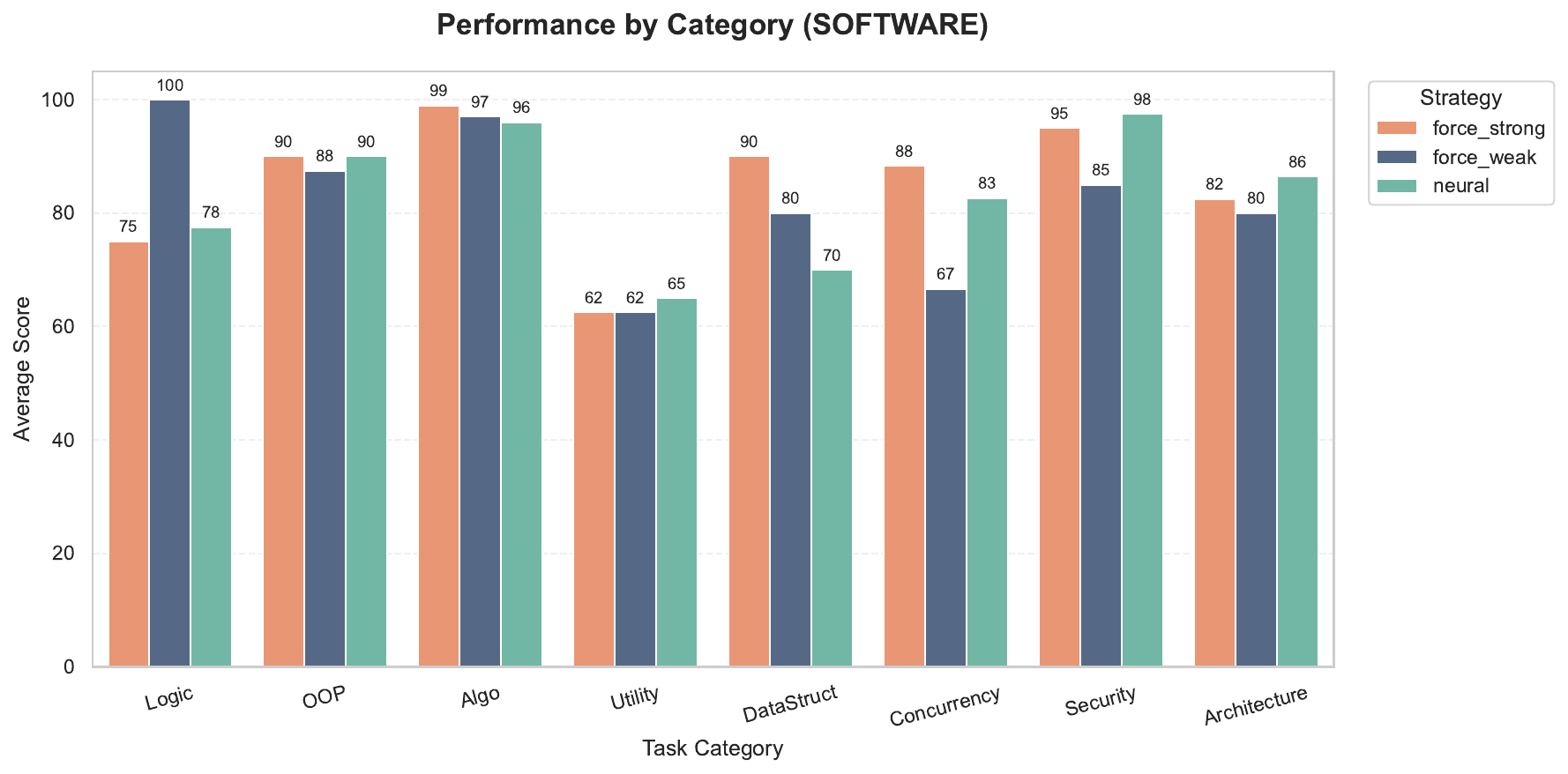}
        \caption{Software Engineering}
        \label{fig:breakdown_se}
    \end{subfigure}
    \hfill 
    \begin{subfigure}{0.48\linewidth}
        \centering
        \includegraphics[width=\linewidth]{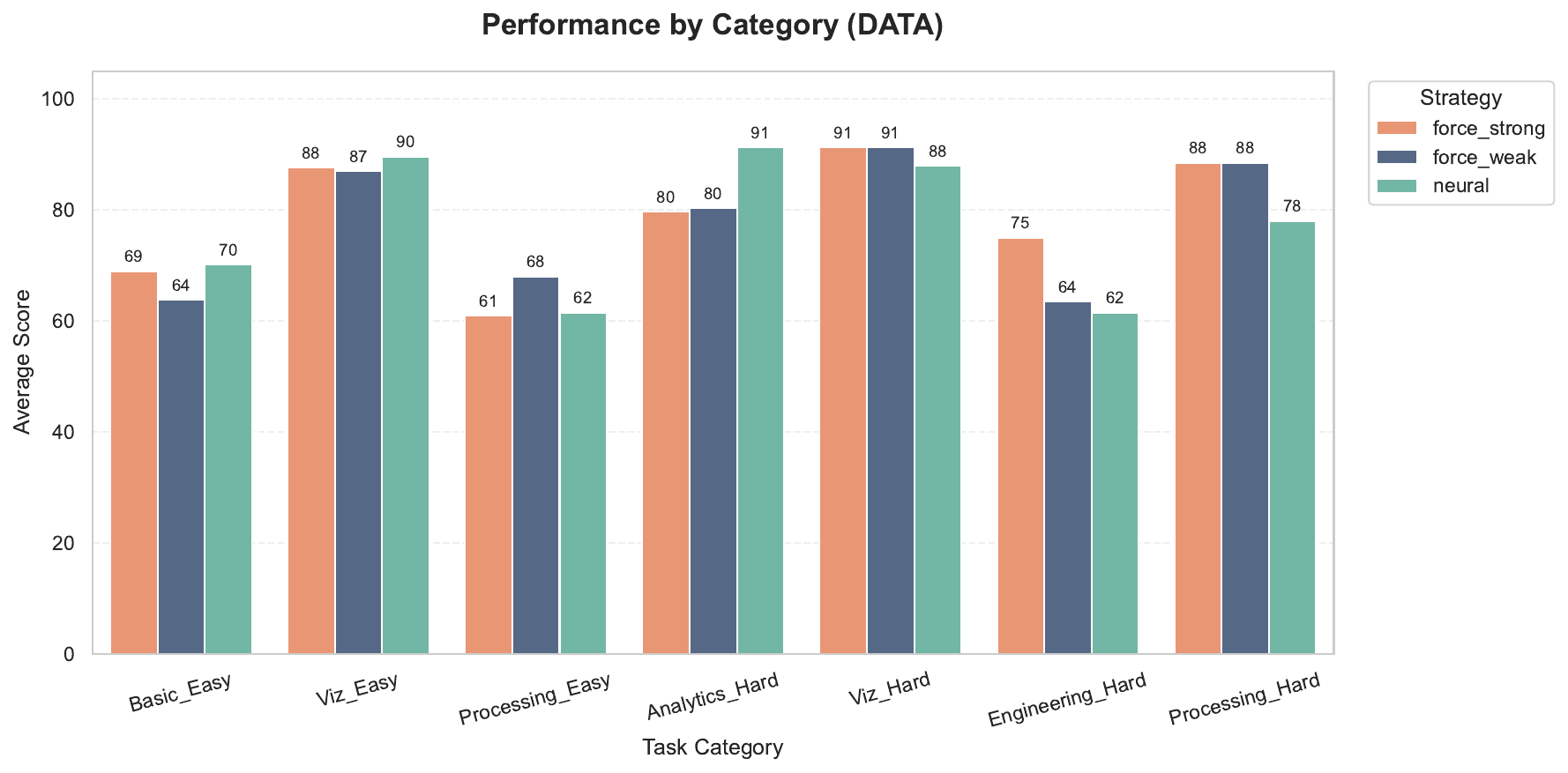}
        \caption{Data Analysis}
        \label{fig:breakdown_data}
    \end{subfigure}
    \begin{subfigure}{0.48\linewidth}
        \centering
        \includegraphics[width=\linewidth]{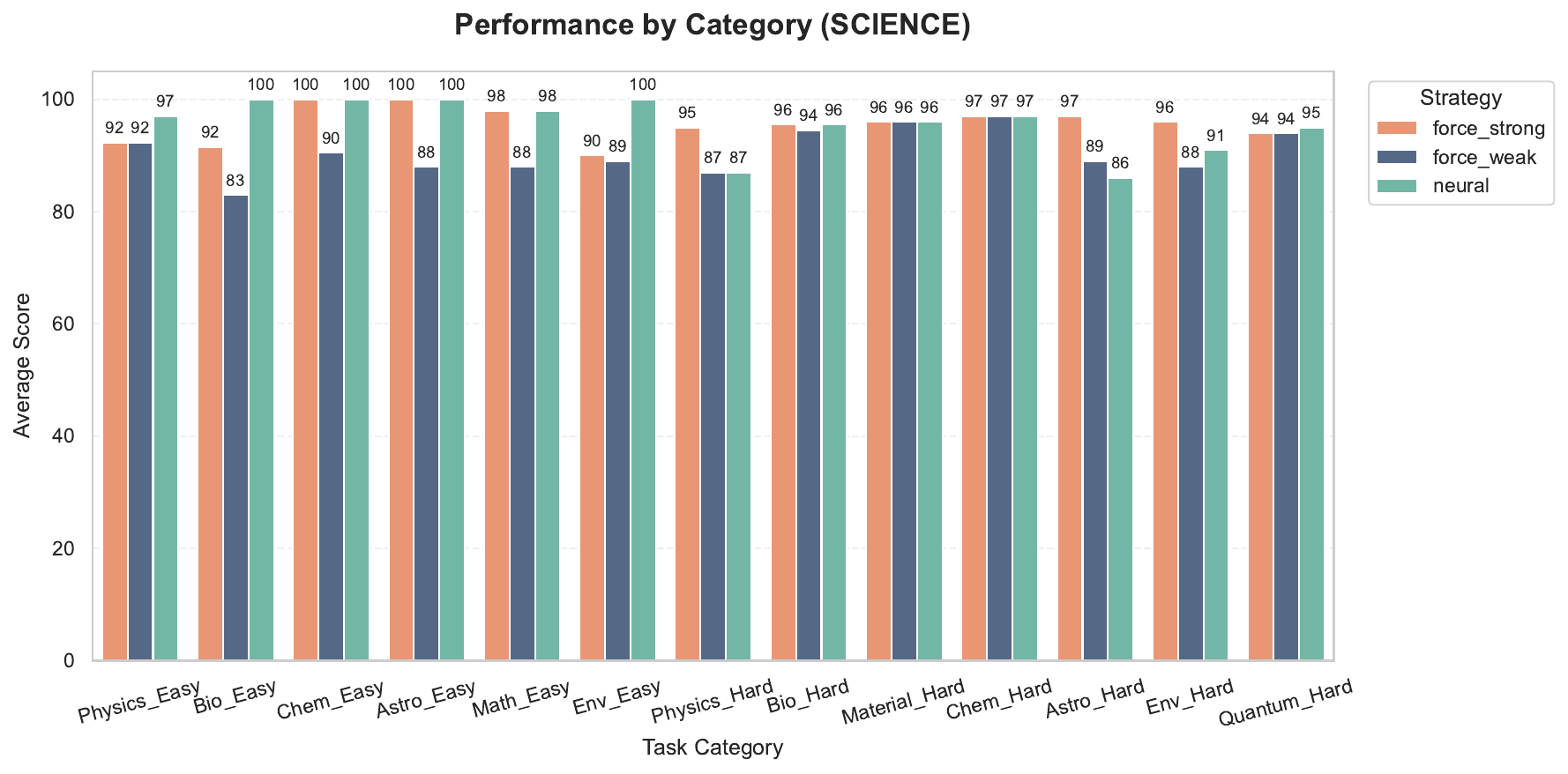}
        \caption{Scientific Discovery} 
        \label{fig:science_inf}
    \end{subfigure}
    \hfill 
    \begin{subfigure}{0.48\linewidth}
        \centering
        \includegraphics[width=\linewidth]{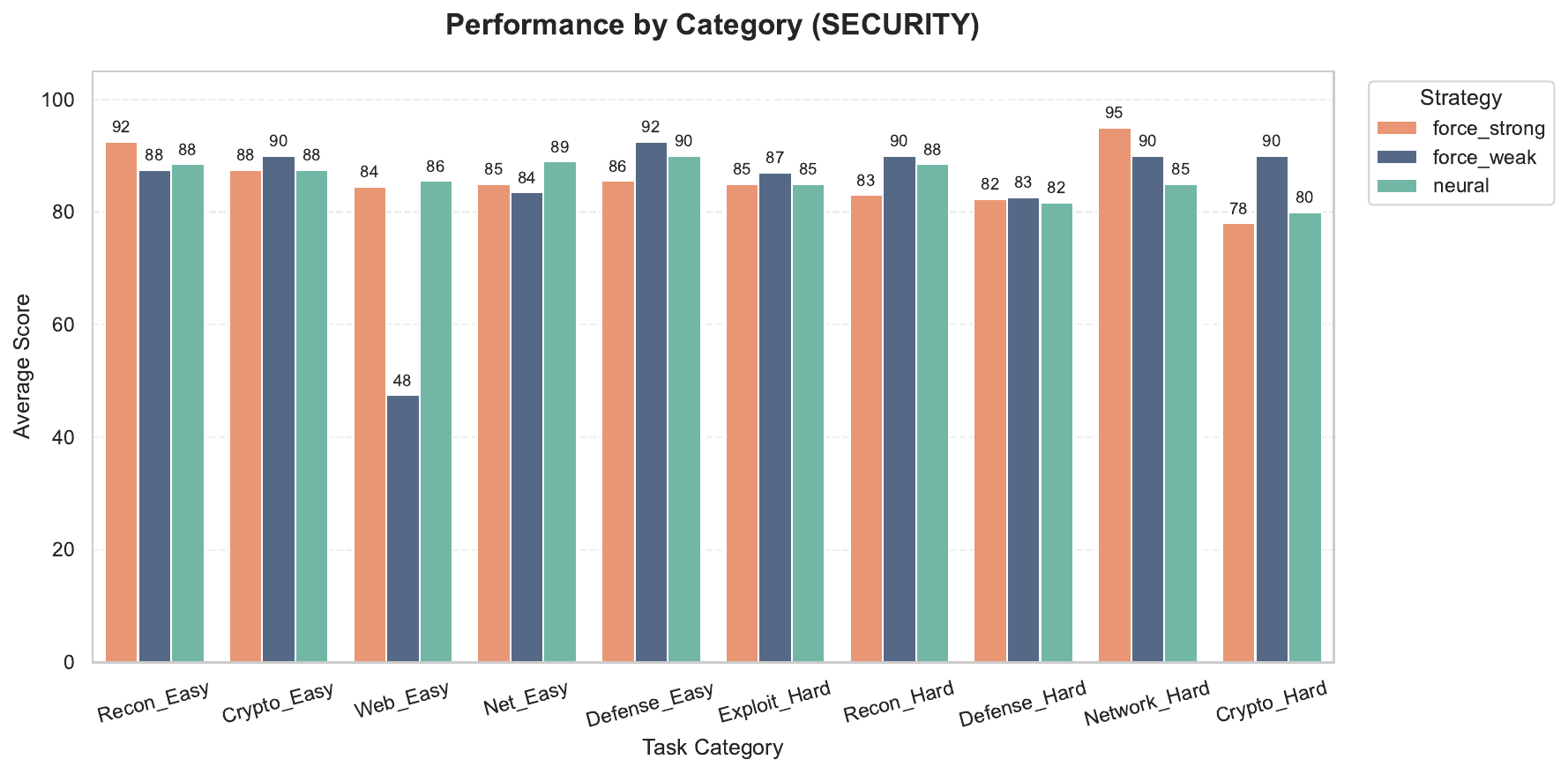}
        \caption{Cybersecurity} 
        \label{fig:security_inf}
    \end{subfigure}
    \caption{\textbf{Performance breakdown by task category across four domains.} The grouped bar charts detail the average scores across specific sub-tasks in Software Engineering, Data Analysis, Scientific Discovery, and Cybersecurity. The results highlight the robustness of CASTER (green): in high-complexity categories such as \textit{Concurrency}, \textit{Defense Operations}, and \textit{Quantum Simulation}, the Force Weak baseline (grey) suffers catastrophic performance drops (e.g., dropping to 48.0 in Security). In contrast, CASTER successfully identifies these challenges and routes them to the strong model, achieving scores comparable to the Force Strong upper bound (orange).}
    \label{fig:category_breakdown}
\end{figure}
\begin{table}[t]
    \centering
    \caption{\textbf{Breakdown of task performance scores by category. }The table highlights the robustness of the CASTER. In complex categories such as \textit{Concurrency}, \textit{Exploitation}, and \textit{Quantum Sim}, the Force Weak baseline suffers significant performance drops (highlighted in red-like low scores). The CASTER successfully detects these complexities and routes them to the strong model, restoring performance to near-upper-bound levels while maintaining high scores in simpler categories.}
    \label{tab:category_breakdown}
    \begin{tabular}{llccc}
        \toprule
        \textbf{Scenario} & \textbf{Category} &  \textbf{Force Strong} &\textbf{Force Weak} & \textbf{CASTER} \\
        \midrule
        \multirow{8}{*}{\textbf{Software}} 
        & Logic &  75 &\textbf{100} & 78 \\
        & OOP &  \textbf{90} &88 & \textbf{90} \\
        & Algorithm &  \textbf{99} &97 & 96 \\
        & Utility &  62 &62 & \textbf{65} \\
        & Data Struct. &  \textbf{90} &80 & 70 \\
        & Concurrency&  \textbf{88}&67 & 83\\ 
        & Security&  95 &85 & \textbf{98} \\ 
        & Architecture&  82 &80 & \textbf{86} \\ 
        \midrule
        \multirow{5}{*}{\textbf{Data}} 
        & Basic &  69&64& \textbf{70}\\
        & Visualization (Easy)&  88&87& \textbf{90}\\
        & Processing (Easy)&  61&\textbf{68}& 62\\ 
        & Analytics &  80&80& \textbf{91}\\
 & Visualization (Hard)&  \textbf{91}&\textbf{91}&88\\
 & Engineering &  \textbf{75}&64&62\\
        & Processing (Hard)&  \textbf{88}&\textbf{88}& 78\\
        \midrule
        \multirow{5}{*}{\textbf{Science}} 
        & Physics (Easy)&  92&92& \textbf{97}\\
        & Biology (Easy)&  92&83& \textbf{100}\\ 
        & Chemistry (Easy)&  \textbf{100}&90& \textbf{100}\\ 
        & Astrophysics (Easy)&  \textbf{100}&88& \textbf{100}\\ 
        & Mathematics (Easy)&  \textbf{98}&88& \textbf{98}\\
 & Environmental Science (Easy)&  90&89&\textbf{100}\\
 & Physics (Hard)&  \textbf{95}&87&87\\
 & Biology (Hard)&  \textbf{96}&94&\textbf{96}\\
 & Material Science (Hard)&  96&96&96\\
 & Chemistry (Hard)&  97&97&97\\
 & Astrophysics (Hard)&  \textbf{97}&89&86\\
 & Environmental Science (Hard)&  \textbf{96}&88&91\\
 & Quantum (Hard)&  94&94&\textbf{95}\\ 
        \midrule
        \multirow{5}{*}{\textbf{Security}} 
        & Reconnaissance (Easy)&  \textbf{92}&88& 88\\
 & Cryptography (Easy)&  88&\textbf{90}&88\\
 & Web Security&  84&48&\textbf{86}\\
 & Network Security (Easy)&  85&84&\textbf{89}\\
 & Defensive Security (Easy)&  86&\textbf{92}&90\\
        & Exploitation&  85&\textbf{87}& 85\\ 
        & Reconnaissance (Hard)&  83&\textbf{90}& 88\\
 & Defensive Security (Hard)&  82&\textbf{83}&82\\ 
        & Network Security (Hard)&  \textbf{95}&90& 85\\ 
        & Cryptography (Hard)&  78&\textbf{90}& 80\\
        \bottomrule
    \end{tabular}
\end{table}
\begin{figure}[ht]
    \centering
    \begin{subfigure}{0.48\linewidth}
        \centering
        \includegraphics[width=\linewidth]{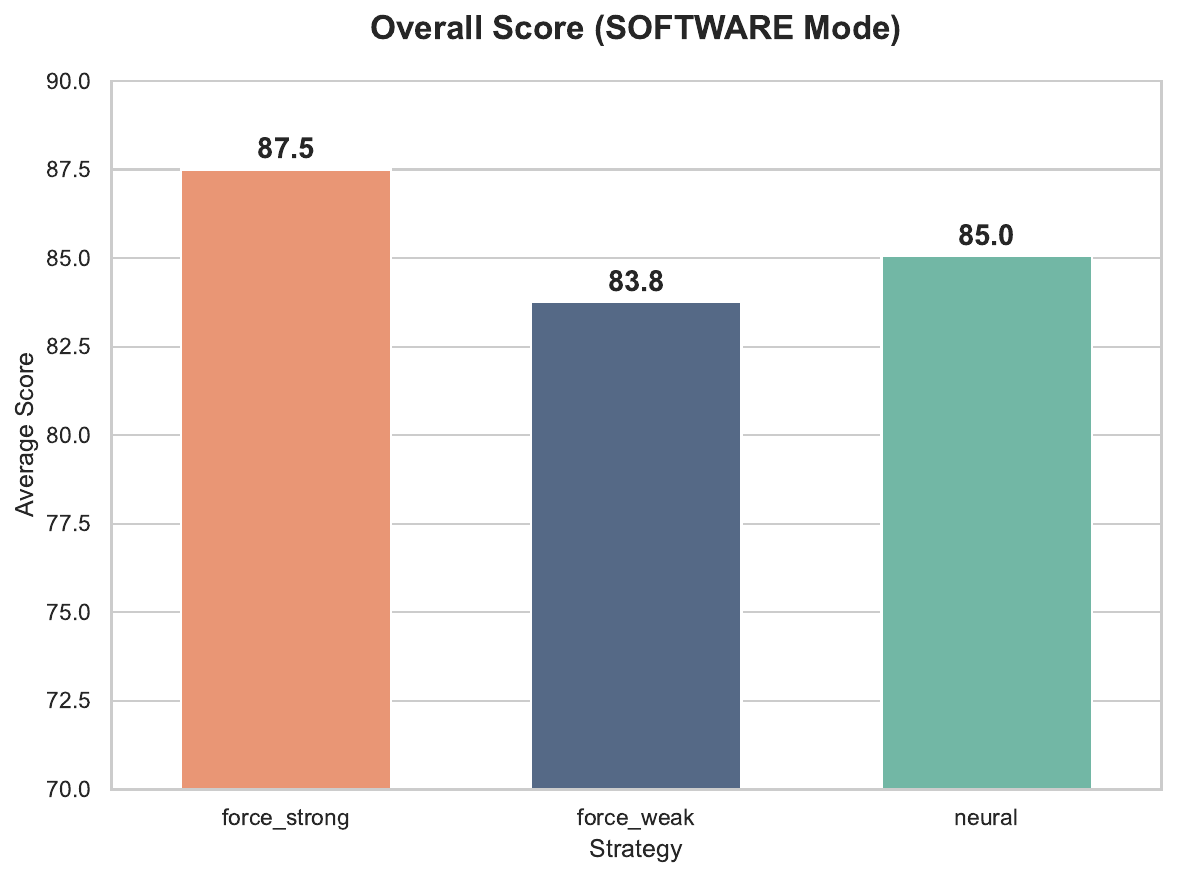}
        \caption{Software Engineering Scenario}
        \label{fig:score_se}
    \end{subfigure}
    \hfill 
    \begin{subfigure}{0.48\linewidth}
        \centering
        \includegraphics[width=\linewidth]{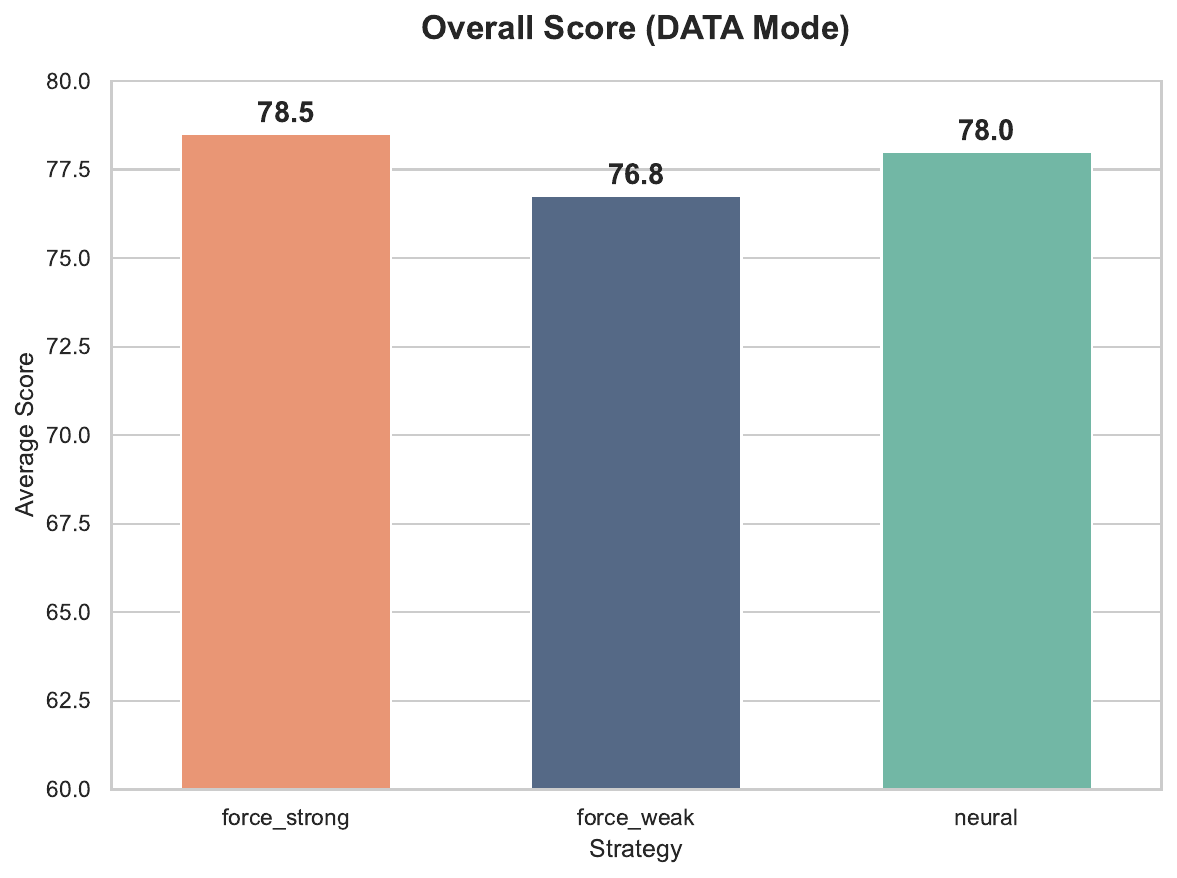}
        \caption{Data Analysis Scenario}
        \label{fig:score_data}
    \end{subfigure}
    \begin{subfigure}{0.48\linewidth}
        \centering
        \includegraphics[width=\linewidth]{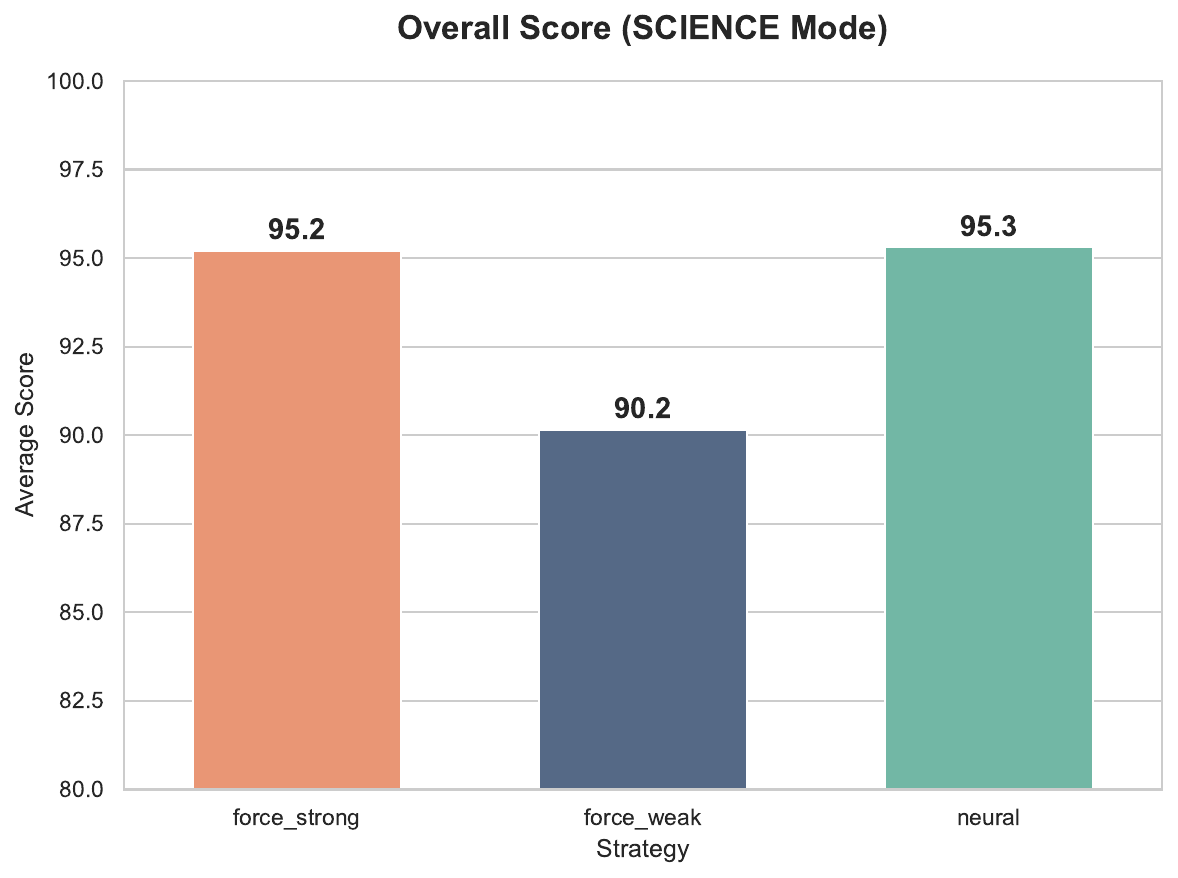}
        \caption{Scientific Discovery} 
        \label{fig:science_inf}
    \end{subfigure}
    \hfill 
    \begin{subfigure}{0.48\linewidth}
        \centering
        \includegraphics[width=\linewidth]{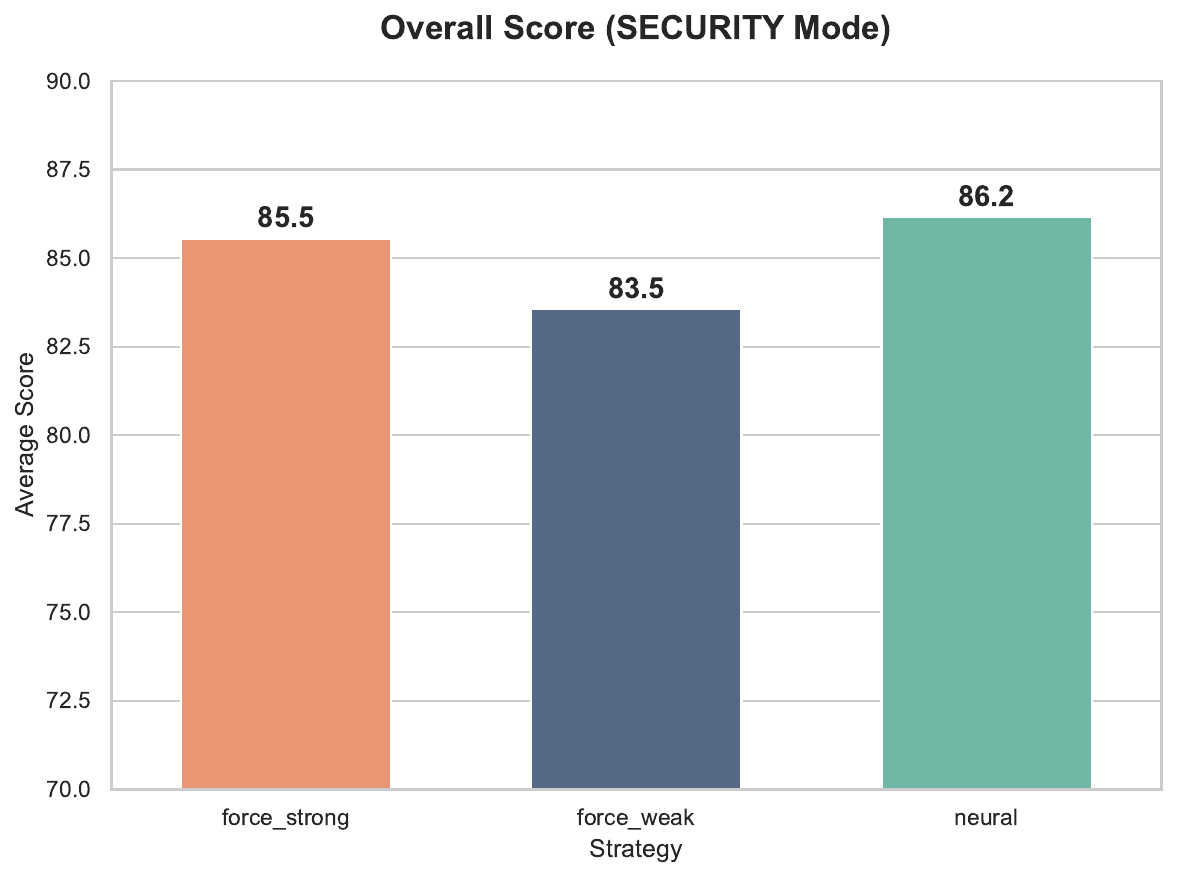}
        \caption{Cybersecurity} 
        \label{fig:security_inf}
    \end{subfigure}
    \caption{\textbf{Comparison of Overall Task Scores across Four Domains.} The bar charts illustrate that CASTER (green) consistently bridges the quality gap between the Force Weak baseline (grey) and the Force Strong upper bound (orange).}
    \label{fig:overall_scores}
\end{figure}
\begin{table}[t]
    \centering
    \caption{\textbf{Comparison of overall task performance scores.} The table validates the effectiveness of our routing strategy. In the Software Engineering scenario, the CASTER recovers most of the performance drop caused by the weak model. Notably, in Science and Security scenarios, the CASTER achieves performance (95.3 and 86.2) that matches or even slightly surpasses the Force Strong baseline, demonstrating that cost-aware routing can effectively mitigate overfitting or "over-thinking" in simple tasks.}
    \label{tab:overall_performance}
    \begin{tabular}{llc}
        \toprule
        \textbf{Scenario} & \textbf{Strategy} & \textbf{Average Score} \\
        \midrule
        \multirow{3}{*}{\textbf{Software}} 
        & Force Weak   & 83.8 \\
        & CASTER& 85.0 \\ 
        & Force Strong & \textbf{87.5}\\ 
        \midrule
        \multirow{3}{*}{\textbf{Data}} 
        & Force Weak   & 76.8\\
        & CASTER& 78.0\\ 
        & Force Strong & \textbf{78.5}\\ 
        \midrule
        \multirow{3}{*}{\textbf{Science}} 
        & Force Weak   & 90.2 \\
        & CASTER& \textbf{95.3} \\ 
        & Force Strong & \textbf{95.2}\\ 
        \midrule
        \multirow{3}{*}{\textbf{Security}} 
        & Force Weak   & 83.5 \\
        & CASTER& \textbf{86.2} \\ 
        & Force Strong & 85.5 \\ 
        \bottomrule
    \end{tabular}
\end{table}

\begin{figure}[ht] 
    \centering
    \begin{subfigure}{0.48\linewidth}
        \centering
        \includegraphics[width=\linewidth]{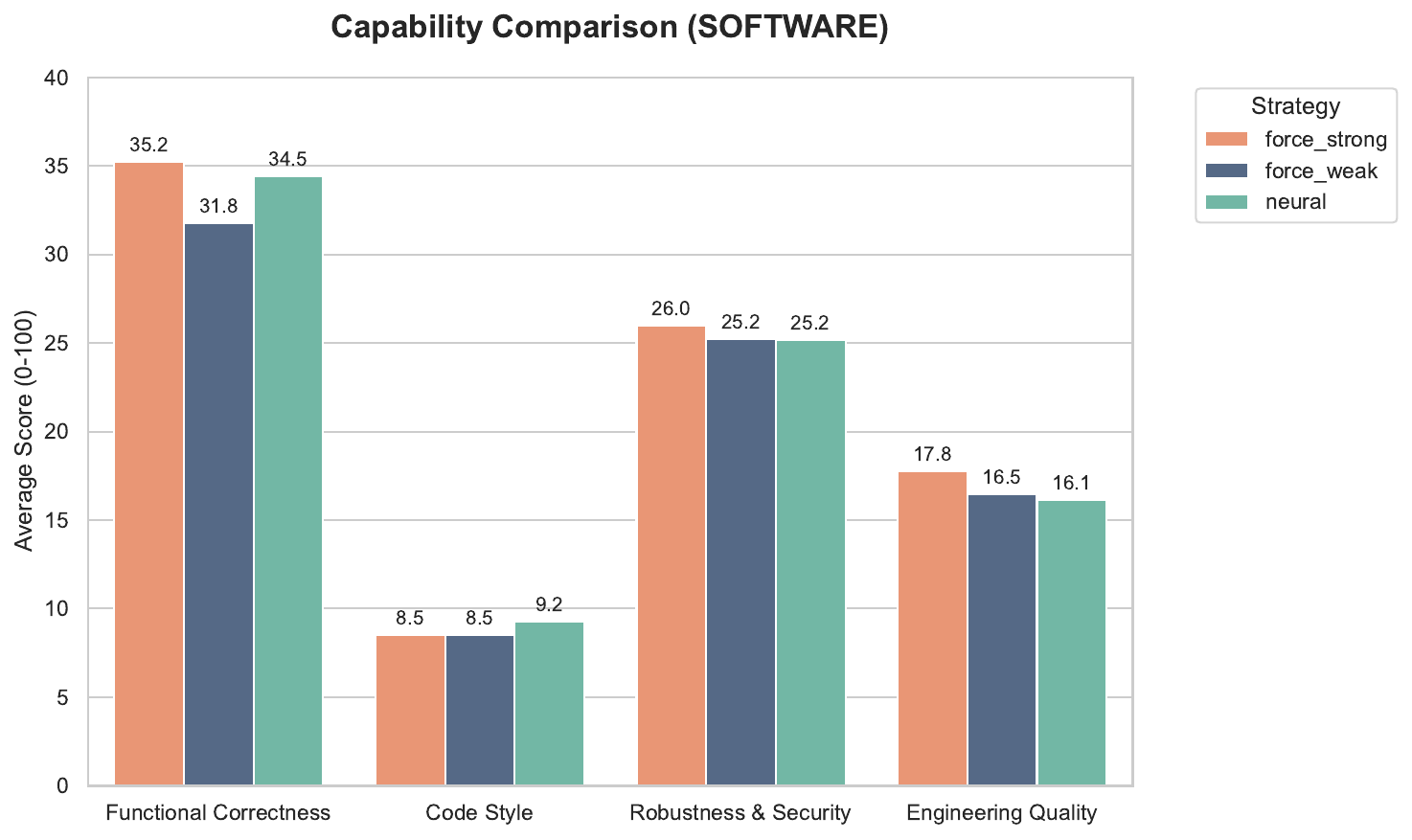}
        \caption{Software Engineering}
        \label{fig:cap_se}
    \end{subfigure}
    \hfill 
    \begin{subfigure}{0.48\linewidth}
        \centering
        \includegraphics[width=\linewidth]{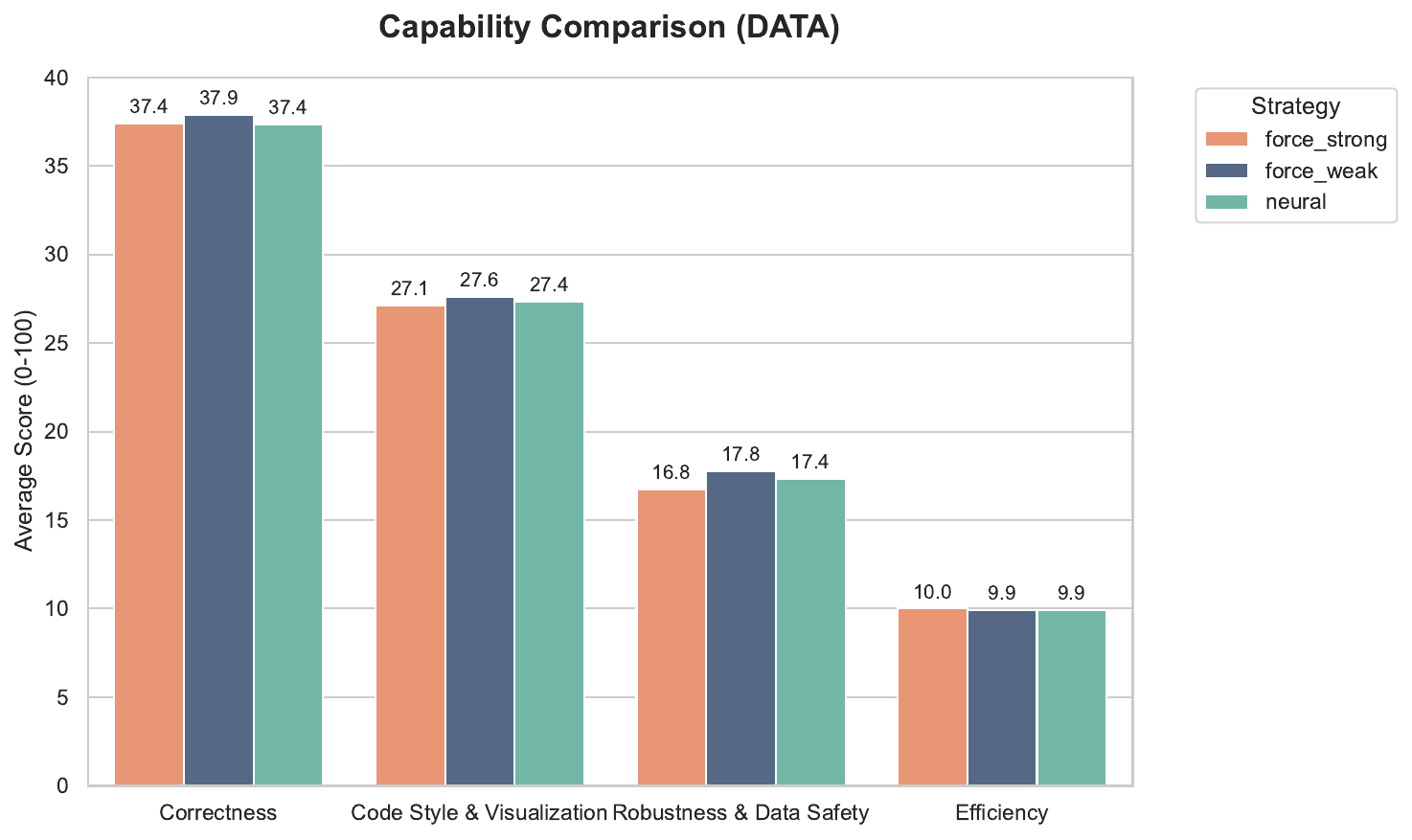}
        \caption{Data Analysis}
        \label{fig:cap_data}
    \end{subfigure}
    \begin{subfigure}{0.48\linewidth}
        \centering
        \includegraphics[width=\linewidth]{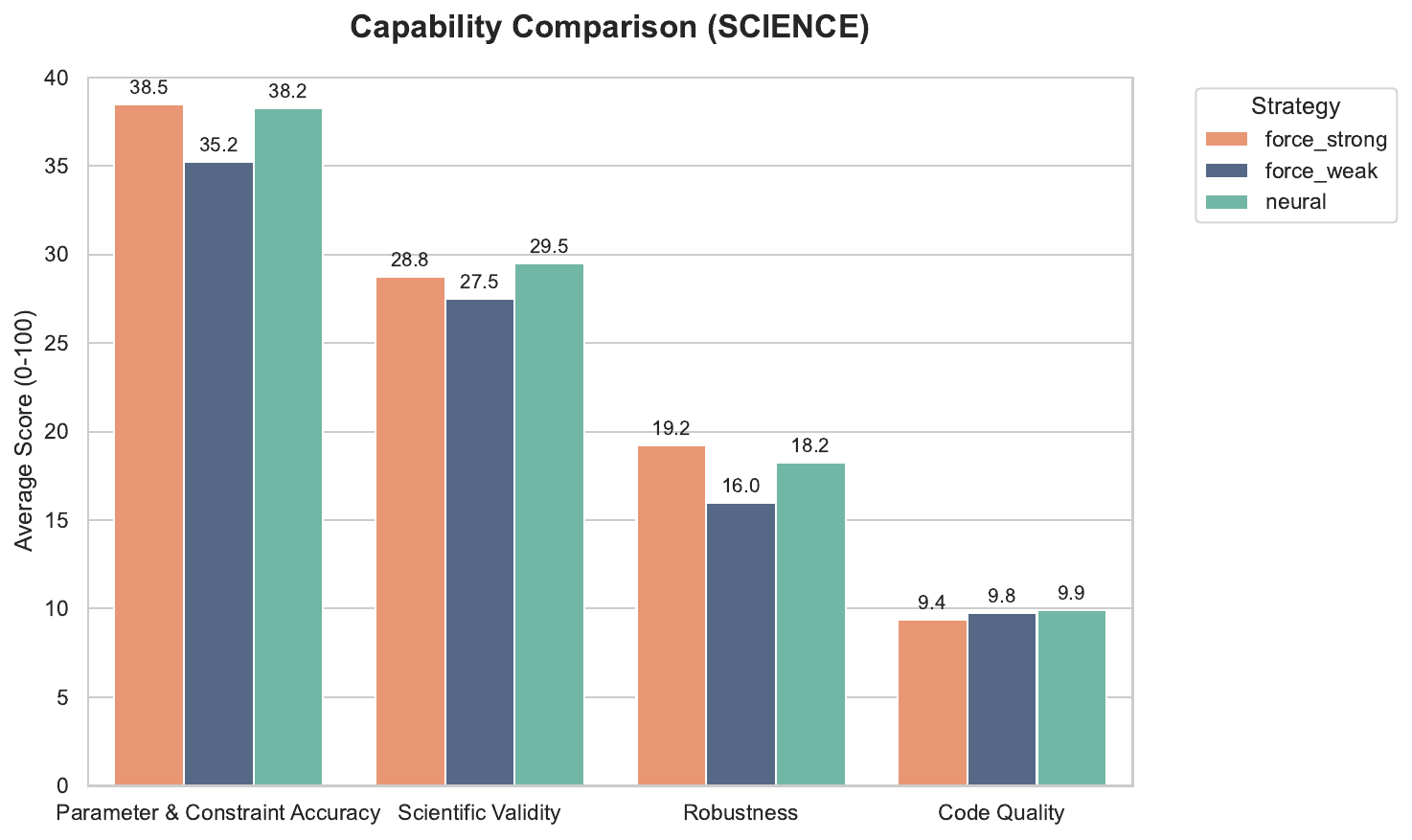}
        \caption{Scientific Discovery} 
        \label{fig:science_inf}
    \end{subfigure}
    \hfill 
    \begin{subfigure}{0.48\linewidth}
        \centering
        \includegraphics[width=\linewidth]{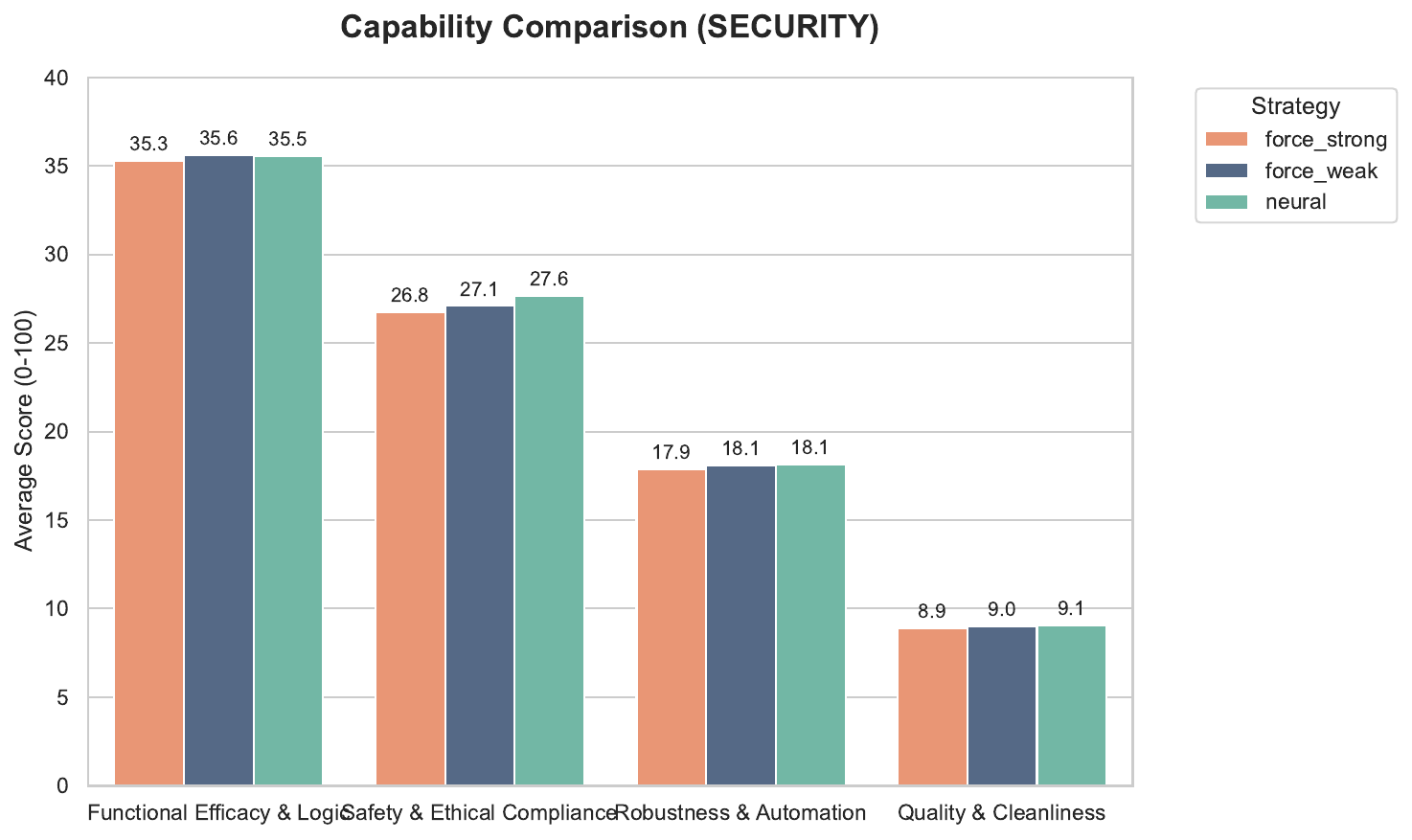}
        \caption{Cybersecurity} 
        \label{fig:security_inf}
    \end{subfigure}
    \caption{\textbf{Multi-dimensional capability assessment across four domains.} The grouped bar charts compare the performance of Force Strong, Force Weak, and CASTER strategies across four key metrics. The results demonstrate that CASTER (green) achieves a dual advantage: it not only recovers the deficit of the weak baseline (e.g., improving from 35.2 to 38.2 in Parameter \& Constraint Accuracy of Science) but also surpasses the Force Strong baseline in some scores across the Security and Software domains (e.g., 27.6 vs. 26.8 in Cleanliness of Security), suggesting an optimization in output formatting and compliance.}
    \label{fig:capability_breakdown}
\end{figure}

\begin{table}[t]
    \centering
    \caption{\textbf{Multi-dimensional quality evaluation.} The results illustrate CASTER's ability to maintain high-performance standards while optimizing specific output traits.  In complex reasoning tasks, CASTER closely mirrors the Force Strong baseline, achieving \textbf{34.5} in \textit{Software} 'Functional Correctness' (vs. 35.2) and \textbf{38.2} in \textit{Science} 'Parameter Accuracy' (vs. 38.5), effectively avoiding the performance dip seen in the Force Weak strategy (31.8 and 35.2, respectively). Notably, CASTER outperforms the Force Strong baseline in qualitative metrics, securing the highest scores in \textit{Software} 'Code Style' (\textbf{9.2}) and \textit{Security} 'Safety \& Compliance' (\textbf{27.6}), suggesting that dynamic routing can leverage model-specific strengths to enhance formatting and adherence to safety protocols.}
    \label{tab:multidimensional_quality}
    \begin{tabular}{llccc}
        \toprule
        \textbf{Scenario} & \textbf{Metric} &\textbf{Force Strong}& \textbf{Force Weak}& \textbf{CASTER} \\
        \midrule
        \multirow{4}{*}{\textbf{Software}} 
        & Functional Correctness&\textbf{35.2} & 31.8& 34.5\\ 
        & Robustness \& Security&\textbf{26.0} & 25.2 & 25.2 \\
        & Engineering Quality&\textbf{17.8} & 16.5 & 16.1 \\
        & Code Style&8.5  & 8.5  & \textbf{9.2} \\ 
        \midrule
        \multirow{4}{*}{\textbf{Data}} 
        & Correctness&37.4& \textbf{37.9}& 37.4\\ 
        & Code Style \& Visualization&27.1& \textbf{27.6}& 27.4\\
        & Robustness \& Data Safety&16.8& \textbf{17.8}& 17.4\\ 
        & Efficiency&\textbf{10.0}& 9.9& 9.9\\ 
        \midrule
        \multirow{4}{*}{\textbf{Science}} 
        & Parameter \& Constraint Accuracy&\textbf{38.5} & 35.2 & \textbf{38.2} \\ 
        & Scientific Validity&28.8 & 27.5& \textbf{29.5} \\ 
        & Robustness&\textbf{19.2} & 16& 18.2 \\
        & Code Quality&9.4  & 9.8  & \textbf{9.9} \\
        \midrule
        \multirow{4}{*}{\textbf{Security}} 
        & Functional Efficacy \& Logic&35.3 & \textbf{35.6} & 35.5 \\ 
        & Safety \& Ethical Compliance&26.8 & 27.1 & \textbf{27.6} \\ 
        & Robustness \& Automation&17.9 & \textbf{18.1} & \textbf{18.1} \\
        & Quality \& Cleanliness&8.9  & 9.0  & \textbf{9.1} \\ 
        
        \bottomrule
    \end{tabular}
\end{table}
\begin{figure}[h] 
    \centering
    \includegraphics[width=0.48\linewidth]{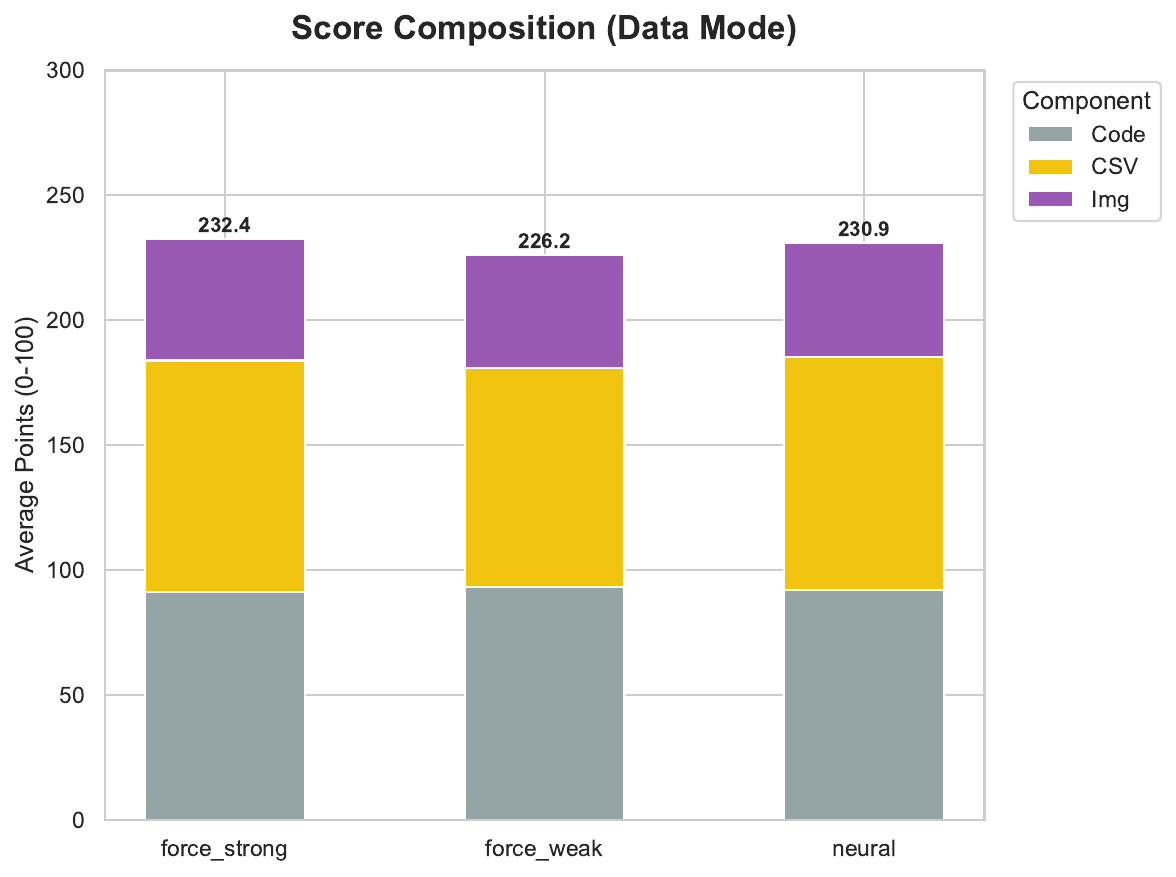}
    \caption{\textbf{Component-wise score composition in Data Analysis tasks.} The total score is decomposed into Code quality (grey), CSV data quality (yellow), and Visualization quality (purple). The CASTER (230.9) successfully mitigates the degradation observed in the weak baseline (226.2), achieving a multi-modal output quality nearly indistinguishable from the strong model (232.4).}
    \label{fig:data_score_composition}
\end{figure}
\begin{table}[t]
    \centering
    \caption{\textbf{Component-wise score composition in Data Analysis.} The evaluation metric aggregates scores from three generated outputs: Executable Code, CSV Data Files, and Image Plots. The breakdown reveals that while the Force Weak baseline lags behind due to lower quality in data and visual artifacts, the CASTER effectively recovers performance, particularly in CSV generation quality, matching the Force Strong upper bound closely.}
    \label{tab:score_composition}
    \begin{tabular}{lcccc}
        \toprule
        \multirow{2}{*}{\textbf{Strategy}} & \multirow{2}{*}{\textbf{Total Score}} & \multicolumn{3}{c}{\textbf{Component Breakdown}} \\
        \cmidrule(lr){3-5}
        & & \textbf{Code} & \textbf{CSV File} & \textbf{Image Plot} \\
        \midrule
        Force Weak    & 226.2 & 93.2 & 88.0 & 45.0 \\
        \textbf{CASTER} & \textbf{230.9} & \textbf{92.0} & \textbf{93.5} & \textbf{45.4} \\ 
        \textbf{Force Strong}& \textbf{232.4}& \textbf{91.3}& \textbf{92.7}& \textbf{48.5}\\
        \bottomrule
    \end{tabular}
\end{table}
\begin{figure}[ht]
    \centering
    \begin{subfigure}{0.48\linewidth}
        \centering
        \includegraphics[width=\linewidth]{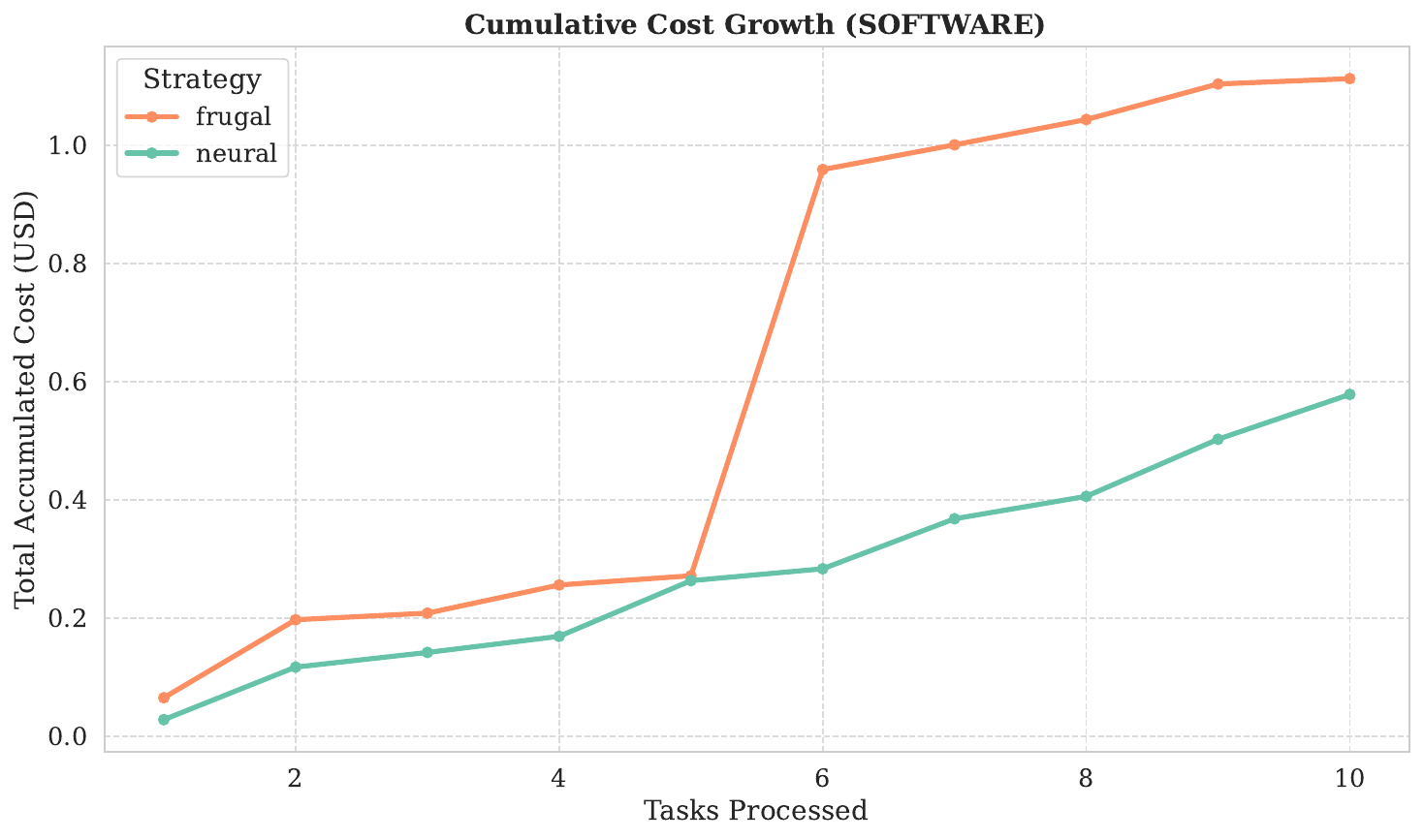}
        \caption{Software Engineering}
        \label{fig:cost_software_with_frugal}
    \end{subfigure}
    \hfill 
    \begin{subfigure}{0.48\linewidth}
        \centering
        \includegraphics[width=\linewidth]{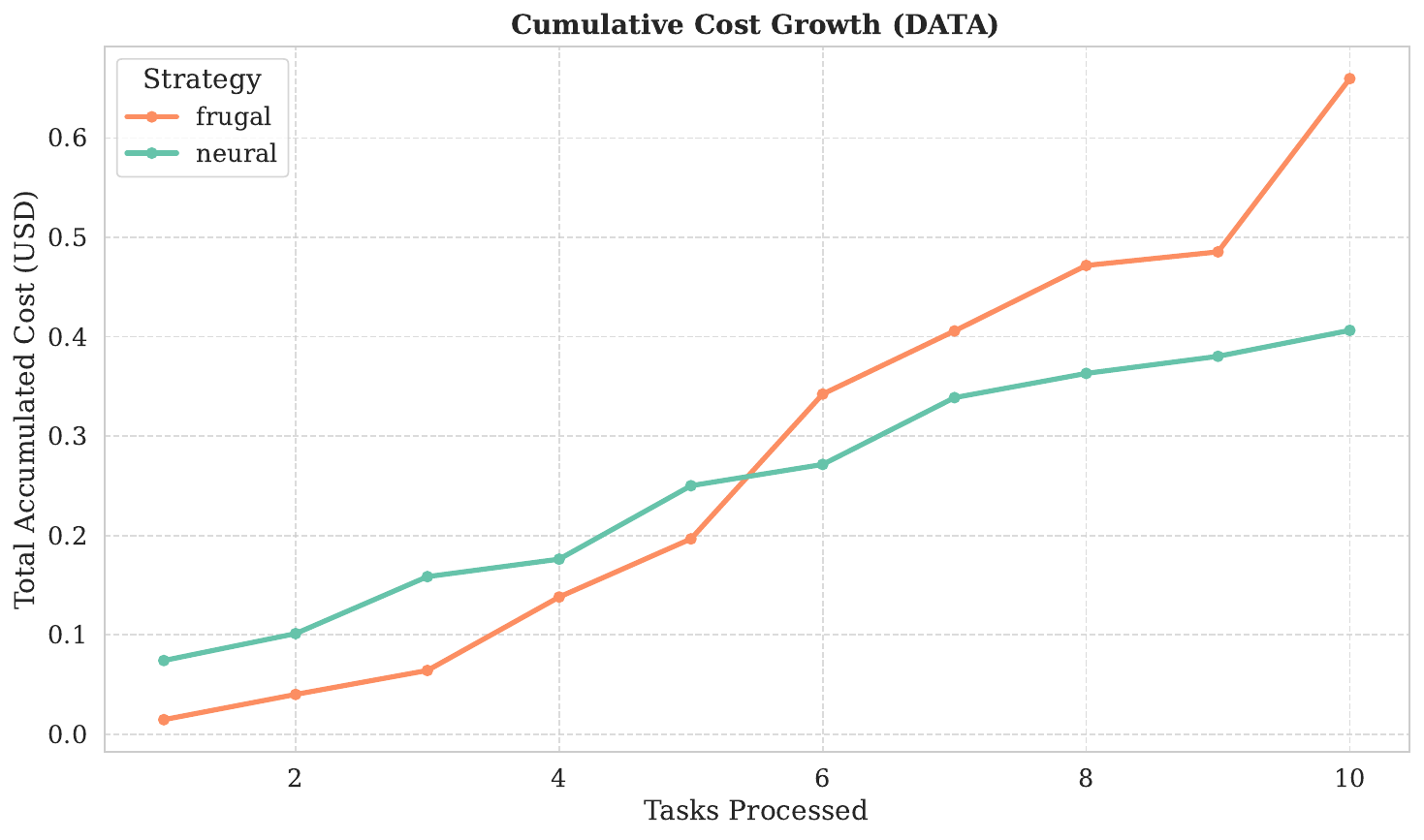}
        \caption{Data Analysis}
        \label{fig:cost_data_with_frugal}
    \end{subfigure}
    \begin{subfigure}{0.48\linewidth}
        \centering
        \includegraphics[width=\linewidth]{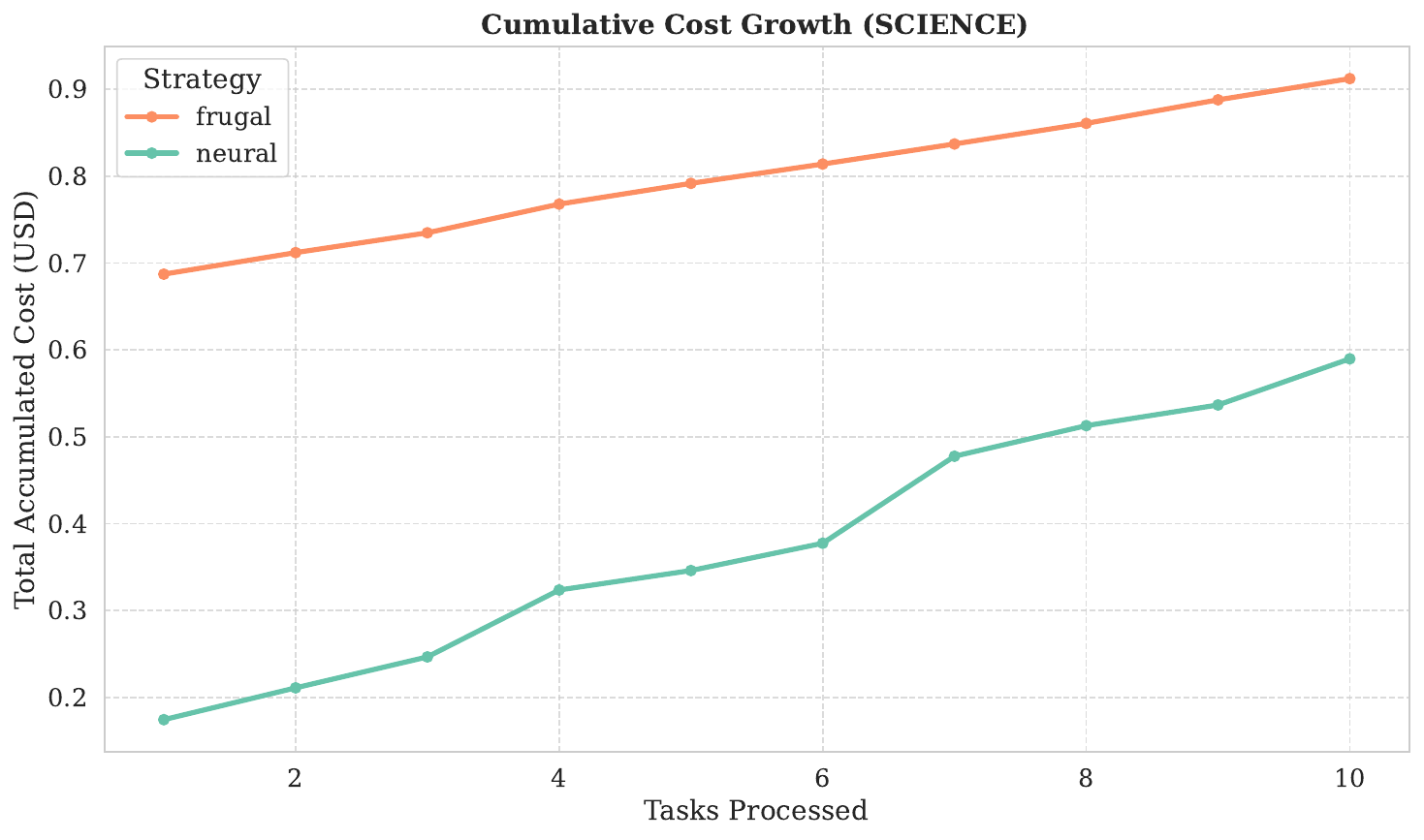}
        \caption{Scientific Discovery} 
        \label{fig:science_inf}
    \end{subfigure}
    \hfill 
    \begin{subfigure}{0.48\linewidth}
        \centering
        \includegraphics[width=\linewidth]{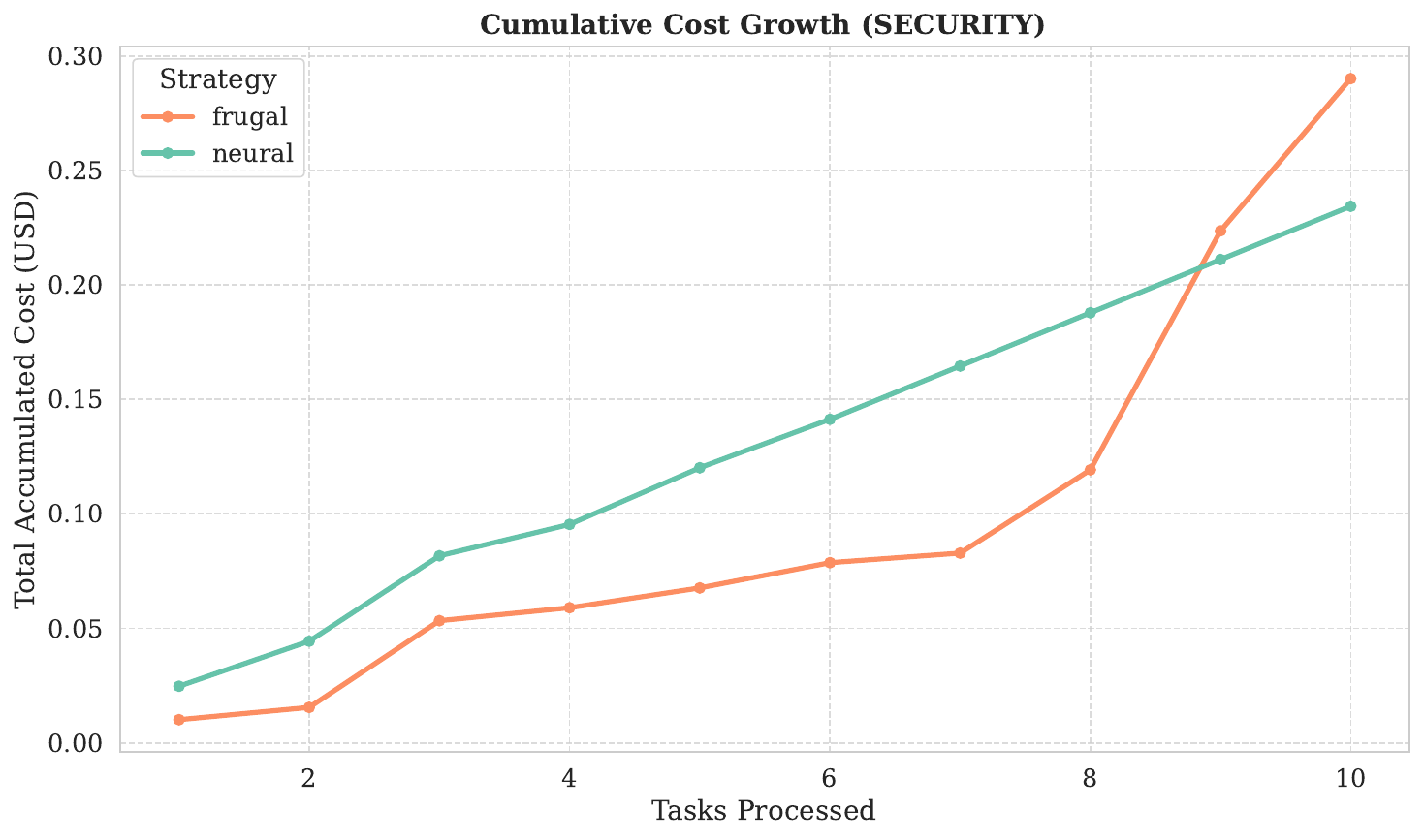}
        \caption{Cybersecurity} 
        \label{fig:security_inf}
    \end{subfigure}
    \caption{\textbf{Cumulative cost growth comparison between CASTER and FrugalGPT across a sequence of complex tasks.} The plot illustrates the accumulated token cost (USD) over 10 consecutive hard tasks (e.g., concurrency control, security architecture). FrugalGPT(Cascade) exhibits a steeper cost trajectory due to the "cascading overhead" incurred by failing with weak models before upgrading. In contrast, CASTER minimizes cost by identifying task complexity upfront and directly routing to the strong model, effectively bypassing wasteful trial-and-error iterations.}
    \label{fig:cumulative_cost_comparison_with_frugal}
\end{figure}
\begin{table}[t]
    \centering
    \caption{\textbf{Comparative analysis of dynamic routing strategies.} FrugalGPT (Cascade) vs. CASTER. The table summarizes the total accumulated cost after processing 10 tasks. Results indicate that the CASTER consistently outperforms the Frugal strategy. By predicting complexity upfront rather than relying on a "fail-then-retry" cascade, the CASTER avoids the overhead of double-billing, achieving cost reductions ranging from \textbf{20.7\%} to \textbf{48.0\%}.}
    \label{tab:frugal_vs_neural}
    \begin{tabular}{llcc}
        \toprule
        \textbf{Scenario} & \textbf{Strategy} & \textbf{Total Cost (USD)} & \textbf{Cost Reduction} \\
        \midrule
        \multirow{2}{*}{\textbf{Software}} 
        & FrugalGPT (Cascade) & \$1.11& - \\ 
        & \textbf{CASTER} & \textbf{\$0.58}& \textbf{48.0\%}\\ 
        \midrule
        \multirow{2}{*}{\textbf{Data}} 
        & FrugalGPT (Cascade) & \$0.66& - \\ 
        & \textbf{CASTER} & \textbf{\$0.41}& \textbf{38.4\%}\\ 
        \midrule
        \multirow{2}{*}{\textbf{Science}} 
        & FrugalGPT (Cascade) & \$0.91& - \\ 
        & \textbf{CASTER} & \textbf{\$0.59}& \textbf{35.3\%}\\ 
        \midrule
        \multirow{2}{*}{\textbf{Security}} 
        & FrugalGPT (Cascade) & \$0.29& - \\ 
        & \textbf{CASTER} & \textbf{\$0.23}& \textbf{20.7\%}\\ 
        \bottomrule
    \end{tabular}
\end{table}

\begin{figure}[ht]
    \centering
    \begin{subfigure}{0.48\linewidth}
        \centering
        \includegraphics[width=\linewidth]{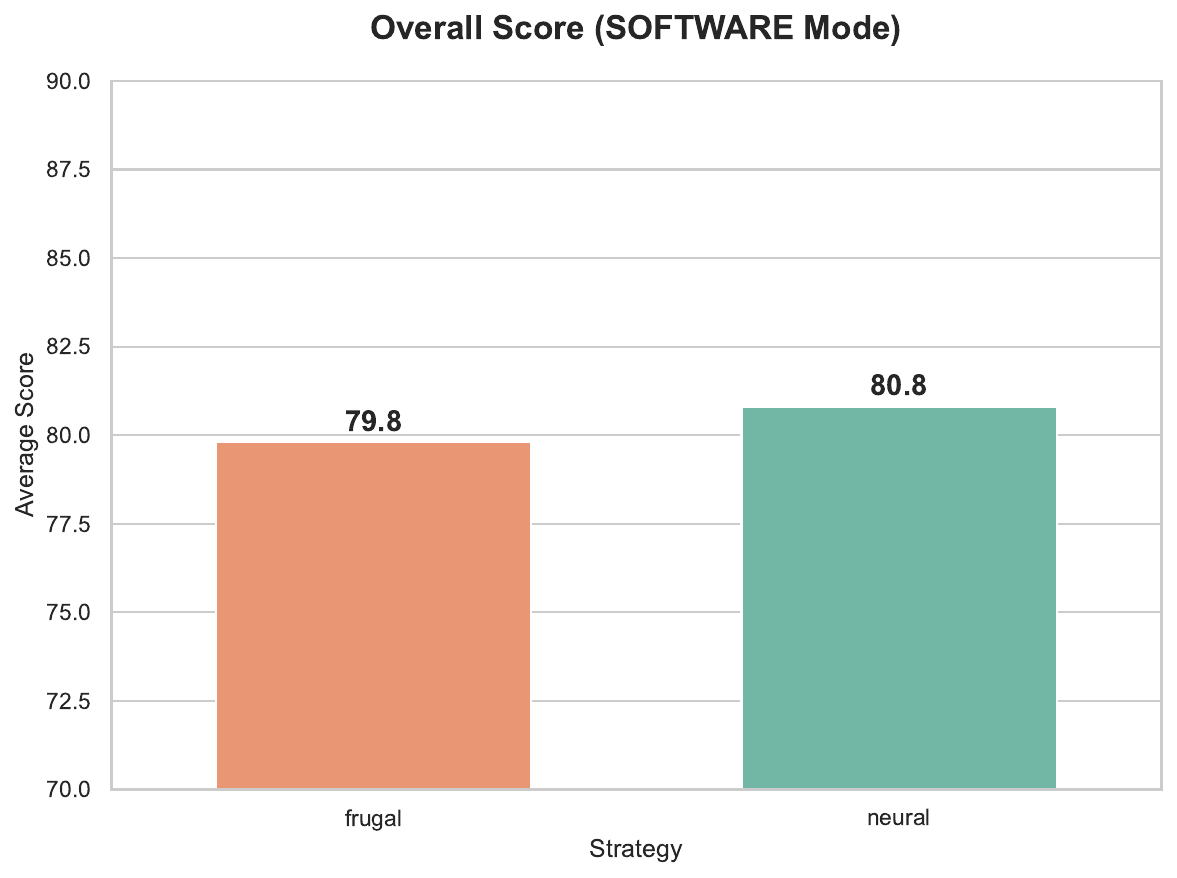}
        \caption{Software Engineering}
        \label{fig:score_software_with_frugal}
    \end{subfigure}
    \hfill 
    \begin{subfigure}{0.48\linewidth}
        \centering
        \includegraphics[width=\linewidth]{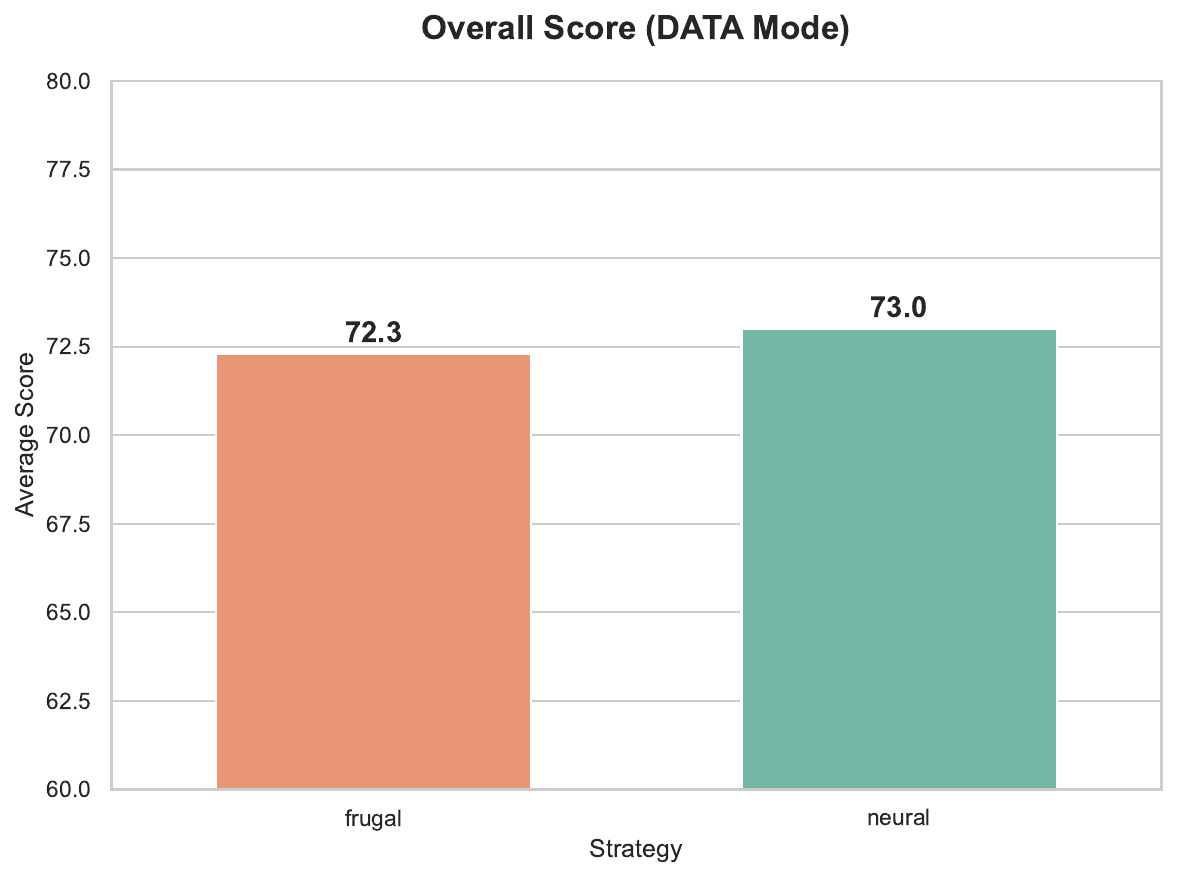}
        \caption{Data Analysis}
        \label{fig:score_data_with_frugal}
    \end{subfigure}
    \begin{subfigure}{0.48\linewidth}
        \centering
        \includegraphics[width=\linewidth]{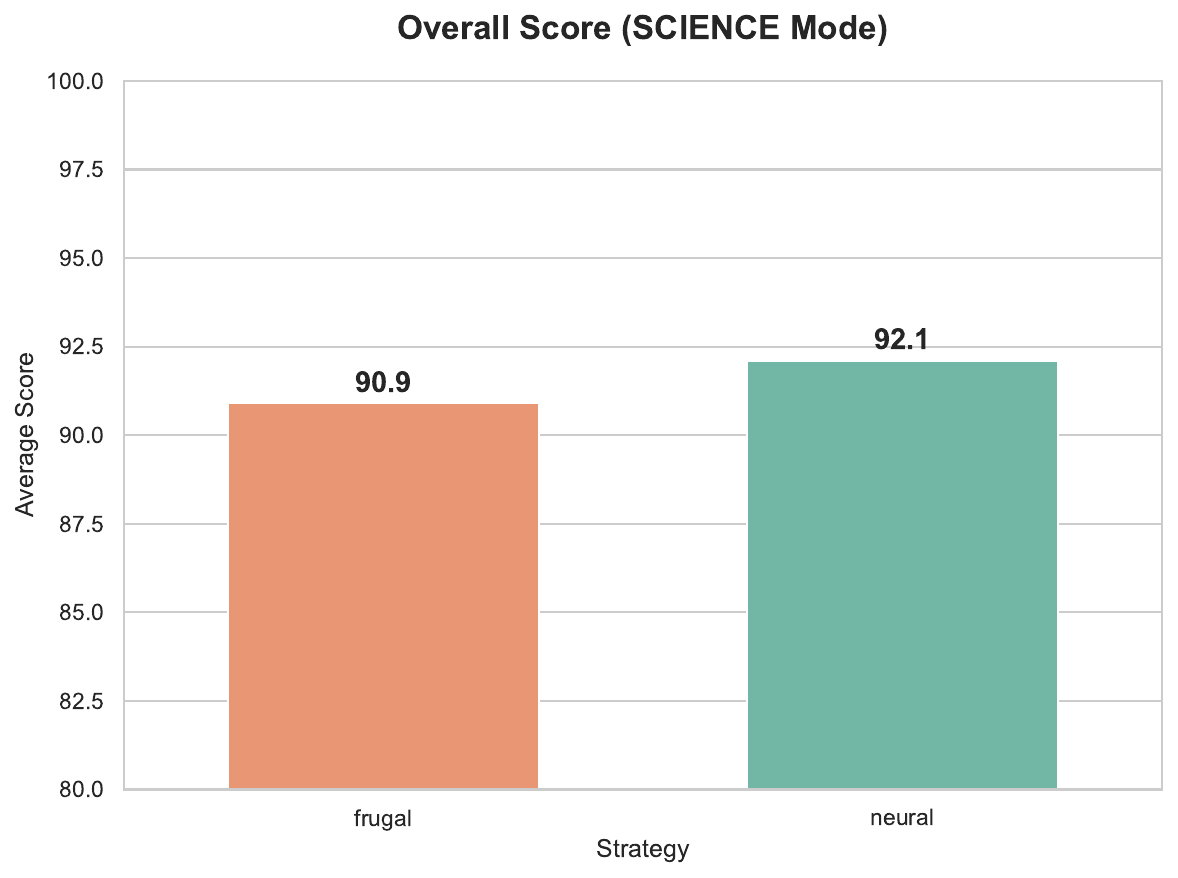}
        \caption{Scientific Discovery} 
        \label{fig:science_inf}
    \end{subfigure}
    \hfill 
    \begin{subfigure}{0.48\linewidth}
        \centering
        \includegraphics[width=\linewidth]{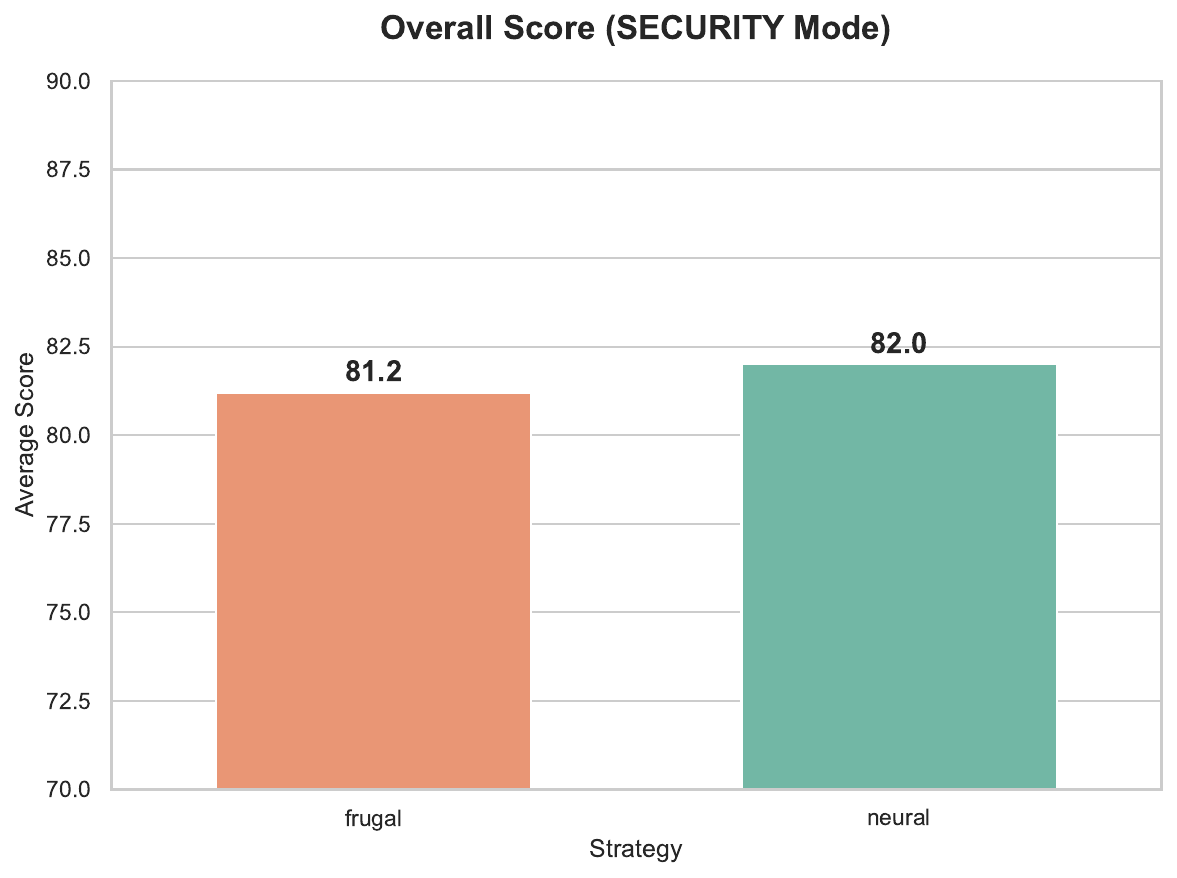}
        \caption{Cybersecurity} 
        \label{fig:security_inf}
    \end{subfigure}
    \caption{\textbf{Comparison of overall performance scores between CASTER and FrugalGPT across varying task difficulties.} The chart displays the average success rates achieved by both strategies on a benchmark containing 10 groups of high-difficulty tasks. CASTER maintains a high performance level comparable to the strong model baseline (GPT-4o), particularly in complex logic and architectural tasks. In contrast, FrugalGPT (Cascade) exhibits performance fluctuations on hard tasks, suggesting that relying on weak models for initial attempts not only incurs higher costs but may also compromise final output quality due to flawed initial reasoning.}
    \label{fig:score_comparison_with_frugal}
\end{figure}
\begin{table}[t]
    \centering
    \caption{\textbf{Performance comparison of dynamic strategies.} FrugalGPT (Cascade) vs. CASTER. The table compares the average quality scores (0-100) across four domains. While FrugalGPT often settles for "acceptable" outputs from weaker models to save cost, the CASTER consistently achieves higher quality scores. Combined with Table \ref{tab:frugal_vs_neural} (Cost Analysis), this confirms that our approach yields a \textit{Pareto-superior} outcome: lower cost AND higher performance.}
    \label{tab:frugal_vs_neural_performance}
    \begin{tabular}{llcc}
        \toprule
        \textbf{Scenario} & \textbf{Strategy} & \textbf{Average Score} & \textbf{Quality Gain} \\
        \midrule
        \multirow{2}{*}{\textbf{Software}} 
        & FrugalGPT (Cascade) & 79.8 & - \\ 
        & \textbf{CASTER} & \textbf{80.8} & \textbf{+1.0} \\ 
        \midrule
        \multirow{2}{*}{\textbf{Data}} 
        & FrugalGPT (Cascade) & 72.3& - \\ 
        & \textbf{CASTER} & \textbf{73.0}& \textbf{+0.7}\\ 
        \midrule
        \multirow{2}{*}{\textbf{Science}} 
        & FrugalGPT (Cascade) & 90.9 & - \\ 
        & \textbf{CASTER} & \textbf{92.1} & \textbf{+1.2} \\ 
        \midrule
        \multirow{2}{*}{\textbf{Security}} 
        & FrugalGPT (Cascade) & 81.2 & - \\ 
        & \textbf{CASTER} & \textbf{82.0} & \textbf{+0.8} \\ 
        \bottomrule
    \end{tabular}
\end{table}
\begin{figure}[ht]
    \centering
    \begin{subfigure}{0.48\linewidth}
        \centering
        \includegraphics[width=\linewidth]{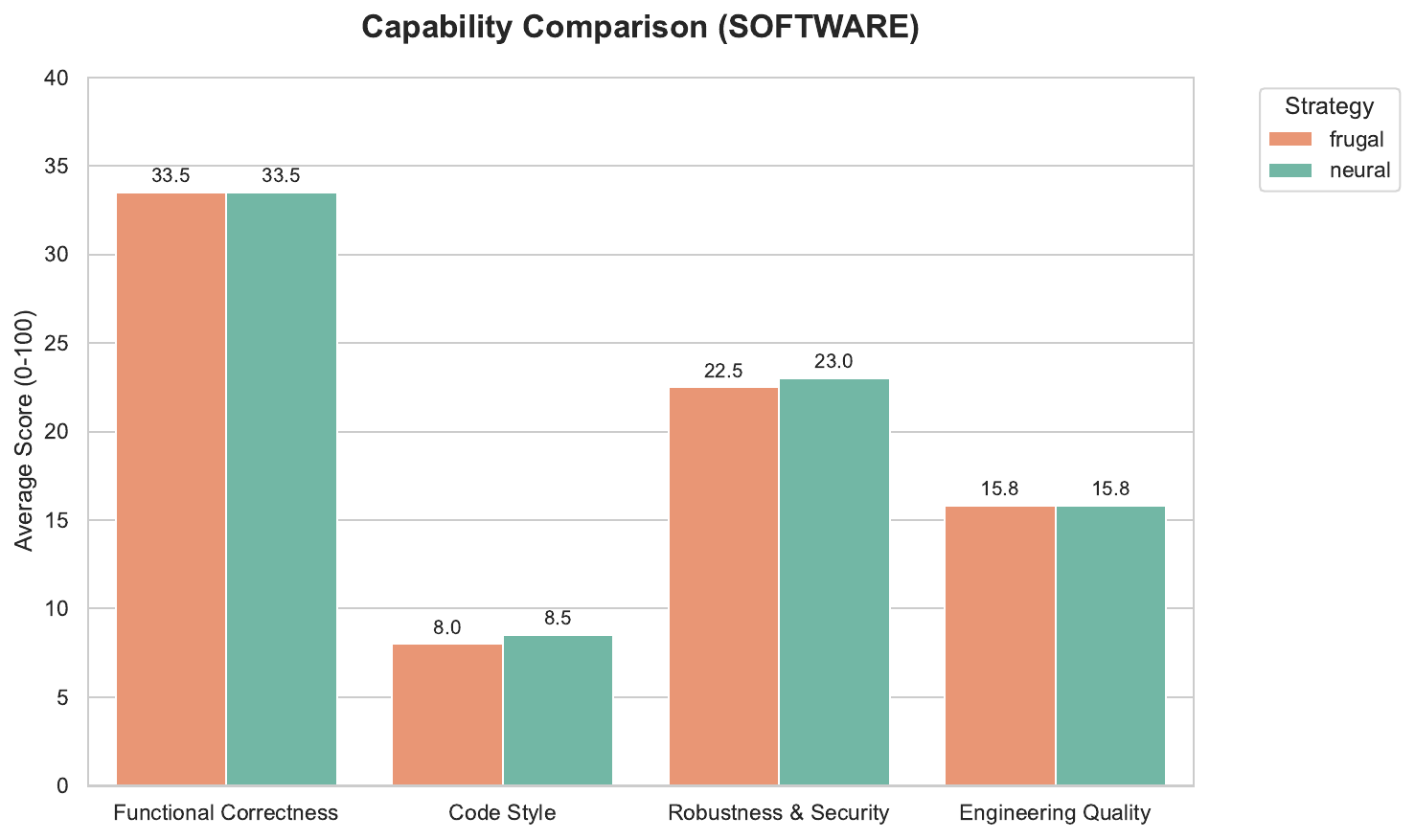}
        \caption{Software Engineering}
        \label{fig:capability_software_with_frugal}
    \end{subfigure}
    \hfill 
    \begin{subfigure}{0.48\linewidth}
        \centering
        \includegraphics[width=\linewidth]{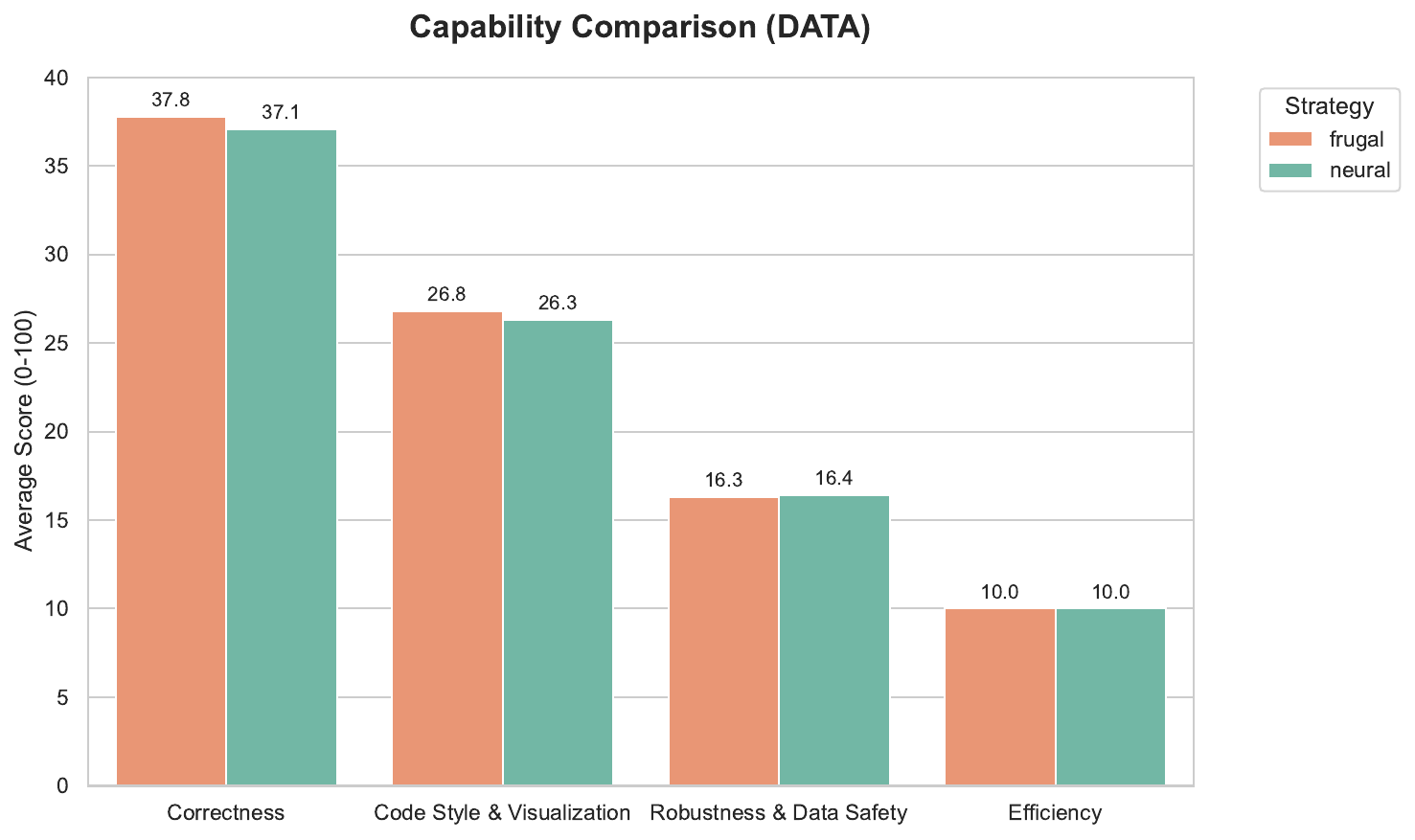}
        \caption{Data Analysis}
        \label{fig:capability_data_with_frugal}
    \end{subfigure}
    \begin{subfigure}{0.48\linewidth}
        \centering
        \includegraphics[width=\linewidth]{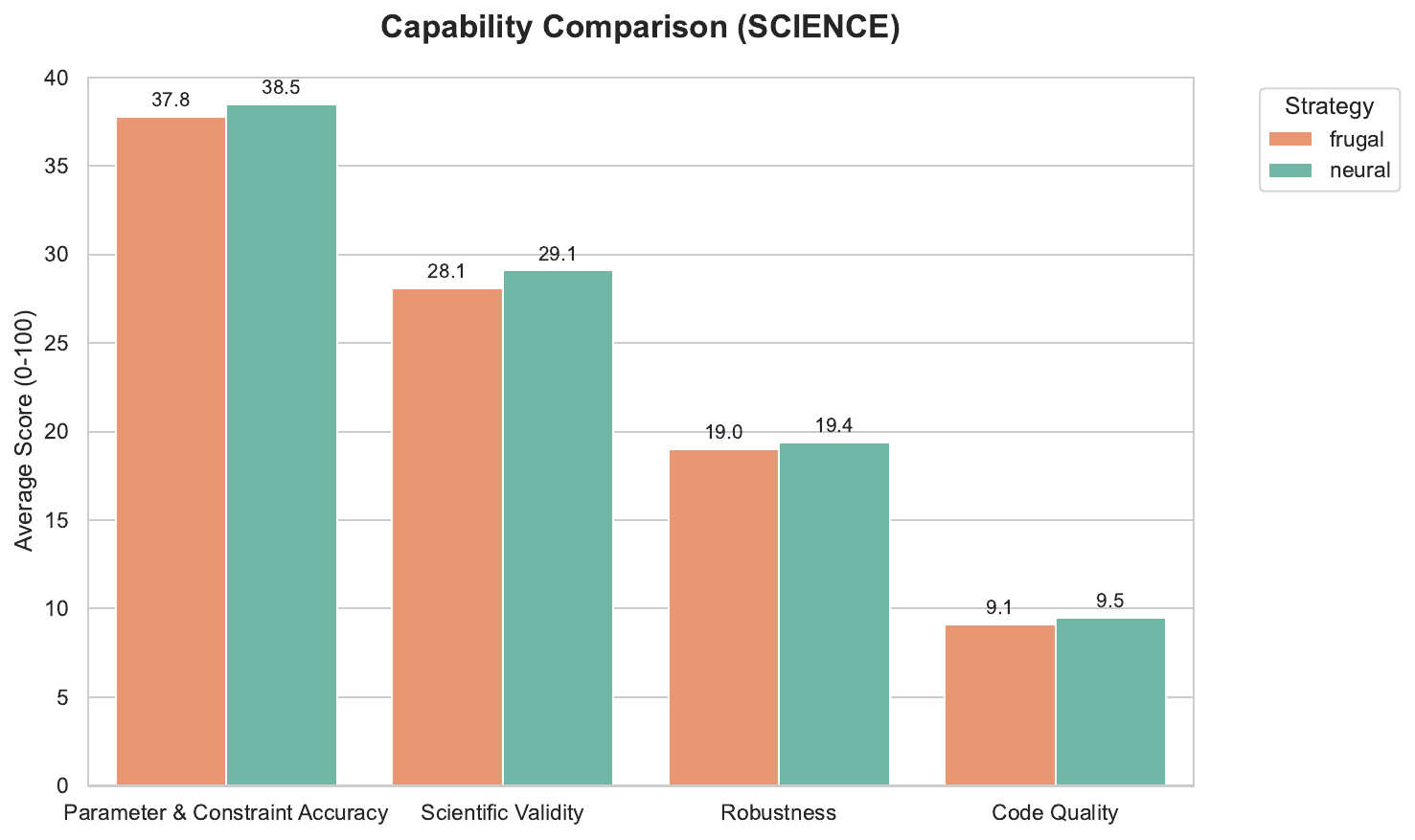}
        \caption{Scientific Discovery} 
        \label{fig:science_inf}
    \end{subfigure}
    \hfill 
    \begin{subfigure}{0.48\linewidth}
        \centering
        \includegraphics[width=\linewidth]{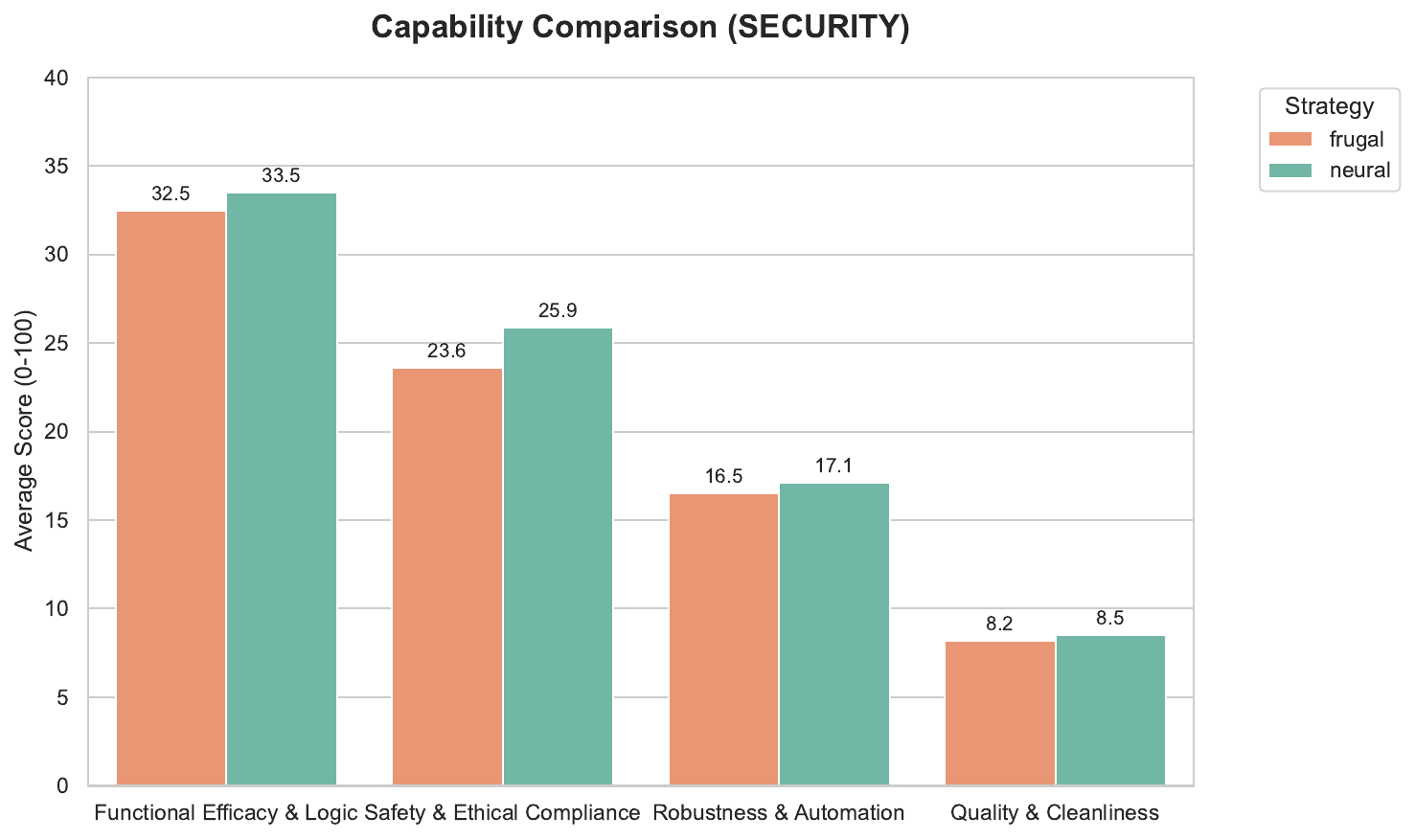}
        \caption{Cybersecurity} 
        \label{fig:security_inf}
    \end{subfigure}
    \caption{\textbf{Comprehensive Cost-Performance comparison of CASTER, FrugalGPT, and GPT-4 Baseline across Software Engineering and Data Analysis domains.} The bar charts illustrate the Average Success Rate and Average Cost per Task for three strategies in the Software and Data Analysis benchmarks. Results indicate that CASTER (green) achieves a "Pareto optimal" balance in both distinct domains: it rivals the strong baseline (GPT-4) in success rate while maintaining a cost profile comparable to the cost-aggressive FrugalGPT, demonstrating its capability as a domain-agnostic CASTER.}
    \label{fig:capability_comparison_with_frugal}
\end{figure}
\begin{table}[t]
    \centering
    \caption{\textbf{Multi-dimensional quality breakdown.} FrugalGPT vs. CASTER. The table details the performance on four sub-metrics. The CASTER consistently outperforms FrugalGPT in complex domains, particularly in \textit{Science} (e.g., Scientific Validity +1.0) and \textit{Security} (e.g., Safety +2.3).}
    \label{tab:multidimensional_frugal_vs_neural}
    \begin{tabular}{llcc}
        \toprule
        \textbf{Scenario} & \textbf{Metric} & \textbf{FrugalGPT} & \textbf{CASTER} \\
        \midrule
        \multirow{4}{*}{\textbf{Software}} 
        & Functional Correctness & 33.5 & 33.5 \\ 
        & Robustness \& Security & 22.5 & \textbf{23.0} \\
        & Engineering Quality    & 15.8 & 15.8 \\
        & Code Style             & 8.0  & \textbf{8.5} \\ 
        \midrule
        \multirow{4}{*}{\textbf{Data}} 
        & Correctness                 & \textbf{37.8} & 37.1 \\ 
        & Code Style \& Visualization & \textbf{26.8} & 26.3 \\
        & Robustness \& Data Safety   & 16.3 & \textbf{16.4} \\ 
        & Efficiency                  & 10.0 & 10.0 \\ 
        \midrule
        \multirow{4}{*}{\textbf{Science}} 
        & Parameter \& Constraint Accuracy & 37.8 & \textbf{38.5} \\
        & Scientific Validity              & 28.1 & \textbf{29.1} \\ 
        & Robustness                       & 19.0 & \textbf{19.4} \\
        & Code Quality                     & 9.1  & \textbf{9.5} \\
        \midrule
        \multirow{4}{*}{\textbf{Security}} 
        & Functional Efficacy \& Logic   & 32.5 & \textbf{33.5} \\
        & Safety \& Ethical Compliance   & 23.6 & \textbf{25.9} \\ 
        & Robustness \& Automation       & 16.5 & \textbf{17.1} \\ 
        & Quality \& Cleanliness         & 8.2  & \textbf{8.5}  \\ 
        
        \bottomrule
    \end{tabular}
    
    \vspace{3pt}
\end{table}
\begin{figure}[ht]
    \centering
    \begin{subfigure}{0.48\linewidth}
        \centering
        \includegraphics[width=\linewidth]{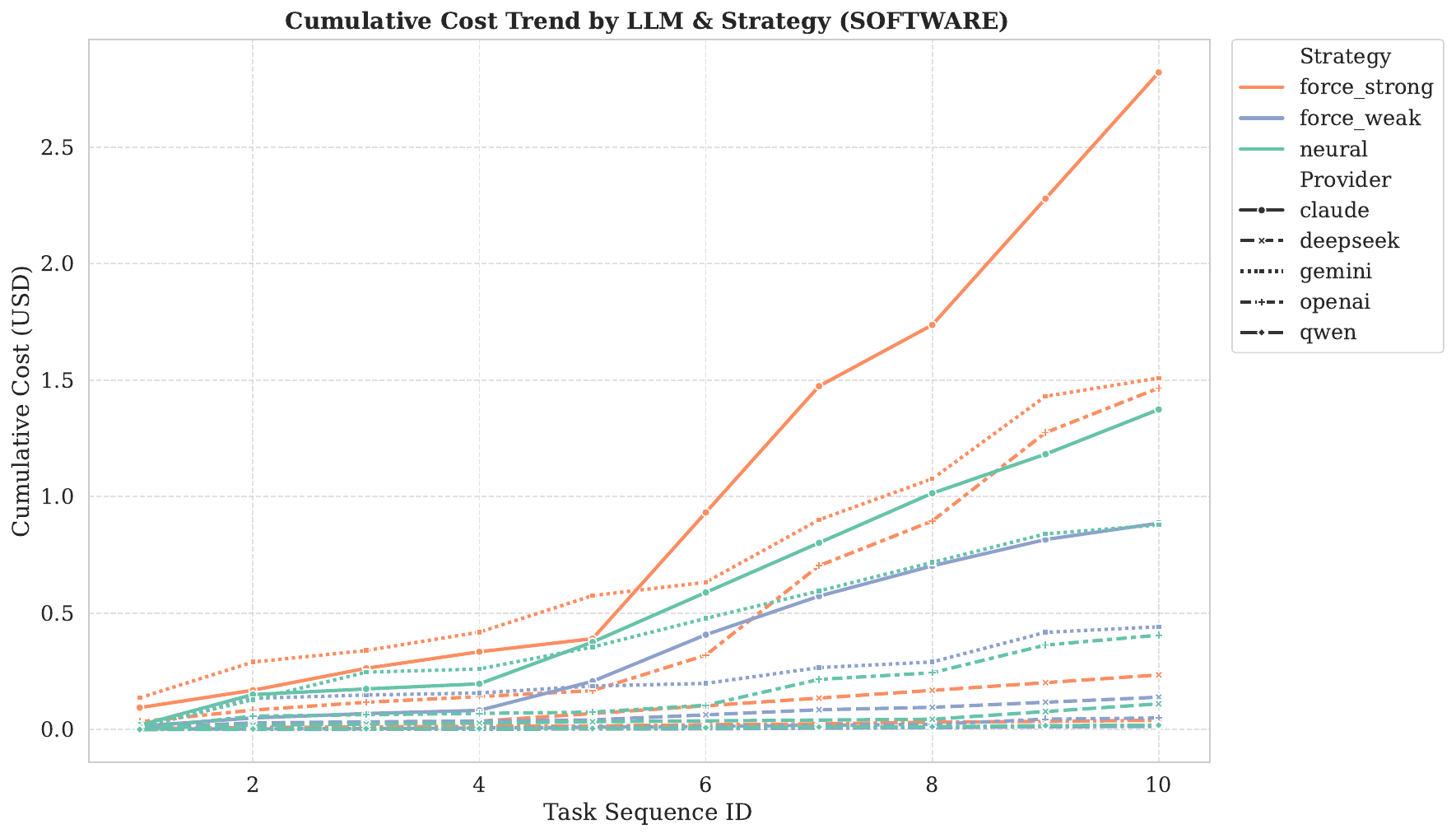}
        \caption{Software Engineering}
        \label{fig:software_cost_LLM_comp}
    \end{subfigure}
    \hfill 
    \begin{subfigure}{0.48\linewidth}
        \centering
        \includegraphics[width=\linewidth]{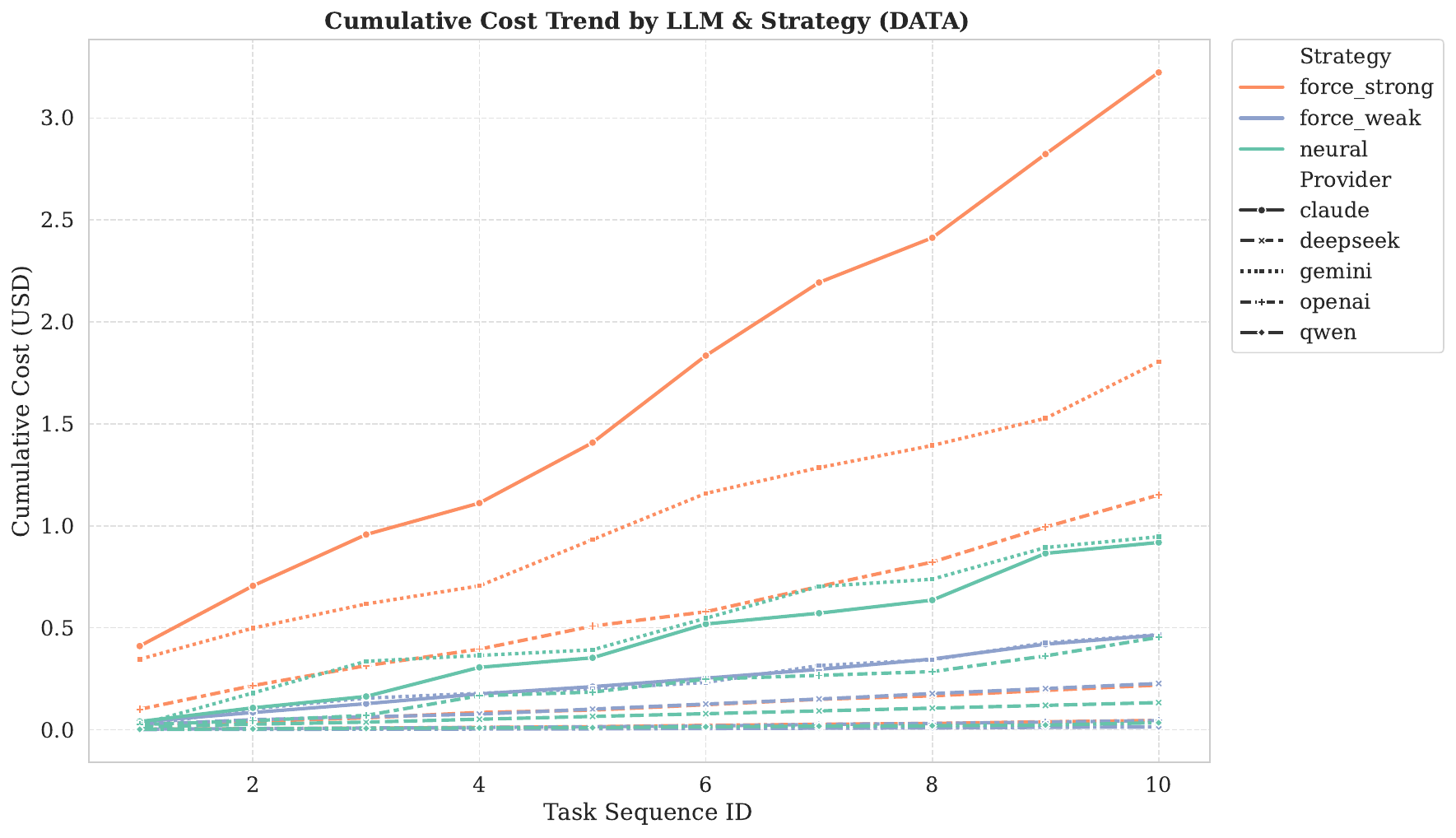}
        \caption{Data Analysis}
        \label{fig:data_cost_LLM_comp}
    \end{subfigure}
    \begin{subfigure}{0.48\linewidth}
        \centering
        \includegraphics[width=\linewidth]{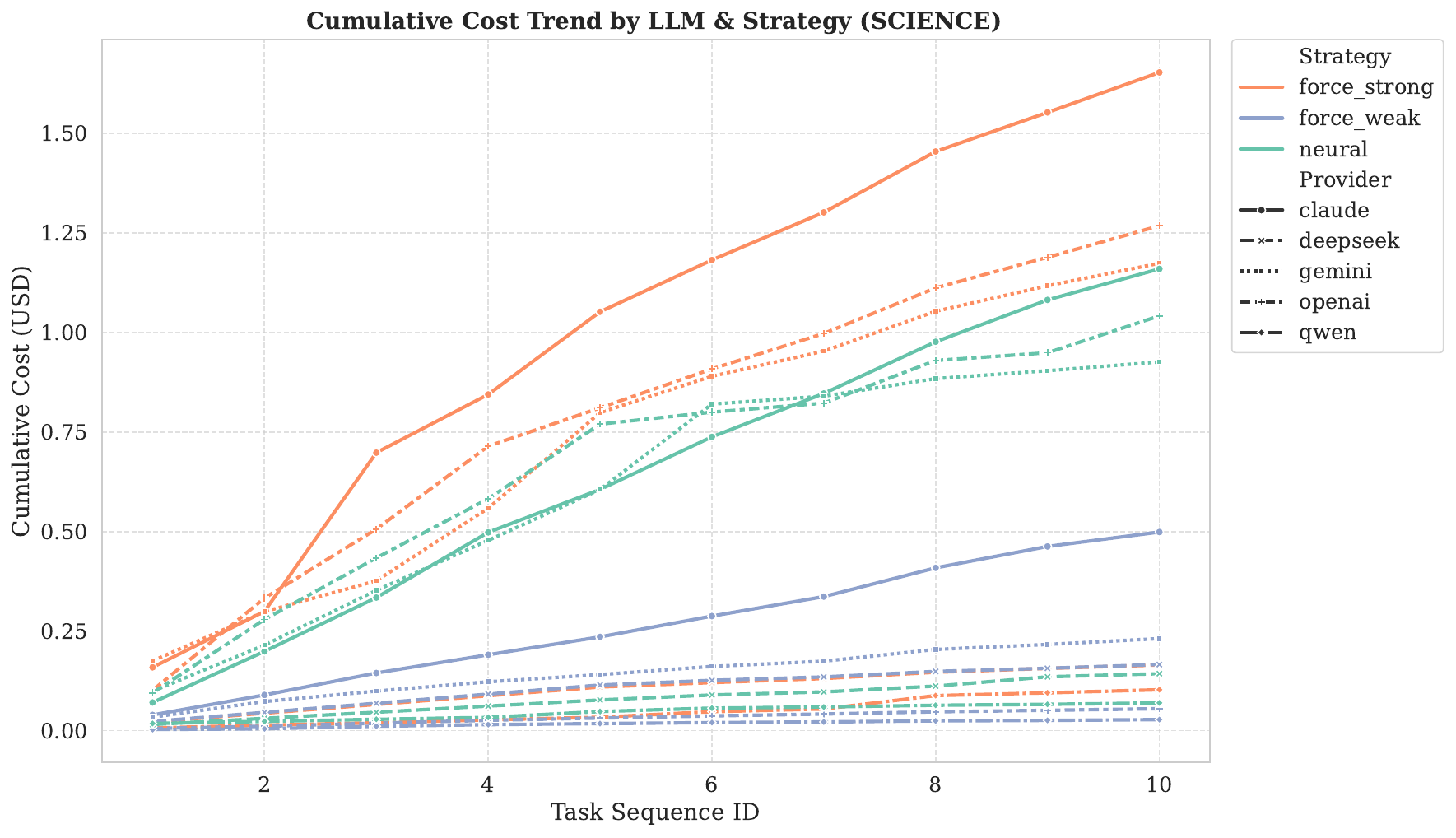}
        \caption{Scientific Discovery} 
        \label{fig:science_cost_LLM_comp}
    \end{subfigure}
    \hfill 
    \begin{subfigure}{0.48\linewidth}
        \centering
        \includegraphics[width=\linewidth]{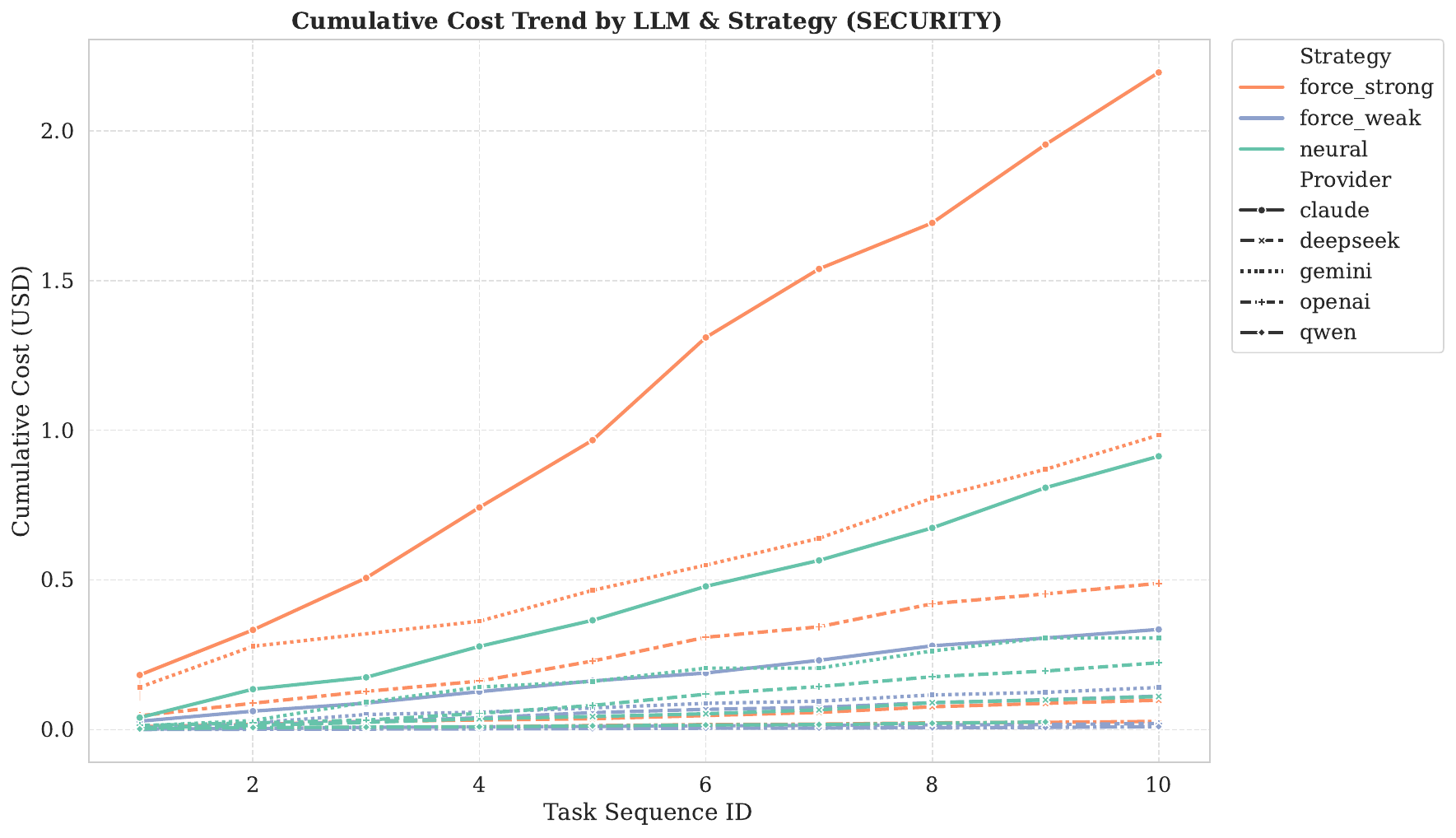}
        \caption{Cybersecurity} 
        \label{fig:security_cost_LLM_comp}
    \end{subfigure}
    \caption{\textbf{Comparative analysis of cumulative cost trends across four domains. }(a) Software Engineering, (b) Data Analysis, (c) Science Discovery, and (d) Cybersecurity. The results consistently demonstrate that Claude, OpenAI, and Gemini incur significantly higher costs compared to Qwen and DeepSeek.}
    \label{fig:cost_LLM_comp}
\end{figure}

\begin{table}[t]
    \centering
    \caption{\textbf{Comparative analysis of dynamic routing strategies.} The results are split into two columns to optimize space. CASTER consistently shows cost reductions compared to the Strong baseline.}
    \label{tab:cost_LLM_comp}
    
    \begin{minipage}{0.48\textwidth}
        \centering
        \resizebox{\linewidth}{!}{%
        \begin{tabular}{lllcc}
            \toprule
            \textbf{Scenario} & \textbf{Model} & \textbf{Strategy} & \textbf{Cost} & \textbf{Red.} \\
            \midrule
        \textbf{Software} & claude & Force Strong& \$2.8204& - \\
         &  & Force Weak & \$0.8858 & 68.6\%\\
         &  & \textbf{CASTER} & \$1.3738& 51.3\%\\
         & deepseek & Force Strong& \$0.2346& - \\
         &  & Force Weak & \$0.1391& 40.7\%\\
         &  & \textbf{CASTER} & \$0.1106& 52.9\%\\
         & gemini & Force Strong& \$1.5087 & - \\
         &  & Force Weak & \$0.4412& 70.8\%\\
         &  & \textbf{CASTER} & \$0.8783& 41.8\%\\
         & openai & Force Strong& \$1.4658 & - \\
         &  & Force Weak & \$0.0498 & 96.6\% \\
         &  & \textbf{CASTER} & \$0.4052 & 72.4\% \\
         & qwen & Force Strong& \$0.0397 & - \\
         &  & Force Weak & \$0.0138 & 65.2\% \\
         &  & \textbf{CASTER} & \$0.0186 & 53.1\% \\
        \midrule
        \textbf{Data} & claude & Force Strong& \$3.2237& - \\
         &  & Force Weak & \$0.4641 & 85.6\%\\
         &  & \textbf{CASTER} & \$0.9186 & 71.5\%\\
         & deepseek & Force Strong& \$0.2193& - \\
         &  & Force Weak & \$0.2268& -3.4\%\\
         &  & \textbf{CASTER} & \$0.1332& 39.3\%\\
         & gemini & Force Strong& \$1.8052& - \\
         &  & Force Weak & \$0.3009 & 74.2\%\\
         &  & \textbf{CASTER} & \$0.3258 & 47.6\%\\
         & openai & Force Strong& \$1.1514 & - \\
         &  & Force Weak & \$0.0452 & 96.1\% \\
 & & \textbf{CASTER}& \$0.4533&60.6\%\\
         & qwen & Force Strong& \$0.0456 & - \\
         &  & Force Weak & \$0.0136 & 70.2\% \\
         &  & \textbf{CASTER} & \$0.0344 & 24.6\% \\
            \bottomrule
        \end{tabular}%
        }
    \end{minipage}
    \hfill
    \begin{minipage}{0.48\textwidth}
        \centering
        \resizebox{\linewidth}{!}{%
        \begin{tabular}{lllcc}
            \toprule
            \textbf{Scenario} & \textbf{Model} & \textbf{Strategy} & \textbf{Cost} & \textbf{Red.} \\
            \midrule
        \textbf{Science} & claude & Force Strong& \$1.6530 & - \\
         &  & Force Weak & \$0.4987 & 69.8\% \\
         &  & \textbf{CASTER} & \$1.1596 & 29.8\% \\
         & deepseek & Force Strong& \$0.1648& - \\
         &  & Force Weak & \$0.1660& -0.7\%\\
         &  & \textbf{CASTER} & \$0.1429& 13.3\%\\
         & gemini & Force Strong& \$1.1737 & - \\
         &  & Force Weak & \$0.2311 & 80.3\% \\
         &  & \textbf{CASTER} & \$0.9259 & 21.1\% \\
         & openai & Force Strong& \$1.2679 & - \\
         &  & Force Weak & \$0.0550 & 95.7\% \\
         &  & \textbf{CASTER} & \$1.0416 & 17.8\% \\
         & qwen & Force Strong& \$0.1026 & - \\
         &  & Force Weak & \$0.0276 & 73.1\% \\
         &  & \textbf{CASTER} & \$0.0695 & 32.3\% \\
        \midrule
        \textbf{Security} & claude & Force Strong& \$2.1948 & - \\
         &  & Force Weak & \$0.3346 & 84.8\% \\
         &  & \textbf{CASTER} & \$0.9129 & 58.4\% \\
         & deepseek & Force Strong& \$0.0983 & - \\
         &  & Force Weak & \$0.1118 & -13.7\% \\
         &  & \textbf{CASTER} & \$0.1105 & -12.4\% \\
         & gemini & Force Strong& \$0.9837 & - \\
         &  & Force Weak & \$0.1408 & 85.7\% \\
         &  & \textbf{CASTER} & \$0.3064 & 68.9\% \\
         & openai & Force Strong& \$0.4886 & - \\
         &  & Force Weak & \$0.0207 & 95.8\% \\
         &  & \textbf{CASTER} & \$0.2234 & 54.3\% \\
         & qwen & Force Strong& \$0.0274 & - \\
         &  & Force Weak & \$0.0100 & 63.5\% \\
         &  & \textbf{CASTER} & \$0.0257 & 6.2\% \\
            \bottomrule
        \end{tabular}%
        }
    \end{minipage}
\end{table}

\begin{figure}[ht]
    \centering
    \begin{subfigure}{0.48\linewidth}
        \centering
        \includegraphics[width=\linewidth]{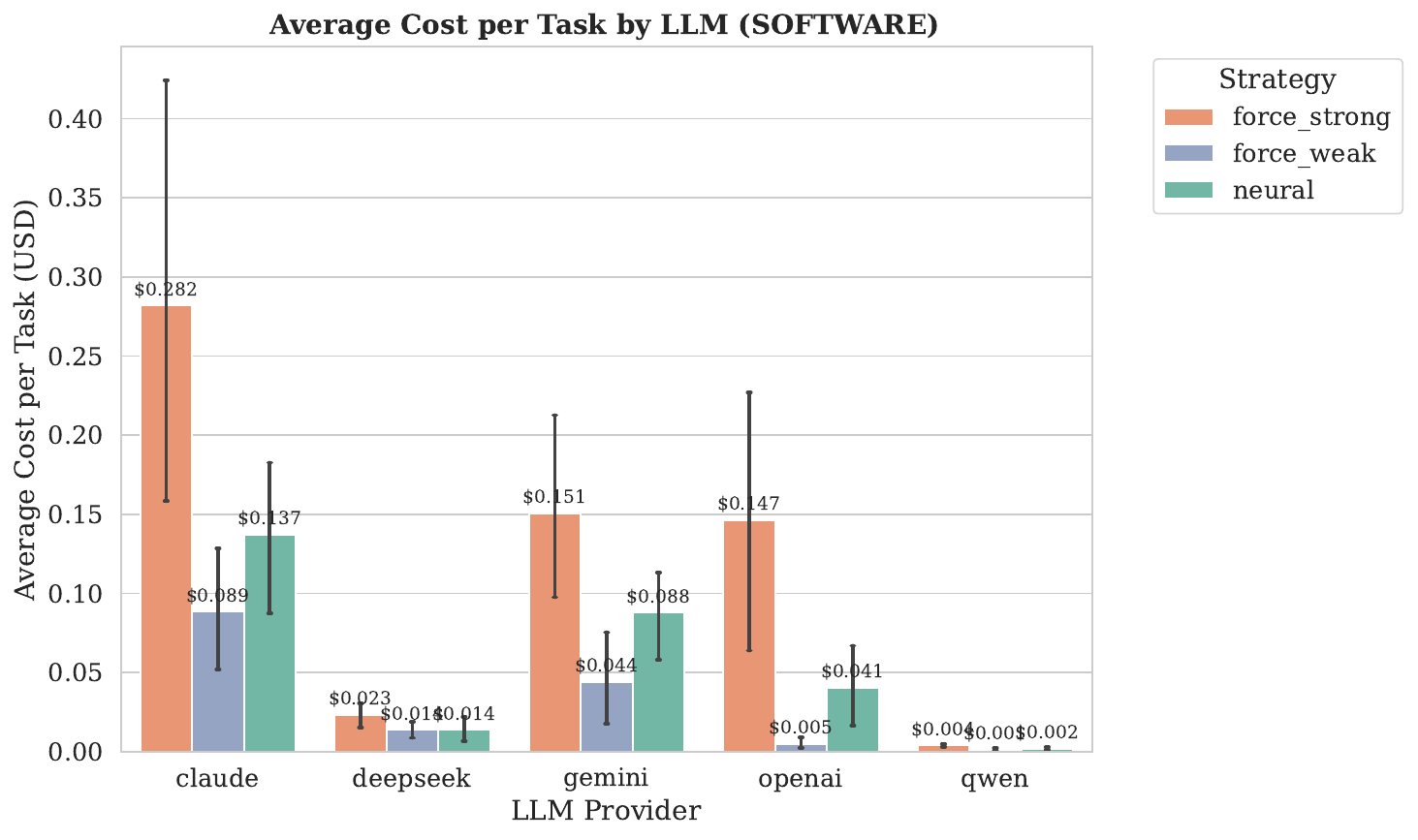}
        \caption{Software Engineering}
        \label{fig:software_avg_cost_LLM_comp}
    \end{subfigure}
    \hfill 
    \begin{subfigure}{0.48\linewidth}
        \centering
        \includegraphics[width=\linewidth]{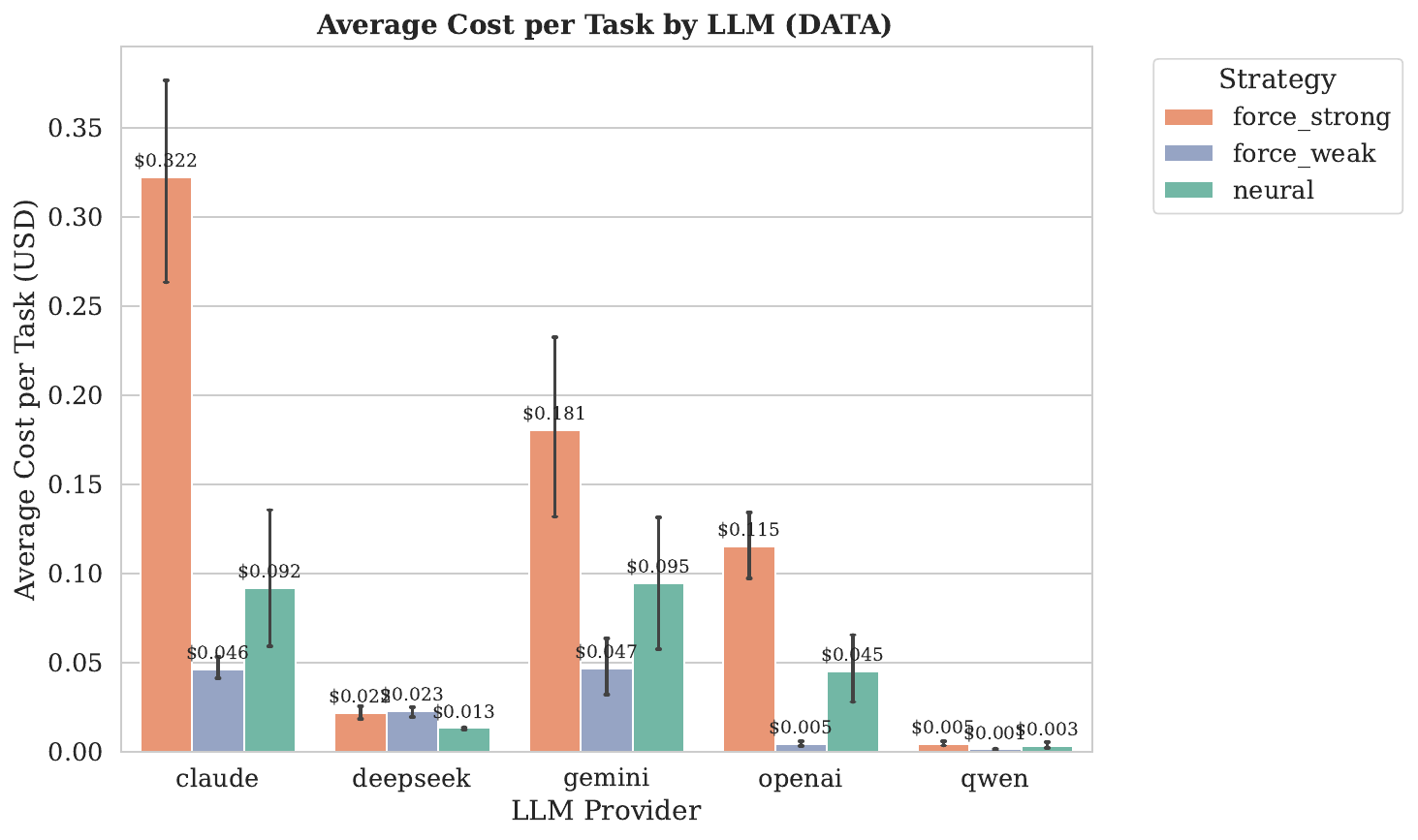}
        \caption{Data Analysis}
        \label{fig:data_avg_cost_LLM_comp}
    \end{subfigure}
    \begin{subfigure}{0.48\linewidth}
        \centering
        \includegraphics[width=\linewidth]{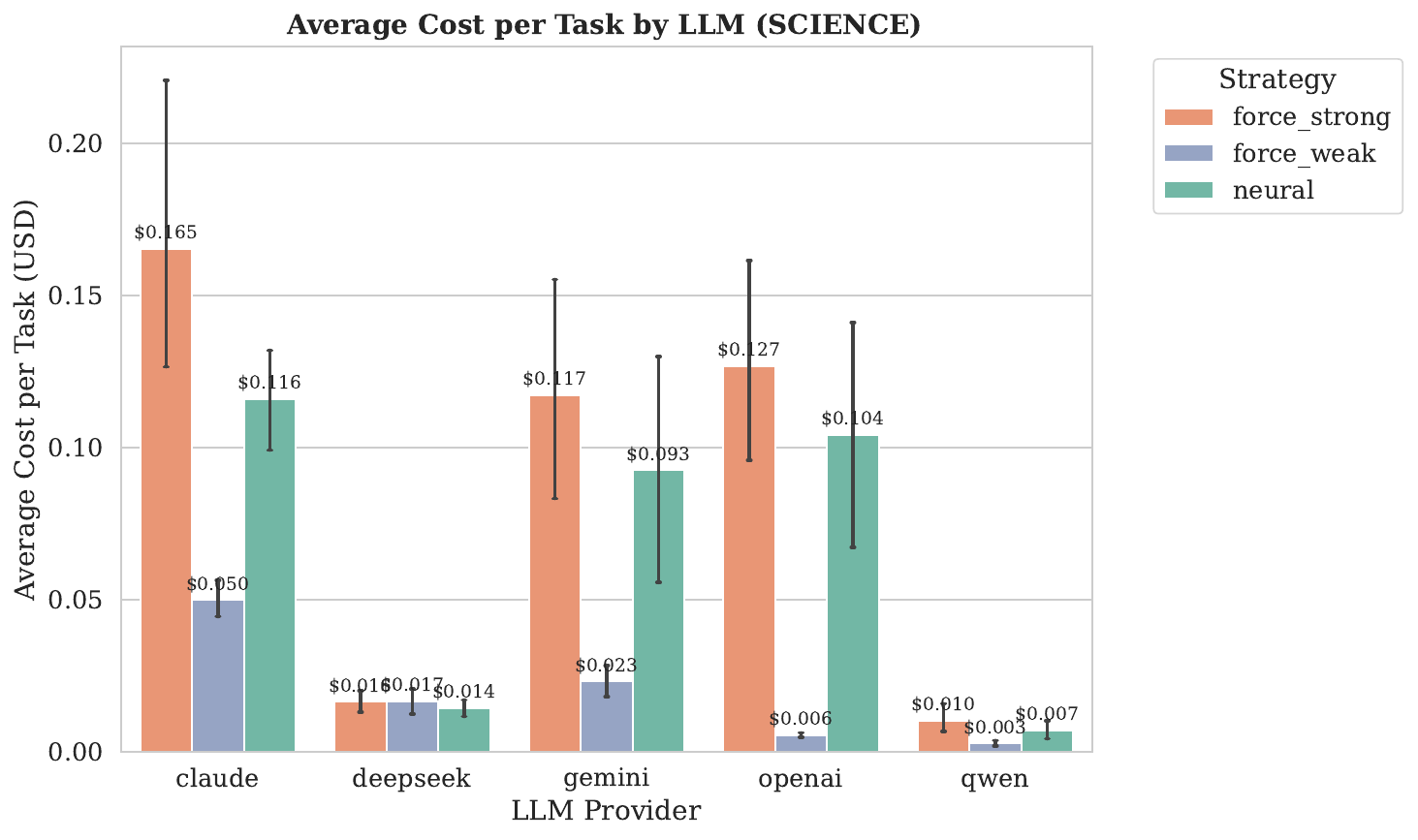}
        \caption{Scientific Discovery} 
        \label{fig:science_avg_cost_LLM_comp}
    \end{subfigure}
    \hfill 
    \begin{subfigure}{0.48\linewidth}
        \centering
        \includegraphics[width=\linewidth]{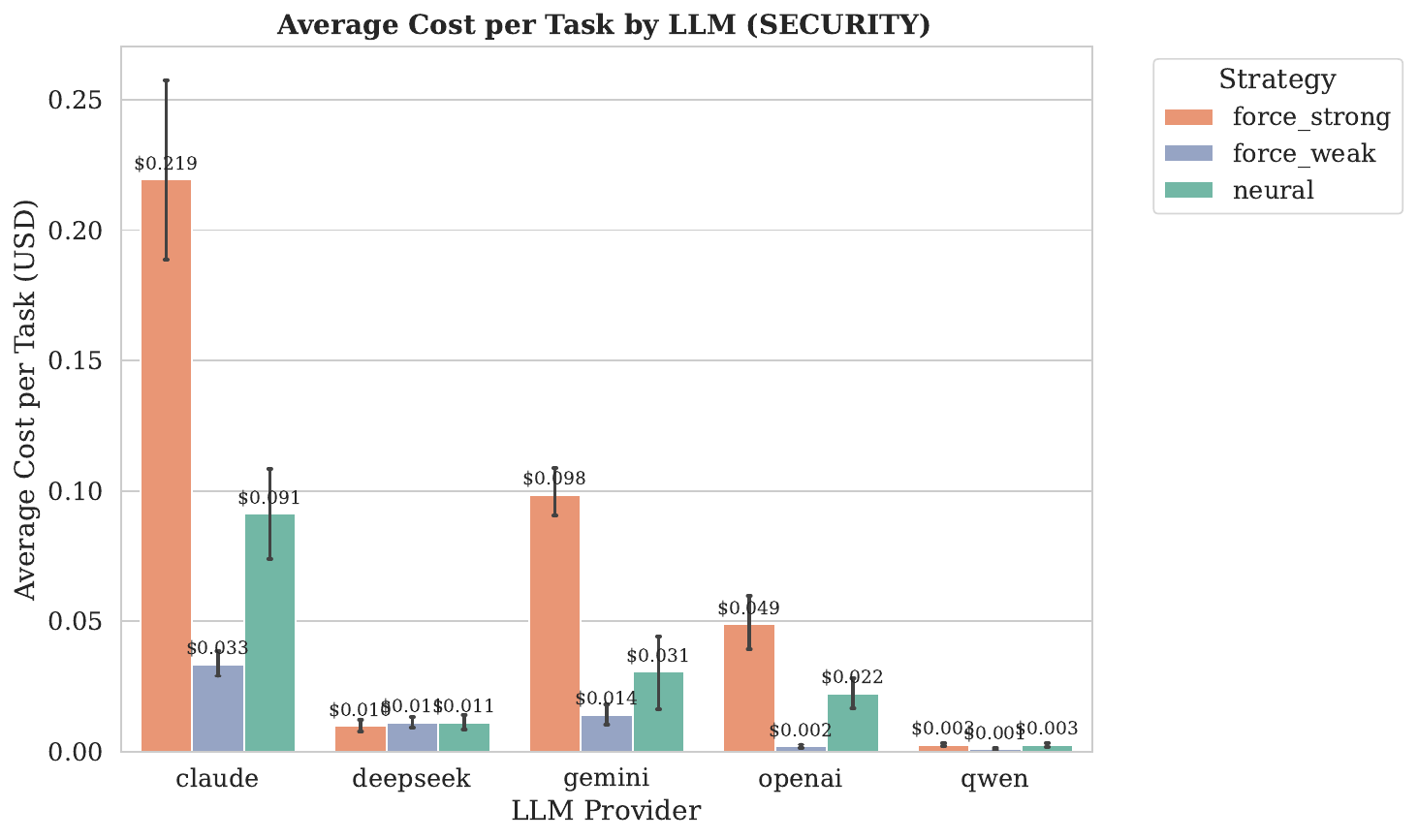}
        \caption{Cybersecurity} 
        \label{fig:security_avg_cost_LLM_comp}
    \end{subfigure}
    \caption{\textbf{Comparison of average cost per task across four domains.} (a) Software Engineering, (b) Data Analysis, (c) Science Discovery, and (d) Cybersecurity. The bar charts clearly illustrate that the CASTER strategy (green) achieves substantial cost reductions compared to the strong model baseline (orange), particularly for high-cost providers such as Claude, Gemini, and OpenAI. This demonstrates the effectiveness of dynamic routing in controlling average budgets without solely relying on expensive models.}
    \label{fig:avg_cost_LLM_comp}
\end{figure}
\begin{table}[t]
    \centering
    \caption{\textbf{Comparative analysis of Average Cost per Task.} This table details the average expenditure for each task across different models and routing strategies. The results highlight the cost-efficiency of the CASTER strategy.}
    \label{tab:avg_cost_LLM_comp}
    
    \begin{minipage}{0.48\textwidth}
        \centering
        \resizebox{\linewidth}{!}{%
        \begin{tabular}{lllcc}
            \toprule
            \textbf{Scenario} & \textbf{Model} & \textbf{Strategy} & \textbf{Avg. Cost} & \textbf{Red.} \\
            \midrule
        \textbf{Software} & claude & Force Strong& \$0.2820 & - \\
         &  & Force Weak & \$0.0886 & 68.6\% \\
         &  & \textbf{CASTER} & \$0.1374 & 51.3\% \\
         & deepseek & Force Strong& \$0.0235 & - \\
         &  & Force Weak & \$0.0139 & 40.7\% \\
         &  & \textbf{CASTER} & \$0.0111 & 52.9\% \\
         & gemini & Force Strong& \$0.1509 & - \\
         &  & Force Weak & \$0.0441 & 70.8\% \\
         &  & \textbf{CASTER} & \$0.0878 & 41.8\% \\
         & openai & Force Strong& \$0.1466 & - \\
         &  & Force Weak & \$0.0050 & 96.6\% \\
         &  & \textbf{CASTER} & \$0.0405 & 72.4\% \\
         & qwen & Force Strong& \$0.0040 & - \\
         &  & Force Weak & \$0.0014 & 65.2\% \\
         &  & \textbf{CASTER} & \$0.0019 & 53.1\% \\
        \midrule
        \textbf{Data} & claude & Force Strong& \$0.3224& - \\
         &  & Force Weak & \$0.0464 & 85.6\% \\
         &  & \textbf{CASTER} & \$0.0919 & 71.5\% \\
         & deepseek & Force Strong& \$0.0219 & - \\
         &  & Force Weak & \$0.0227 & -3.4\% \\
         &  & \textbf{CASTER} & \$0.0133 & 39.3\% \\
         & gemini & Force Strong& \$0.1805& - \\
         &  & Force Weak & \$0.0301 & 83.3\% \\
         &  & \textbf{CASTER} & \$0.0326 & 82.0\% \\
         & openai & Force Strong& \$0.1151 & - \\
         &  & Force Weak & \$0.0045 & 96.1\% \\
         &  & \textbf{CASTER} & \$0.0453 & 60.6\% \\
         & qwen & Force Strong& \$0.0046 & - \\
         &  & Force Weak & \$0.0014 & 70.2\% \\
         &  & \textbf{CASTER} & \$0.0034 & 24.6\% \\
            \bottomrule
        \end{tabular}%
        }
    \end{minipage}
    \hfill
    \begin{minipage}{0.48\textwidth}
        \centering
        \resizebox{\linewidth}{!}{%
        \begin{tabular}{lllcc}
            \toprule
            \textbf{Scenario} & \textbf{Model} & \textbf{Strategy} & \textbf{Avg. Cost} & \textbf{Red.} \\
            \midrule
        \textbf{Science} & claude & Force Strong& \$0.1653 & - \\
         &  & Force Weak & \$0.0499 & 69.8\% \\
         &  & \textbf{CASTER} & \$0.1160 & 29.8\% \\
         & deepseek & Force Strong& \$0.0165 & - \\
         &  & Force Weak & \$0.0166 & -0.7\% \\
         &  & \textbf{CASTER} & \$0.0143 & 13.3\% \\
         & gemini & Force Strong& \$0.1174 & - \\
         &  & Force Weak & \$0.0231 & 80.3\% \\
         &  & \textbf{CASTER} & \$0.0926 & 21.1\% \\
         & openai & Force Strong& \$0.1268 & - \\
         &  & Force Weak & \$0.0055 & 95.7\% \\
         &  & \textbf{CASTER} & \$0.1042 & 17.8\% \\
         & qwen & Force Strong& \$0.0103 & - \\
         &  & Force Weak & \$0.0028 & 73.1\% \\
         &  & \textbf{CASTER} & \$0.0070 & 32.3\% \\
        \midrule
        \textbf{Security} & claude & Force Strong& \$0.2195 & - \\
         &  & Force Weak & \$0.0335 & 84.8\% \\
         &  & \textbf{CASTER} & \$0.0913 & 58.4\% \\
         & deepseek & Force Strong& \$0.0098 & - \\
         &  & Force Weak & \$0.0112 & -13.7\% \\
         &  & \textbf{CASTER} & \$0.0111 & -12.4\% \\
         & gemini & Force Strong& \$0.0984 & - \\
         &  & Force Weak & \$0.0141 & 85.7\% \\
         &  & \textbf{CASTER} & \$0.0306 & 68.9\% \\
         & openai & Force Strong& \$0.0489 & - \\
         &  & Force Weak & \$0.0021 & 95.8\% \\
         &  & \textbf{CASTER} & \$0.0223 & 54.3\% \\
         & qwen & Force Strong& \$0.0027 & - \\
         &  & Force Weak & \$0.0010 & 63.5\% \\
         &  & \textbf{CASTER} & \$0.0026 & 6.2\% \\
            \bottomrule
        \end{tabular}%
        }
    \end{minipage}
\end{table}
\begin{figure}[ht]
    \centering
    \begin{subfigure}{0.48\linewidth}
        \centering
        \includegraphics[width=\linewidth]{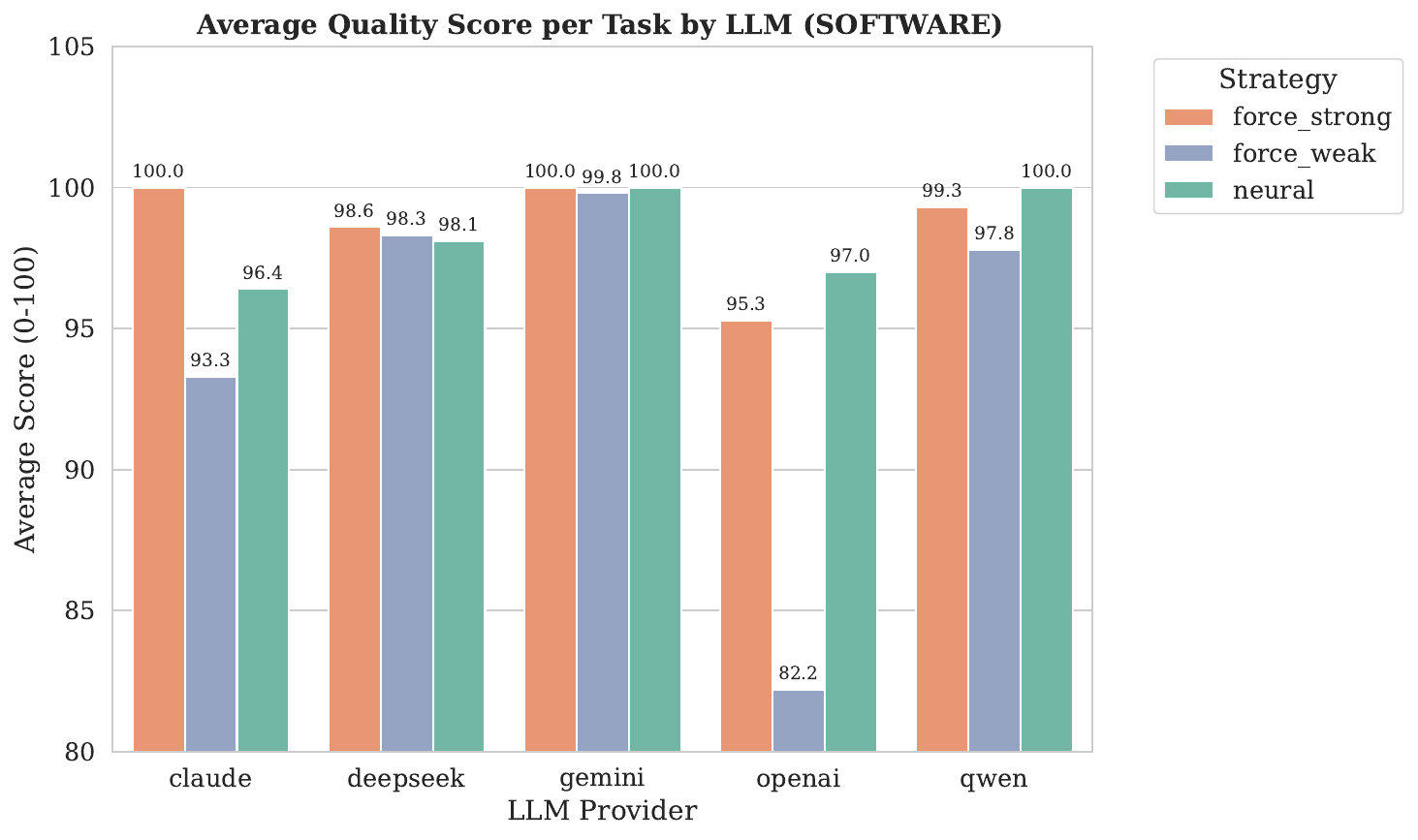}
        \caption{Software Engineering}
        \label{fig:software_score_LLM_comp}
    \end{subfigure}
    \hfill 
    \begin{subfigure}{0.48\linewidth}
        \centering
        \includegraphics[width=\linewidth]{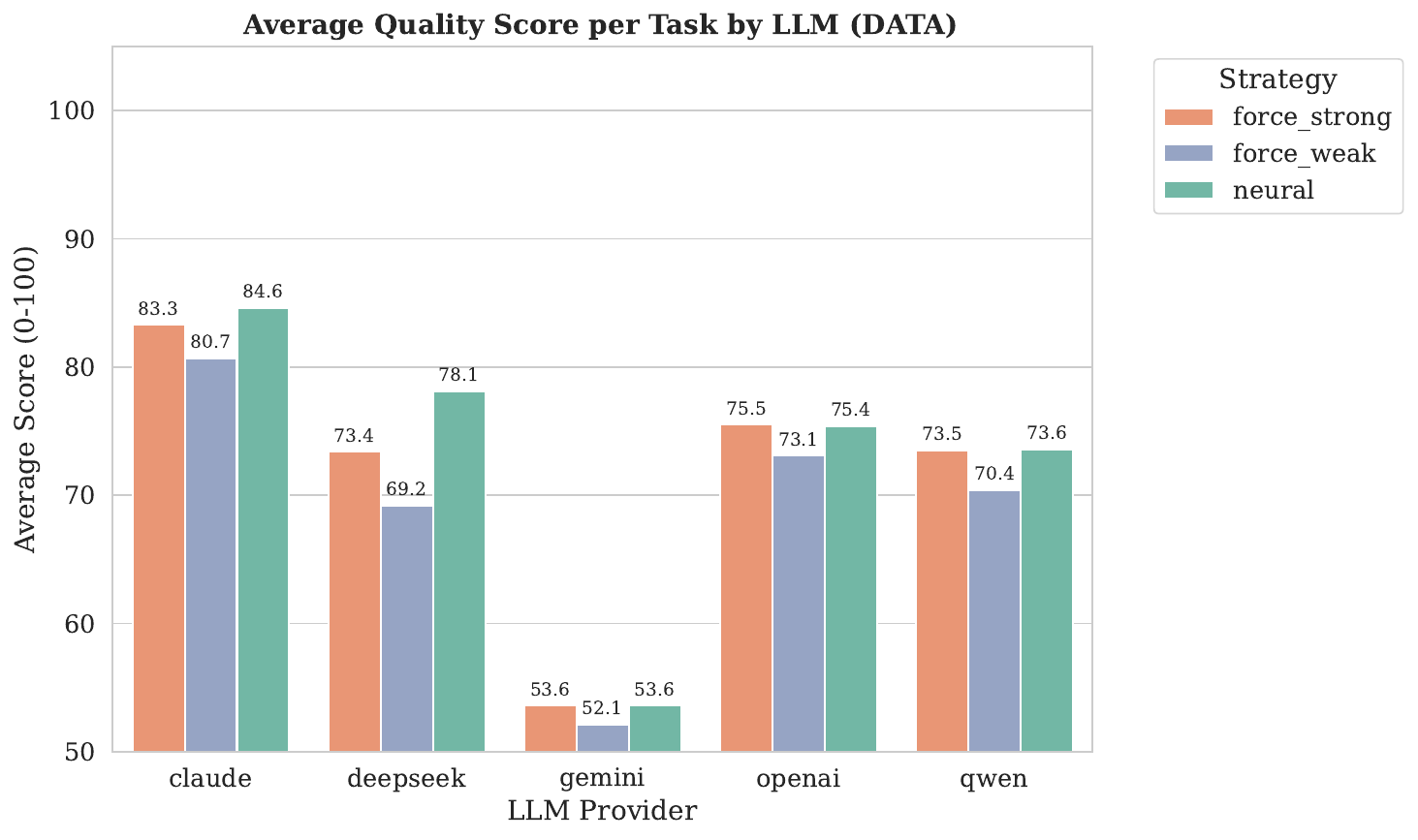}
        \caption{Data Analysis}
        \label{fig:data_score_LLM_comp}
    \end{subfigure}
    \begin{subfigure}{0.48\linewidth}
        \centering
        \includegraphics[width=\linewidth]{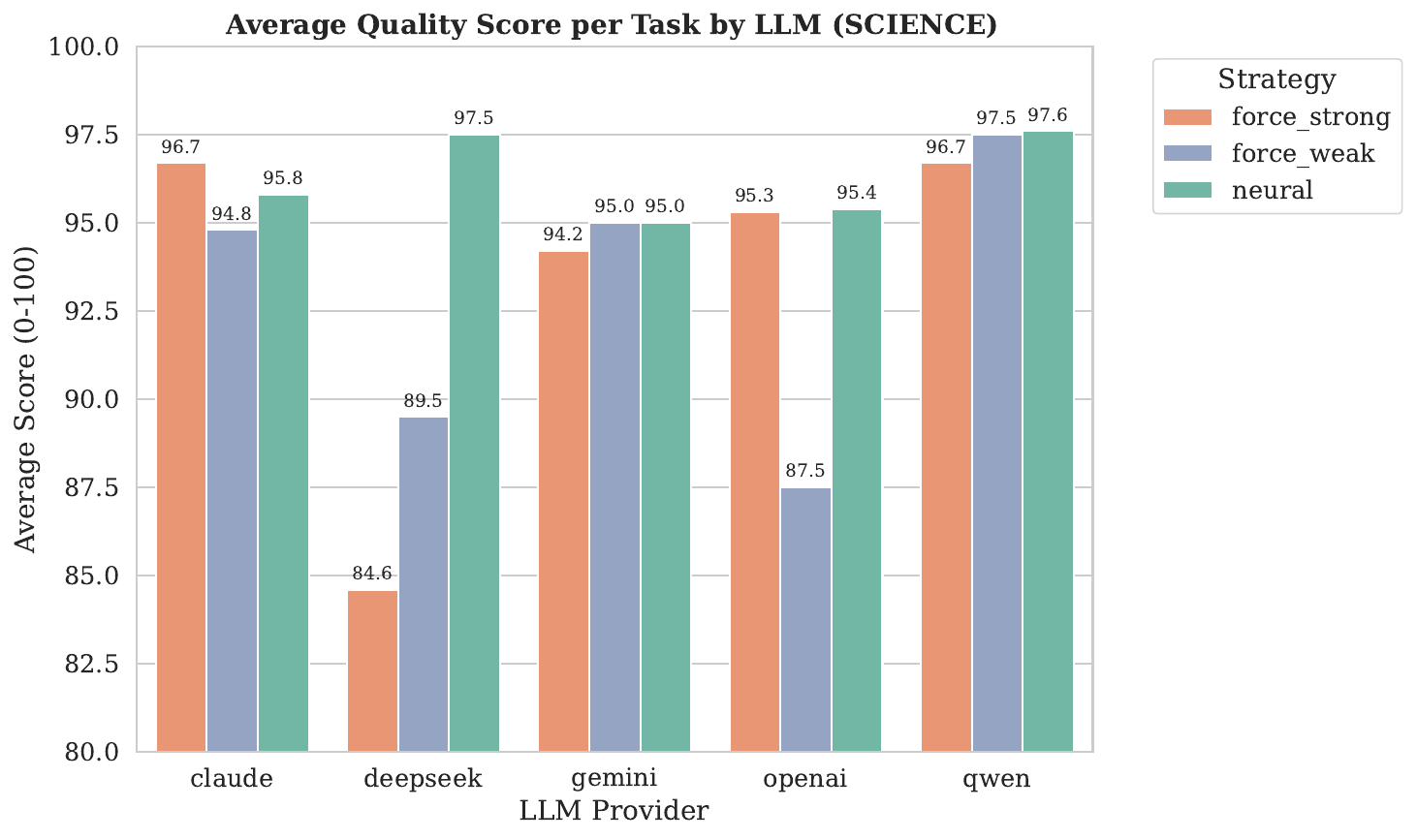}
        \caption{Scientific Discovery} 
        \label{fig:science_score_LLM_comp}
    \end{subfigure}
    \hfill 
    \begin{subfigure}{0.48\linewidth}
        \centering
        \includegraphics[width=\linewidth]{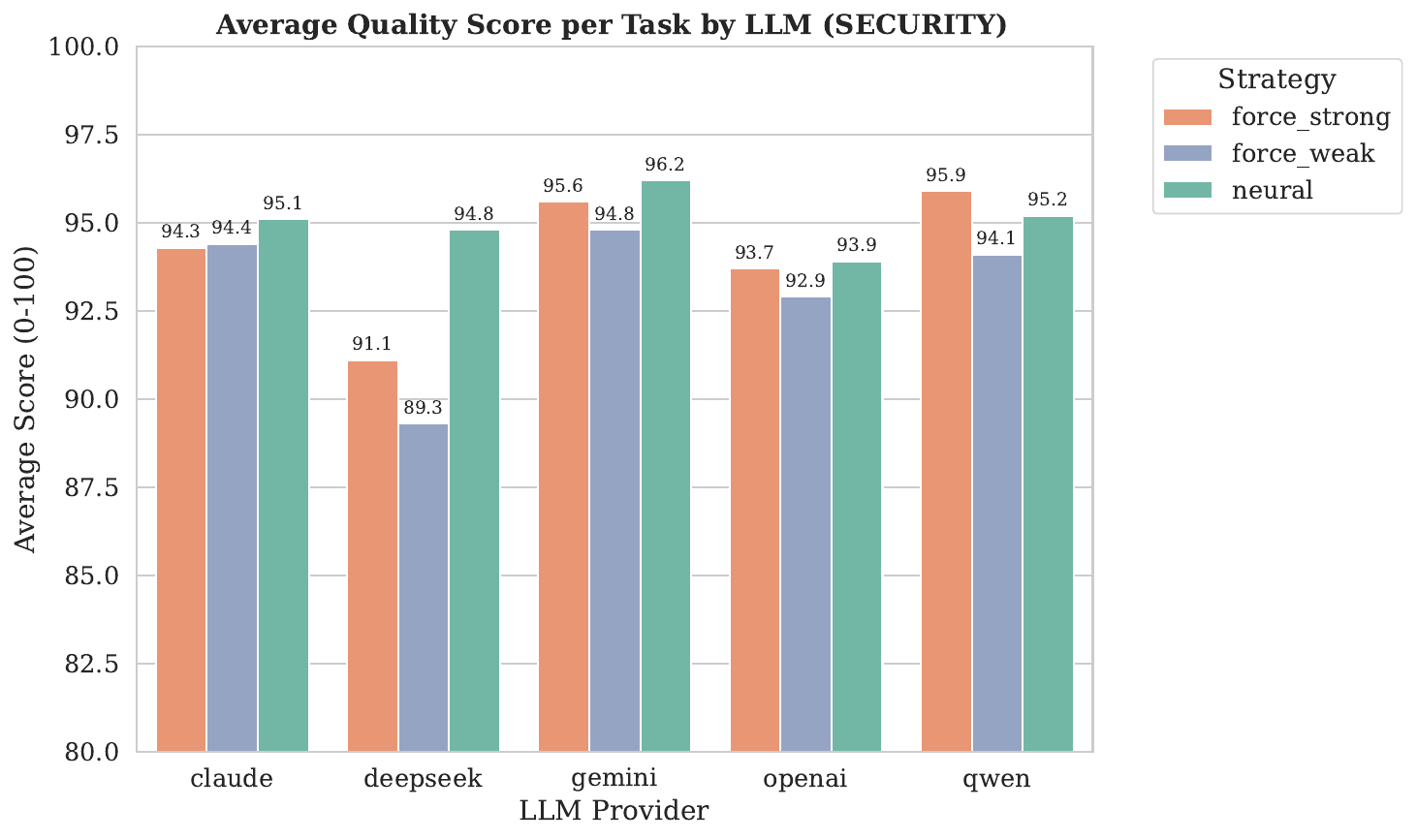}
        \caption{Cybersecurity} 
        \label{fig:security_score_LLM_comp}
    \end{subfigure}
    \caption{\textbf{Comparison of average quality scores across four domains. }(a) Software Engineering, (b) Data Analysis, (c) Science Discovery, and (d) Cybersecurity. The results indicate that despite the significant cost reductions shown in \cref{fig:software_avg_cost_LLM_comp}, the CASTER strategy (green) maintains high performance levels comparable to the strong model baseline (orange) and significantly outperforms the weak model baseline (grey). This validates the strategy's capability to optimize costs without compromising task quality.}
    \label{fig:score_LLM_comp}
\end{figure}
\begin{table}[t]
    \centering
    \caption{\textbf{Comparative analysis of Average Quality Score.} The table presents the average performance scores (0-100) across different models and routing strategies. CASTER maintains high performance comparable to Strong baselines while significantly reducing costs.}
    \label{tab:score_LLM_comp}
    
    \begin{minipage}{0.48\textwidth}
        \centering
        \resizebox{\linewidth}{!}{%
        \begin{tabular}{lllc}
            \toprule
            \textbf{Scenario} & \textbf{Model} & \textbf{Strategy} & \textbf{Avg. Score} \\
            \midrule
        \textbf{Software} & claude & Force Strong& \textbf{100.0} \\
         &  & Force Weak& 93.3 \\
         &  & \textbf{CASTER} & 96.4\\
         & deepseek & Force Strong& \textbf{98.6} \\
         &  & Force Weak& 98.3 \\
         &  & \textbf{CASTER} & 98.1 \\
         & gemini & Force Strong& \textbf{100.0}\\
         &  & Force Weak& 99.8\\
         &  & \textbf{CASTER} & \textbf{100.0}\\
         & openai & Force Strong& 95.3\\
         &  & Force Weak& 82.2 \\
         &  & \textbf{CASTER} & \textbf{97.0}\\
         & qwen & Force Strong& 99.3\\
         &  & Force Weak& 97.8\\
         &  & \textbf{CASTER} & \textbf{100.0} \\
        \midrule
        \textbf{Data} & claude & Force Strong& 83.3\\
         &  & Force Weak& 80.7\\
         &  & \textbf{CASTER} & \textbf{84.6}\\
         & deepseek & Force Strong& 73.4\\
         &  & Force Weak& 69.2\\
         &  & \textbf{CASTER} & \textbf{78.1}\\
         & gemini & Force Strong& \textbf{53.6}\\
         &  & Force Weak& 52.1\\
         &  & \textbf{CASTER} & \textbf{53.6}\\
         & openai & Force Strong& \textbf{75.5}\\
         &  & Force Weak& 73.1\\
         &  & \textbf{CASTER} & 75.4\\
         & qwen & Force Strong& 73.5\\
         &  & Force Weak& 70.4\\
         &  & \textbf{CASTER} & \textbf{73.6}\\
            \bottomrule
        \end{tabular}%
        }
    \end{minipage}
    \hfill
    \begin{minipage}{0.48\textwidth}
        \centering
        \resizebox{\linewidth}{!}{%
        \begin{tabular}{lllc}
            \toprule
            \textbf{Scenario} & \textbf{Model} & \textbf{Strategy} & \textbf{Avg. Score} \\
            \midrule
        \textbf{Science} & claude & Force Strong& \textbf{96.7} \\
         &  & Force Weak& 94.8 \\
         &  & \textbf{CASTER} & 95.8 \\
         & deepseek & Force Strong& 84.6 \\
         &  & Force Weak& 89.5 \\
         &  & \textbf{CASTER} & \textbf{97.5} \\
         & gemini & Force Strong& 94.2 \\
         &  & Force Weak& \textbf{95.0} \\
         &  & \textbf{CASTER} & \textbf{95.0} \\
         & openai & Force Strong& 95.3 \\
         &  & Force Weak& 87.5 \\
         &  & \textbf{CASTER} & \textbf{95.4}\\
         & qwen & Force Strong& 96.7 \\
         &  & Force Weak& 97.5 \\
         &  & \textbf{CASTER} & \textbf{97.6}\\
        \midrule
        \textbf{Security} & claude & Force Strong& 94.3 \\
         &  & Force Weak& 94.4 \\
         &  & \textbf{CASTER} & \textbf{95.1} \\
         & deepseek & Force Strong& 91.1 \\
         &  & Force Weak& 89.3 \\
         &  & \textbf{CASTER} & \textbf{94.8} \\
         & gemini & Force Strong& 95.6\\
         &  & Force Weak& 94.8\\
         &  & \textbf{CASTER} & \textbf{96.2}\\
         & openai & Force Strong& 93.7 \\
         &  & Force Weak& 92.9 \\
         &  & \textbf{CASTER} & \textbf{93.9} \\
         & qwen & Force Strong& \textbf{95.9} \\
         &  & Force Weak& 94.1\\
         &  & \textbf{CASTER} & 95.2 \\
            \bottomrule
        \end{tabular}%
        }
    \end{minipage}
\end{table}


\end{document}